\pdfoutput=1
\documentclass[11pt]{article}

\usepackage[preprint]{acl}

\renewcommand{\dblfloatpagefraction}{0.1}

\usepackage{times}
\usepackage{latexsym}
\usepackage[T1]{fontenc}
\usepackage[utf8]{inputenc}
\usepackage{microtype}
\usepackage{inconsolata}
\usepackage{graphicx}

\usepackage{booktabs}
\usepackage{amsfonts}
\usepackage{amsmath,amssymb}
\usepackage{nicefrac}
\usepackage{xcolor}
\usepackage{colortbl}
\usepackage{multirow}
\usepackage{pifont}
\usepackage{subcaption}
\usepackage{adjustbox}
\usepackage{enumitem}
\usepackage{wrapfig}
\usepackage{comment}

\setlist[itemize]{leftmargin=0.3cm, itemsep=0.2pt, parsep=0pt, topsep=1pt}
\setlist[enumerate]{leftmargin=0.3cm, itemsep=0.2pt, parsep=0pt, topsep=1pt}
\definecolor{revision}{rgb}{0.0, 0.0, 0.8}

\definecolor{improve}{rgb}{0.0, 0.5, 0.0}
\definecolor{degrade}{rgb}{0.8, 0.0, 0.0}
\definecolor{firstcolor}{rgb}{0.8, 0.0, 0.0}
\definecolor{secondcolor}{rgb}{0.0, 0.0, 0.8}
\definecolor{darkorange}{rgb}{0.85, 0.40, 0.00}
\definecolor{deepgreen}{rgb}{0.00, 0.55, 0.25}
\newcommand{\up}[1]{\textcolor{improve}{#1}}

\newcommand{\first}[1]{\textcolor{firstcolor}{\textbf{#1}}}
\newcommand{\second}[1]{\textcolor{secondcolor}{\underline{#1}}}

\newcommand\blfootnote[1]{%
  \begingroup
  \renewcommand\thefootnote{}\footnote{#1}%
  \addtocounter{footnote}{-1}%
  \endgroup
}

\title{FinSTaR: Towards Financial Reasoning \\ with Time Series Reasoning Models}

\author{
  Seunghan Lee \quad Jun Seo \quad Jaehoon Lee \quad Sungdong Yoo \quad Minjae Kim \\
  \textbf{Tae Yoon Lim} \quad \textbf{Dongwan Kang} \quad \textbf{Hwanil Choi} \quad \textbf{SoonYoung Lee} \quad \textbf{Wonbin Ahn} \\
  \\
  LG AI Research, Seoul, South Korea
}

\begin{document}

\maketitle

\blfootnote{Accepted to the EMNLP 2026 Industry Track.}

\vspace{-8pt}
\begin{abstract}
\vspace{-8pt}
Time series (TS) reasoning models (TSRMs) have shown promising capabilities in general domains, yet they consistently fail 
on financial domain, 
which exhibit unique characteristics. 
We propose a general $2 \times 2$ capability taxonomy for TSRMs by crossing 1) \textit{single-entity vs.\ multi-entity} analysis with 2) \textit{assessment} of the current state vs.\ \textit{prediction} of future behavior.
We instantiate this taxonomy in the financial domain---where the distinction between deterministic assessment and stochastic prediction is particularly critical---as ten financial reasoning tasks, forming the \textbf{FinTSR-Bench} benchmark based on S\&P stocks.
To this end, we propose \textbf{FinSTaR} (\textbf{Fin}ancial Time \textbf{S}eries \textbf{T}hinking \textbf{a}nd \textbf{R}easoning), 
trained on FinTSR-Bench with distinct chain-of-thought (CoT) strategies tailored to each category.
For \textit{assessment}, which is \textit{deterministic},
we employ \textit{Compute-in-CoT}, a programmatic CoT that enables models to derive answers directly from raw prices. 
For \textit{prediction}, which is inherently \textit{stochastic},
we adopt \textit{Scenario-Aware CoT}, which generates diverse scenarios before making a judgment, mirroring how financial analysts reason under uncertainty.
FinSTaR
achieves 78.9\% average accuracy 
on FinTSR-Bench,
substantially outperforming 
LLM and TSRM baselines.
Furthermore, we show that 
the 
four capability categories are complementary and mutually reinforcing through joint training.
Code is publicly available at: https://github.com/seunghan96/FinSTaR.
\end{abstract}
\vspace{-10pt}

\section{Introduction}
Time series (TS) reasoning models (TSRMs) have emerged as a promising direction for enabling LLMs to understand and reason over temporal data~\cite{timeomni1, thoth2026, patra2026}.
Unlike conventional forecasting, TSRMs produce interpretable reasoning chains that explain \textit{why} a particular conclusion follows from the data.
However, existing TSRMs are primarily developed 
on general domains and perform poorly 
on
financial TS~\cite{merrill2024towards}.
We find that existing 
models achieve accuracy barely above chance level on financial 
tasks, confirming that \textit{TSRMs lack financial reasoning capabilities}.

\begin{figure}[t]
\centering
\adjustbox{width=\linewidth}{%
\includegraphics[width=\linewidth]{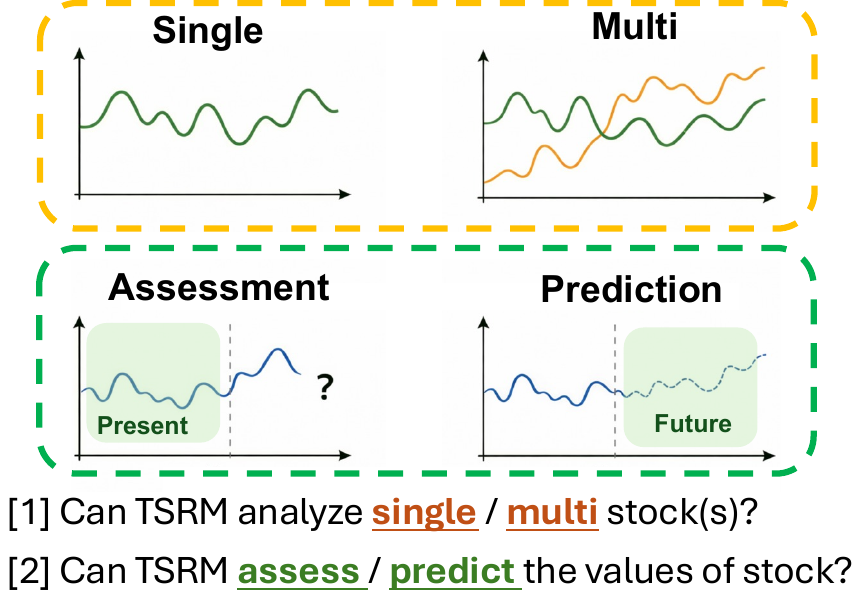}
}
\vspace{-15pt}
\caption{Capability taxonomy for TSRMs.}
\label{fig:taxonomy}
\vspace{-15pt}
\end{figure}

\textit{\textbf{Why do TSRMs fail on finance?}}
Financial TS exhibit characteristics that 
differ from general domains,
including volatility clustering, momentum and mean-reversion dynamics, and complex cross-asset dependencies.
More importantly, financial reasoning involves two distinct modes that existing TSRMs fail to distinguish:
\setlist[itemize]{leftmargin=0.56cm, itemsep=1.2pt, parsep=1.2pt, topsep=1pt}
\begin{itemize}
\item \textbf{[1] Assessment} (\textit{``What is the current state?''}): \textit{Deterministic} reasoning that can be resolved by computing observable quantities from price data (e.g., measuring drawdown severity).
\item \textbf{[2] Prediction} (\textit{``What will happen in the future?''}): \textit{Probabilistic} reasoning under uncertainty, where outcomes depend on unobservable 
factors 
(e.g., macroeconomic shifts), 
and even a perfect model cannot achieve 100\% accuracy.
\end{itemize}

\begin{figure*}[t]
\centering
\footnotesize

\begin{minipage}[t]{0.49\linewidth}
\centering
(\textbf{Single-Assessment}) \textbf{Volatility Regime}: \textit{Classify recent vs.\ overall volatility ratio. GT: (C) High.}
\vspace{2pt}

\noindent
\begin{minipage}[t]{0.46\linewidth}
\vspace{0pt}%
\centering
\adjustbox{width=\linewidth}{%
\includegraphics[width=\linewidth]{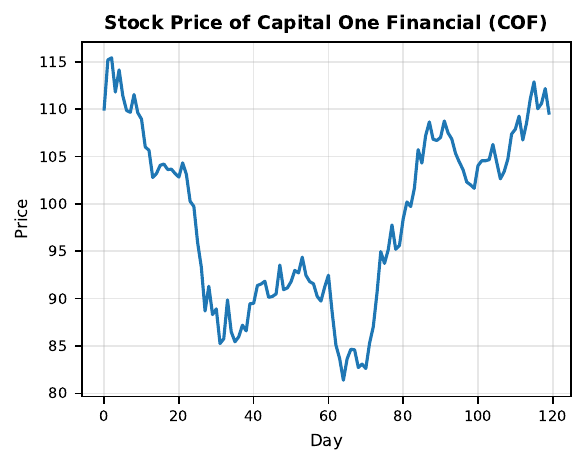}}
\end{minipage}%
\hfill
\begin{minipage}[t]{0.52\linewidth}
\vspace{0pt}%
\colorbox{red!8}{\parbox[t]{\dimexpr\linewidth-2\fboxsep}{%
\textbf{Qwen2.5-7B} \textcolor{red}{(A)} \ding{55}\\[2pt]
\fontsize{5.5}{6.6}\selectfont
\texttt{Calculate the overall volatility [...] Calculate the recent volatility [...] Compute the ratio [...] Let's perform these calculations.}\\[2pt]
\textcolor{gray}{\textit{$\triangleright$ No computation; directly outputs answer.}}
}}\\[2pt]
\colorbox{red!8}{\parbox[t]{\dimexpr\linewidth-2\fboxsep}{%
\textbf{TimeOmni-1} \textcolor{red}{(B)} \ding{55}\\[2pt]
\fontsize{5.5}{6.6}\selectfont
\texttt{Estimate recent SD $\approx$ 2--3. Estimate overall SD $\approx$ 2--3. Ratio $\approx$ 1.0 $\Rightarrow$ Normal.}\\[2pt]
\textcolor{gray}{\textit{$\triangleright$ Rough estimates; actual recent SD is 10.75\%.}}
}}
\end{minipage}

\vspace{3pt}

\noindent
\colorbox{green!10}{\parbox[t]{\dimexpr\linewidth-2\fboxsep}{%
\textbf{FinSTaR (Ours)} \textcolor{green!50!black}{(C)} \ding{51}\\[2pt]
\fontsize{5.5}{6.6}\selectfont
\texttt{Step 1: Compute daily returns.\\
Step 2: Overall volatility = 4.51\%.\\
Step 3: Recent 20-day vol = 10.75\%.\\
Step 4: Ratio = 10.75/4.51 = 2.38, recent vol well above long-term avg. $\Rightarrow$ (C) High volatility.}
}}
\end{minipage}%
\hfill
\begin{minipage}[t]{0.49\linewidth}
\centering
(\textbf{Single-Prediction}) \textbf{Event Response}: \textit{Predict mean-reversion or persistence after a shock. GT: (B) Persistence.}
\vspace{2pt}

\noindent
\begin{minipage}[t]{0.46\linewidth}
\vspace{0pt}%
\centering
\adjustbox{width=\linewidth}{%
\includegraphics[width=\linewidth]{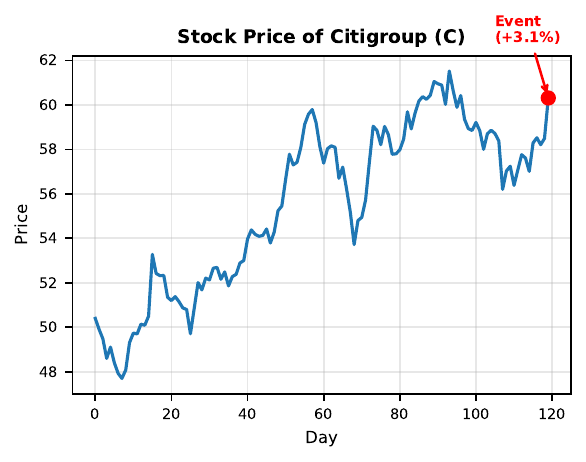}}
\end{minipage}%
\hfill
\begin{minipage}[t]{0.52\linewidth}
\vspace{0pt}%
\colorbox{red!8}{\parbox[t]{\dimexpr\linewidth-2\fboxsep}{%
\textbf{Qwen2.5-7B} \textcolor{red}{(A)} \ding{55}\\[2pt]
\fontsize{5.5}{6.6}\selectfont
\texttt{The overall pattern suggests the stock tends to revert to a mean after sharp moves. [...] the most likely outcome is that the price will move back toward pre-event levels.}
}}\\[2pt]
\colorbox{red!8}{\parbox[t]{\dimexpr\linewidth-2\fboxsep}{%
\textbf{TimeOmni-1} \textcolor{red}{(A)} \ding{55}\\[2pt]
\fontsize{5.5}{6.6}\selectfont
\texttt{The last value is 148.40, a significant jump from the previous day's low. This is an outlier [...] mean reversion is the most plausible scenario.}
}}
\end{minipage}

\vspace{3pt}

\noindent
\colorbox{green!10}{\parbox[t]{\dimexpr\linewidth-2\fboxsep}{%
\textbf{FinSTaR (Ours)} \textcolor{green!50!black}{(B)} \ding{51}\\[2pt]
\fontsize{5.5}{6.6}\selectfont
\texttt{Step 1: Event: Positive shock, z=4.52, +13.70\%.\\
Step 2: Pre-event trend: downward.\\
Step 3: Scenario analysis -- Base: Shock reflects genuine new info; Adverse: Mean-reversion forces kick in; Favorable: Additional news amplifies move.\\
Step 4: 2/3 support Persistence.\\
Step 5: Judgment $\Rightarrow$ (B).}
}}
\end{minipage}
\vspace{-5pt}
\caption{\textbf{Qualitative comparison.} 
Two simplified examples where 
baselines either
fail to compute precise quantities
or
default to heuristic reasoning
while FinSTaR's CoT produces grounded, step-by-step reasoning.
}
\label{fig:qualitative}
\vspace{-10pt}
\end{figure*}

\begin{table*}[t]
\centering
\caption{\textbf{Comparison with related works.} We compare across 1) domain, 2) input modality, 3) task, 4) CoT, and 5) uncertainty modeling.
FinSTaR is the
first \first{TSRM} designed for \first{financial} domain.
}
\label{tab:comparison}
\vspace{-3pt}
\adjustbox{width=\linewidth}{
\begin{tabular}{l|c|cc||c|cc|c|c|c}
\toprule
\multicolumn{1}{c|}{\multirow{2.5}{*}{\textbf{Work}}} & \multirow{2.5}{*}{\textbf{Venue}} & \multicolumn{2}{c||}{\textbf{Contribution}} & \multirow{2.5}{*}{\textbf{[1] Domain}} & \multicolumn{2}{c|}{\textbf{[2] Input}} & \multirow{2.5}{*}{\textbf{[3] Task}} & \multirow{2.5}{*}{\textbf{[4] CoT}} & \multirow{2.5}{*}{\textbf{[5] Uncertainty}} \\
\cmidrule(lr){3-4} \cmidrule(lr){6-7}
& & Benchmark & Model & & TS & Text & & & \\
\midrule
\midrule
S2TS-LLM~\cite{s2tsllm2025} & NeurIPS'25 & & \ding{51} & \multirow{5}{*}{General} & \ding{51} &  & Analysis &  &  \\
Thoth~\cite{thoth2026} & arXiv'26 & & \ding{51} &  & \ding{51} &  & Understanding &  &  \\
PATRA~\cite{patra2026} & ICML'26 & & \ding{51} &  & \ding{51} &  & QA & \ding{51} &  \\
SenTSR-Bench~\cite{sentsr2026} & AISTATS'26 & \ding{51} & \ding{51} &  & \ding{51} &  & Reasoning & \ding{51} &  \\
TimeOmni-1~\cite{timeomni1} & ICLR'26 & \ding{51} & \ding{51} &  & \ding{51} & \ding{51} & Reasoning QA & \ding{51} &  \\
\midrule
FinTSB~\cite{liu2025fintsb} & arXiv'25 & \ding{51} & & \multirow{5.5}{*}{\first{Finance}} & \ding{51} & & Forecasting &  &  \\
FinBen~\cite{finben2024} & NeurIPS'24 & \ding{51} & &  & & \ding{51} & NLP tasks &  &  \\
CausalStock~\cite{causalstock2024} & NeurIPS'24 & & \ding{51} &  & \ding{51} & \ding{51} & Forecasting &  &  \\
FinZero~\cite{wang2025finzero} & arXiv'25 & \ding{51} & \ding{51} &  &  & \ding{51} & Forecasting &  & \ding{51} \\
\cmidrule{1-4} \cmidrule{6-10}
\multicolumn{2}{c|}{\textbf{FinSTaR (Ours)}} & \textbf{\ding{51}} & \textbf{\ding{51}} &  & \textbf{\ding{51}} & \textbf{\ding{51}} & \first{Reasoning QA} & \textbf{\ding{51}}  & \textbf{\ding{51}} \\
\bottomrule
\end{tabular}
}
\vspace{-13pt}
\end{table*}

\begin{table}[t]
\centering
\caption{\textbf{Overview of FinTSR-Bench.} 
The 
ten financial reasoning tasks are organized along two axes.
}
\label{tab:taxonomy}
\adjustbox{width=\linewidth}{
\begin{tabular}{c|c|c}
\toprule
 & \textcolor{darkorange}{\textbf{Single}} & \textcolor{darkorange}{\textbf{Multi}} \\
\midrule
\multirow{3}{*}{\textcolor{deepgreen}{\textbf{Assessment}}} & Drawdown, & \multirow{3}{*}{Correlation} \\
 & Volatility, & \\
 & Trend & \\
\hline
\multirow{4}{*}{\textcolor{deepgreen}{\textbf{Prediction}}} & Event Response, & \multirow{4}{*}{\shortstack{Relative Performance,\\Pair Convergence}} \\
 & Support/Resistance, & \\
 & Drawdown Recovery, & \\
 & Volatility Forecast & \\
\bottomrule
\end{tabular}
}
\vspace{-13pt}
\end{table}

\begin{table*}[t]
\centering
\caption{
\textbf{Tasks of FinTSR-Bench.} 
Ten tasks span four capability categories across two axes. 
Matching symbols ($^\dagger$, $^\ddagger$, $^\S$) denote assessment--prediction pairs that share the same underlying financial skill.}
\label{tab:task_spec}
\adjustbox{width=0.8085\linewidth}{
\begin{tabular}{c|c|l|l}
\toprule
\multicolumn{2}{c|}{\textbf{Task category}} & \multicolumn{1}{c|}{\multirow{2.5}{*}{\textbf{Task (\# Classes)}}} & \multicolumn{1}{c}{\multirow{2.5}{*}{\textbf{Description}}} \\
\cmidrule(lr){1-2}
\textbf{Type} & \textbf{\# Stock} & & \\
\midrule
\multirow{4.5}{*}{\rotatebox{90}{\textbf{Assessment}}} & \multirow{3}{*}{Single} & Drawdown$^\dagger$ (4) & Classify peak-to-trough decline severity \\
 & & Volatility Regime$^\ddagger$ (3) & Classify recent vs.\ overall volatility ratio \\
 & & Trend Direction (5) & Classify 120-day cumulative return regime \\
\cmidrule{2-4}
& Multi & Correlation$^\S$ (3) & Classify return correlation of two stocks \\
\midrule
\multirow{6.5}{*}{\rotatebox{90}{\textbf{Prediction}}} & \multirow{4}{*}{Single} & Event Response (2) & Mean-reversion or persistence after extreme move \\
 & & Support/Resistance (2) & Breakout or bounce near key technical level \\
 & & Drawdown Recovery$^\dagger$ (2) & Recovery toward peak or further decline \\
 & & Volatility Forecast$^\ddagger$ (2) & Volatility increase or decrease \\
\cmidrule{2-4}
 & \multirow{2}{*}{Multi} & Relative Performance (2) & Which stock delivers higher returns \\
 & & Pair Convergence$^\S$ (2) & Price spread converges or diverges \\
\bottomrule
\end{tabular}
}
\end{table*}

However, no existing TSRM has been specifically designed for the financial domain or distinguishes between these two reasoning modes.
We propose a general $2 \times 2$ capability taxonomy for TSRMs along two 
axes: (data scope) \textit{single-entity vs.\ multi-entity} and (temporal scope) \textit{assessment vs.\ prediction},
as shown in Figure~\ref{fig:taxonomy}.
While this taxonomy is applicable to TSRMs in any domain, we instantiate it in the financial domain, where the distinction between deterministic assessment and stochastic prediction is particularly critical,
as ten financial reasoning tasks (Table~\ref{tab:taxonomy}).
To support training and evaluation, we construct \textbf{FinTSR-Bench} from 250 S\&P~500 stocks spanning 2010--2025, with ${\sim}$35K training samples and three out-of-distribution (OOD) test splits.

To this end, we propose \textbf{FinSTaR} (\textbf{Fin}ancial Time \textbf{S}eries \textbf{T}hinking \textbf{a}nd \textbf{R}easoning), the first TSRM designed for the financial domain,
as shown in Table~\ref{tab:comparison}.
Trained on FinTSR-Bench, FinSTaR employs different chain-of-thought (CoT) strategies for assessment and prediction.
Specifically, since assessment is deterministic and prediction is inherently stochastic, we use \textbf{Compute-in-CoT} for assessment tasks to derive answers from observable prices. For prediction, \textbf{Scenario-Aware CoT} generates multiple scenarios before making a probabilistic judgment.
various
scenarios before making a probabilistic judgment.
As illustrated in Figure~\ref{fig:qualitative}, existing models~\cite{qwen2025qwen25,timeomni1} 
either fail to compute precise quantities or rely on heuristic reasoning, 
while FinSTaR
produces correct answers through grounded, step-by-step reasoning.
Our main contributions are:
\begin{itemize}
\item We propose a \textbf{$2 \times 2$ capability taxonomy} for TSRMs by crossing \textit{assessment} vs.\ \textit{prediction} with \textit{single-} vs.\ \textit{multi-entity} analysis, and instantiate it as ten financial reasoning tasks.
\item We construct \textbf{FinTSR-Bench} from 250 S\&P~500 stocks (2010--2025) with 35K training samples
and three OOD test sets for
generalization evaluation, each containing 10K samples.
\item We propose \textbf{FinSTaR}, which is trained with 
CoT strategies tailored to each reasoning mode: Compute-in-CoT for assessment tasks and Scenario-Aware CoT for prediction tasks.
\item FinSTaR achieves 78.9\% overall accuracy, outperforming 15+ baselines, including LLMs, TSRMs, and TS forecasting methods. Through systematic ablation, we show that the four capability categories are complementary and mutually reinforcing under joint training.
\end{itemize}

\section{FinTSR-Bench}
\label{sec:benchmark}

We instantiate the proposed $2 \times 2$ taxonomy in the financial domain as \textbf{FinTSR-Bench}, both a training dataset and evaluation benchmark for financial TSRMs.
As shown in Table~\ref{tab:task_spec}, the proposed
ten tasks
cover all four capability categories.

\begin{table}[t]
\centering
\caption{\textbf{FinTSR-Bench.} 
Three test sets 
vary stock universe and time period to evaluate both \textcolor{red}{ID} and \textcolor{blue}{OOD}.}
\label{tab:splits}
\adjustbox{width=\linewidth}{
\begin{tabular}{cc|cc|cc}
\toprule
\multicolumn{2}{c} {\textbf{Split}} & \textbf{\# Stocks} & \textbf{Period} & \textbf{Universe} & \textbf{Period} \\
\midrule
\multicolumn{2}{c|}{Train} & \textcolor{red}{200} & \textcolor{red}{2010--2022} & --- & --- \\
\cmidrule(lr){1-6}
Test & A & \textcolor{red}{200} & \textcolor{blue}{2023--2025} & ID & OOD \\
Test & B & \textcolor{blue}{50} & \textcolor{red}{2010--2022} & OOD & ID \\
Test & C & \textcolor{blue}{50} & \textcolor{blue}{2023--2025} & OOD & OOD \\
\bottomrule
\end{tabular}
}
\vspace{-7pt}
\end{table}
We use the top 250 S\&P~500 stocks by market capitalization as of January 2024, with daily closing prices from 2010--2025.
The top 200 stocks serve as in-domain (ID) training data, 
with the remaining held out for OOD evaluation.
As shown in Table~\ref{tab:splits}, we split along both the stock universe and temporal axes, yielding three test sets that 
vary ID vs.\ OOD conditions.
Dataset statistics and task examples are provided in Appendix~\ref{app:dataset} and~\ref{app:tasks}.

\definecolor{cotboxA}{rgb}{0.94,0.97,1.00}
\definecolor{cotboxB}{rgb}{0.95,1.00,0.95}
\definecolor{cotruleA}{rgb}{0.55,0.68,0.85}
\definecolor{cotruleB}{rgb}{0.50,0.75,0.55}
\newsavebox{\cotpanelA}
\newsavebox{\cotpanelB}
\newlength{\cotpanelw}
\newlength{\cotinnerw}
\newlength{\cotpanelh}

\begin{figure*}[t]
\setlength{\cotpanelw}{0.485\textwidth}
\setlength{\cotinnerw}{\dimexpr\cotpanelw-2\fboxsep-2\fboxrule\relax}

\sbox{\cotpanelA}{%
\begin{minipage}{\cotinnerw}\centering
\vspace{2pt}
\includegraphics[width=\linewidth]{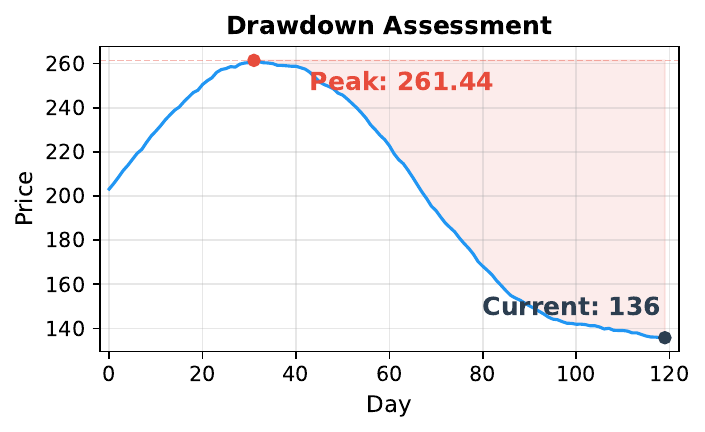}\\[5pt]
\begin{minipage}{\linewidth}
\fontsize{7.5pt}{9pt}\selectfont\ttfamily
Step 1: Peak = 261.44 (day 30).\\
Step 2: Current price = 136.00.\\
Step 3: Drawdown = (261.44$-$136.00)/261.44\\
\hspace*{1.0em}= 48.0\% $>$ 20\% $\Rightarrow$ (D) Severe Decline.
\end{minipage}
\vspace{2pt}
\end{minipage}}

\sbox{\cotpanelB}{%
\begin{minipage}{\cotinnerw}\centering
\vspace{2pt}
\includegraphics[width=\linewidth]{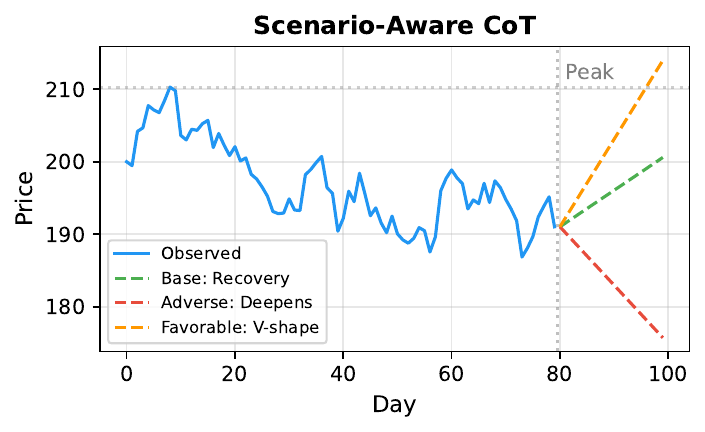}\\[5pt]
\begin{minipage}{\linewidth}
\fontsize{7.5pt}{9pt}\selectfont\ttfamily
Step 1: Peak = 210.15 (day 8), Current = 191.08.\\
Step 2: Drawdown = 9.1\%, momentum = $-$2.1\%.\\
Step 3: Scenario analysis:\\
\hspace*{0.8em}$\bullet$ Base: Stabilizes $\rightarrow$ recovery.\\
\hspace*{0.8em}$\bullet$ Adverse: Earnings revision $\rightarrow$ decline.\\
\hspace*{0.8em}$\bullet$ Favorable: Upgrade $\rightarrow$ V-recovery.\\
Step 4: Consider all scenarios $\Rightarrow$ recovery.\\
Step 5: Judgment: Recovery $\Rightarrow$ (A).
\end{minipage}
\vspace{2pt}
\end{minipage}}

\setlength{\cotpanelh}{\dimexpr\ht\cotpanelA+\dp\cotpanelA\relax}
\ifdim\dimexpr\ht\cotpanelB+\dp\cotpanelB\relax>\cotpanelh
  \setlength{\cotpanelh}{\dimexpr\ht\cotpanelB+\dp\cotpanelB\relax}%
\fi

\centering
\begin{subfigure}[t]{\cotpanelw}
\centering
\fcolorbox{cotruleA}{cotboxA}{%
  \begin{minipage}[c][\cotpanelh][t]{\cotinnerw}\usebox{\cotpanelA}\end{minipage}}
\caption{\textbf{Compute-in-CoT}: Assessment task.}
\label{fig:cot_example_assess}
\end{subfigure}
\hfill
\begin{subfigure}[t]{\cotpanelw}
\centering
\fcolorbox{cotruleB}{cotboxB}{%
  \begin{minipage}[c][\cotpanelh][t]{\cotinnerw}\usebox{\cotpanelB}\end{minipage}}
\caption{\textbf{Scenario-Aware CoT}: Prediction task.}
\label{fig:cot_example_scenario}
\end{subfigure}
\caption{
Two CoT strategies (simplified).
}
\label{fig:cot_examples}
\vspace{-10pt}
\end{figure*}

\section{Method: FinSTaR}
\label{sec:method}

We train FinSTaR via supervised fine-tuning (SFT) on FinTSR-Bench data with structurally different chain-of-thought supervision for 1) assessment tasks and 2) prediction tasks.

\textbf{Why two CoT strategies?}
Assessment and prediction tasks have fundamentally different epistemological properties.
Assessment tasks (e.g., ``What is the current drawdown?'') have answers that are \textit{fully determined} by the observable price data, and the model simply needs to compute the right quantity.
Prediction tasks (e.g., ``Will this drawdown recover?'') have answers that depend on \textit{future events} influenced by unobservable factors such as earnings surprises.
Even with a perfect model, prediction accuracy cannot reach 100\% because the necessary information is not contained in the price history alone.
This fundamental distinction between \textit{deterministic} and \textit{probabilistic} reasoning motivates our design of two complementary CoT strategies tailored to each reasoning mode: Compute-in-CoT for \textit{deterministic} assessment and Scenario-Aware CoT for \textit{stochastic} prediction.

\subsection{Compute-in-CoT (Assessment Tasks)}

For assessment tasks, the answer is deterministically computable from observable prices. We generate CoT chains programmatically by executing the same computation the model should learn. Each chain follows a three-phase structure (Figure~\ref{fig:cot_examples}a):
\begin{enumerate}[wide=0pt, labelsep=0.3em, itemsep=2pt, topsep=2pt]
\item[\textbf{Phase 1.}] 
\textbf{Extract}: Identify relevant quantities from input prices (e.g., peak price, recent returns).
\item[\textbf{Phase 2.}] 
\textbf{Compute}: Perform the calculation (e.g., drawdown percentage, volatility ratio).
\item[\textbf{Phase 3.}] 
\textbf{Classify}: Map the computed result to the answer using the thresholds in the question.
\end{enumerate}
This guarantees 100\% correctness and unlimited scalability, as models never reference unobservable information at inference time.

\begin{table*}[t]
\centering
\caption{\textbf{Main results of assessment/prediction tasks across Test A/B/C.} FinSTaR outperforms all baselines on 10 tasks (4 assessment, 6 prediction) under both in-domain (Test A) and out-of-distribution (Test B, C) settings. \first{Red bold} = best, \second{blue underline} = second best per split.}
\label{tab:main_results}
\vspace{-3pt}
\adjustbox{width=\linewidth}{
\scriptsize
\begin{tabular}{l|l|cccc|cccccc|c}
\toprule
\multirow{2.5}{*}{\textbf{Model}} & \multirow{2.5}{*}{\textbf{Split}} & \multicolumn{4}{c|}{\textit{Assessment}} & \multicolumn{6}{c|}{\textit{Prediction}} & \multirow{2.5}{*}{\textbf{Avg.}} \\
\cmidrule(lr){3-6} \cmidrule(lr){7-12}
 &  & \textbf{Draw.} & \textbf{Vol.} & \textbf{Trend} & \textbf{Corr.} & \textbf{Event} & \textbf{S/R} & \textbf{DDR} & \textbf{V.F.} & \textbf{R.P.} & \textbf{P.C.} & \\
\midrule
\rowcolor{gray!10}
\textit{Random} & --- & 25.0 & 33.3 & 20.0 & 33.3 & 50.0 & 50.0 & 50.0 & 50.0 & 50.0 & 50.0 & 41.2 \\
\midrule
\multicolumn{13}{l}{\textbf{\textit{Language Models}}} \\
\midrule
\multirow{3}{*}{Phi-3.5-mini} & A & 1.8 & 0.4 & 9.6 & 8.2 & 0.2 & 0.8 & 1.2 & 2.0 & 1.5 & 1.7 & 2.7 \\
 & B & 3.0 & 1.1 & 12.4 & 5.7 & 0.2 & 1.1 & 1.6 & 0.5 & 1.3 & 1.8 & 2.9 \\
 & C & 2.0 & 0.3 & 10.5 & 7.3 & 0.0 & 0.5 & 1.1 & 0.8 & 1.8 & 0.8 & 2.5 \\
\hline
\multirow{3}{*}{Mistral-7B} & A & 8.8 & 0.7 & 1.0 & 0.0 & 45.3 & 52.4 & 4.9 & 40.4 & 23.7 & 34.8 & 21.2 \\
 & B & 4.0 & 1.1 & 0.5 & 0.0 & 37.6 & 46.3 & 2.3 & 40.7 & 24.6 & 36.6 & 19.4 \\
 & C & 10.4 & 0.3 & 1.4 & 0.0 & 46.1 & 51.0 & 1.8 & 40.2 & 28.3 & 36.9 & 21.6 \\
\hline
\multirow{3}{*}{Gemma-2-9B} & A & 79.6 & 0.0 & 59.6 & 29.4 & 45.4 & 29.4 & 22.3 & 17.2 & 0.0 & 0.1 & 28.3 \\
 & B & 78.9 & 0.0 & 58.4 & 30.9 & 47.4 & 36.9 & 28.0 & 16.4 & 0.1 & 0.0 & 29.7 \\
 & C & 77.3 & 0.0 & 50.7 & 30.8 & 45.4 & 30.6 & 23.2 & 20.1 & 0.0 & 0.2 & 27.8 \\
\hline
\multirow{3}{*}{Llama-3.1-8B} & A & 57.2 & 30.3 & 75.7 & 1.4 & 47.3 & 50.2 & 47.3 & 48.0 & 29.9 & 33.6 & 42.1 \\
 & B & 59.9 & 30.5 & 73.4 & 2.8 & 46.5 & 48.3 & 45.2 & 48.8 & 26.9 & 31.7 & 41.4 \\
 & C & 57.8 & 30.2 & 76.4 & 2.0 & 46.4 & 50.2 & 46.3 & 45.9 & 31.5 & 35.1 & 42.2 \\
\hline
\multirow{3}{*}{Qwen2.5-7B} & A & 72.9 & 37.1 & 76.9 & 26.9 & 52.2 & 51.2 & 54.1 & 44.9 & 49.2 & 50.4 & 51.6 \\
 & B & 71.5 & 37.0 & 73.3 & 30.3 & \second{53.6} & 48.6 & 58.8 & 46.7 & \first{50.2} & 46.6 & 51.7 \\
 & C & 73.7 & 36.6 & 77.5 & 25.8 & \second{51.4} & 50.4 & 55.0 & 44.9 & \second{51.3} & 47.4 & 51.4 \\
\midrule
\multicolumn{13}{l}{\textbf{\textit{TS Language Models}}} \\
\midrule
\multirow{3}{*}{TimeMQA-Qwen} & A & 5.9 & 1.6 & 3.3 & 0.4 & 1.2 & 2.1 & 1.6 & 1.7 & 1.4 & 2.0 & 2.1 \\
 & B & 5.5 & 0.8 & 2.9 & 0.6 & 0.7 & 1.7 & 0.8 & 2.1 & 0.7 & 0.8 & 1.7 \\
 & C & 5.1 & 1.0 & 3.3 & 0.3 & 1.2 & 2.0 & 1.1 & 1.7 & 1.4 & 1.4 & 1.8 \\
\hline
\multirow{3}{*}{TimeMQA-Mistral} & A & 5.5 & 1.6 & 4.7 & 1.3 & 2.5 & 11.5 & 10.8 & 11.0 & 9.9 & 5.4 & 6.4 \\
 & B & 3.5 & 2.8 & 5.4 & 0.9 & 3.0 & 10.1 & 10.9 & 10.8 & 7.0 & 3.9 & 5.8 \\
 & C & 4.7 & 2.6 & 5.5 & 0.8 & 3.9 & 12.7 & 13.3 & 11.1 & 7.9 & 4.9 & 6.7 \\
\hline
\multirow{3}{*}{TimeMQA-Llama} & A & 6.9 & 7.5 & 6.6 & 3.3 & 16.0 & 20.4 & 17.6 & 7.2 & 8.7 & 9.7 & 10.4 \\
 & B & 9.6 & 8.6 & 7.9 & 6.1 & 15.6 & 20.1 & 16.8 & 7.6 & 7.9 & 11.9 & 11.2 \\
 & C & 10.3 & 6.3 & 5.8 & 6.9 & 16.6 & 18.6 & 15.7 & 8.7 & 9.7 & 10.2 & 10.9 \\
\midrule
\multicolumn{13}{l}{\textbf{\textit{TS Reasoning Models}}} \\
\midrule
\multirow{3}{*}{TimeOmni-1-7B} & A & 75.7 & 35.6 & 50.5 & 31.5 & 49.1 & \second{54.1} & 59.3 & 58.2 & \second{50.7} & 48.4 & 51.3 \\
 & B & 75.0 & 35.6 & 53.1 & 31.9 & 51.7 & 50.7 & 57.6 & 55.8 & 48.3 & 49.0 & 50.9 \\
 & C & 73.0 & 33.7 & 49.7 & 34.5 & 48.6 & 52.8 & 53.7 & 54.1 & 50.8 & 48.4 & 49.9 \\
\midrule
\multicolumn{13}{l}{\textbf{\textit{SFT Baselines (trained on FinTSR-Bench)}}} \\
\midrule
\multirow{3}{*}{Qwen2.5-7B (w/o CoT)} & A & 77.5 & 40.4 & 75.2 & 34.7 & \second{53.0} & 53.7 & \second{67.1} & \second{60.6} & 50.7 & 51.3 & 56.4 \\
 & B & 79.6 & 41.6 & 75.9 & \second{36.2} & 50.0 & \second{51.1} & \second{68.7} & \second{58.2} & 46.9 & 48.1 & 55.6 \\
 & C & 77.9 & 39.1 & 72.8 & 30.3 & 50.4 & \second{55.2} & \second{61.5} & \second{59.0} & 50.3 & 49.4 & 54.6 \\
\hline
\multirow{3}{*}{Qwen2.5-7B (w/ CoT)} & A & \second{96.4} & \second{46.6} & \second{92.3} & \second{35.4} & 47.6 & 51.1 & 57.2 & 48.7 & 47.3 & \second{55.0} & \second{57.8} \\
 & B & \second{97.1} & \second{52.6} & \second{93.5} & 36.0 & 50.7 & 49.4 & 59.1 & 46.8 & 49.9 & \second{51.6} & \second{58.7} \\
 & C & \second{96.5} & \second{45.7} & \second{91.6} & \second{38.5} & 48.7 & 51.0 & 56.4 & 49.8 & 45.8 & \second{53.6} & \second{57.8} \\
\midrule
\multicolumn{13}{l}{\textbf{\textit{Financial TSRMs (Ours)}}} \\
\midrule
\multirow{3}{*}{\textbf{FinSTaR}} & \cellcolor{yellow!15}  A & \cellcolor{yellow!15} \first{99.3} & \cellcolor{yellow!15} \first{93.5} & \cellcolor{yellow!15} \first{99.3} & \cellcolor{yellow!15} \first{88.1} & \cellcolor{yellow!15} \first{57.2} & \cellcolor{yellow!15} \first{80.5} & \cellcolor{yellow!15} \first{74.4} & \cellcolor{yellow!15} \first{79.5} & \cellcolor{yellow!15} \first{51.5} & \cellcolor{yellow!15} \first{65.8} & \cellcolor{yellow!15}\first{78.9} \\
& \cellcolor{yellow!15}  B & \cellcolor{yellow!15}  \first{99.6} & \cellcolor{yellow!15}  \first{92.2} & \cellcolor{yellow!15}  \first{99.5} & \cellcolor{yellow!15}  \first{89.0} & \cellcolor{yellow!15}  \first{55.6} & \cellcolor{yellow!15}  \first{80.0} & \cellcolor{yellow!15}  \first{75.0} & \cellcolor{yellow!15}  \first{77.3} & \cellcolor{yellow!15}  \second{50.1} & \cellcolor{yellow!15} \first{64.3} & \cellcolor{yellow!15}  \first{78.3} \\
& \cellcolor{yellow!15}   C & \cellcolor{yellow!15}  \first{99.0} & \cellcolor{yellow!15}  \first{92.7} & \cellcolor{yellow!15}  \first{99.4} & \cellcolor{yellow!15}  \first{87.9} & \cellcolor{yellow!15}  \first{53.5} & \cellcolor{yellow!15}  \first{79.5} & \cellcolor{yellow!15}  \first{72.9} & \cellcolor{yellow!15}  \first{78.8} & \cellcolor{yellow!15}  \first{53.0} & \cellcolor{yellow!15}  \first{64.6} & \cellcolor{yellow!15}  \first{78.1} \\
\bottomrule
\end{tabular}
}
\vspace{-6pt}
\end{table*}

\subsection{Scenario-Aware CoT (Prediction Tasks)}
\label{sec:scenario_cot}

For prediction tasks, the answer depends on future outcomes influenced by unobservable factors, 
where a \textit{deterministic CoT can be problematic as it may lead the model to be overconfident} about 
uncertain outcomes.
Instead, we propose Scenario-Aware CoT, which teaches the model to consider \textit{multiple scenarios} before making a 
\textit{probabilistic}
judgment.
Each 
CoT follows a five-phase structure:
\begin{enumerate}[wide=0pt, labelsep=0.3em, itemsep=2pt, topsep=2pt]
\item[\textbf{Phase 1.}] \textbf{Extract}: (Same as assessment) Identify relevant quantities from the data.
\item[\textbf{Phase 2.}] \textbf{Compute}: Calculate price-based features (e.g., drawdown depth, recent momentum).
\item[\textbf{Phase 3.}] \textbf{Scenario}: Generate three scenarios:
  \begin{itemize}[leftmargin=1.5em, labelsep=0.3em]
  \item \textit{Base case}: The most probable outcome given current price patterns and no external events.
  \item \textit{Adverse scenario}: An external shock that could reverse the expected outcome.
  \item \textit{Favorable scenario}: A positive catalyst that strengthens the expected outcome.
  \end{itemize}
\item[\textbf{Phase 4.}] \textbf{Assessment}: Consider all scenarios to determine which outcome is most supported.
\item[\textbf{Phase 5.}] \textbf{Judgment}: Select the outcome most supported by the scenario analysis and input data.
\end{enumerate}

\textbf{Scenario templates.}
For each (task, answer) pair, we define domain-specific scenario templates grounded in financial knowledge.
For example, \textit{Volatility Forecast} scenarios reference volatility clustering (base), event-driven spikes (adverse), and market calming (favorable).
During training, the model learns these patterns; at inference, it generates scenario reasoning adapted to the specific input, not template copies (see Section~\ref{sec:analysis}).
Scenario templates 
are provided in Appendix~\ref{app:scenario_templates}.

\section{Experiments}
\label{sec:experiments}

\textbf{Baseline methods.}
We compare FinSTaR against four categories of \textit{reasoning/language model} baselines:
\textbf{1)~Language Models}: Qwen2.5-7B-Instruct \cite{qwen2025qwen25}, Llama-3.1-8B-Instruct \cite{llama31}, Mistral-7B-Instruct \cite{mistral7b}, Gemma-2-9B-it \cite{gemma2}, Phi-3.5-mini \cite{phi35}.
\textbf{2)~TS Language Models}: TimeMQA \cite{timemqa2025} with Qwen-2.5-7B/Mistral-7B/Llama-3-8B backbones.
\textbf{3)~TS Reasoning Models}: TimeOmni-1-7B \cite{timeomni1}\footnote{As of this work, TimeOmni-1 is the only TSRM that publicly releases model weights at the 7B scale. The remaining TSRMs either lack public code/weights or are available only at larger scales (e.g., Thoth, 30B).}, which serves as FinSTaR's base model before fine-tuning.
\textbf{4)~SFT Baselines}: To ensure fair comparison, we also fine-tune Qwen2.5-7B on FinTSR-Bench with both answer-only (w/o CoT) and CoT supervision, providing matched baselines that isolate the contribution of TS reasoning pre-training and our CoT design.
We also compare against 
TS forecasting models
on prediction tasks in Appendix~\ref{app:forecast_protocol}.
All LLM baselines in 1)--3) are evaluated zero-shot, following the 
standard protocol \cite{timeomni1, merrill2024towards}. 

\begin{table*}[t]
\centering
\caption{\textbf{Ablation study w/ and w/o CoT}. CoT consistently improves accuracy across all 10 tasks.
}
\label{tab:ao_vs_cot}
\vspace{-3pt}
\adjustbox{width=0.854\linewidth}{
\begin{tabular}{l|cccc|cccccc|c}
\toprule
 & \multicolumn{4}{c|}{\textit{Assessment}} & \multicolumn{6}{c|}{\textit{Prediction}} & \multirow{2.5}{*}{\textbf{Avg.}} \\
\cmidrule(lr){2-5} \cmidrule(lr){6-11}
 & \textbf{Draw.} & \textbf{Vol.} & \textbf{Trend} & \textbf{Corr.} & \textbf{Event} & \textbf{S/R} & \textbf{DDR} & \textbf{V.F.} & \textbf{R.P.} & \textbf{P.C.} & \\
\midrule
w/o CoT & 78.3 & 37.2 & 68.3 & 32.1 & 49.0 & 56.3 & 66.8 & 69.1 & 46.9 & 54.6 & 55.9 \\
\rowcolor{yellow!15} w/ CoT &  \first{99.3} &  \first{93.5} &  \first{99.3} &  \first{88.1} &  \first{57.2} &  \first{80.5} &  \first{74.4} &  \first{79.5} &  \first{51.5} &  \first{65.8} &  \first{78.9} \\
\midrule
$\Delta$ & \up{+21.0} & \up{+56.3} & \up{+31.0} & \up{+56.0} & \up{+8.2} & \up{+24.2} & \up{+7.6} & \up{+10.4} & \up{+4.6} & \up{+11.2} & \up{+23.0} \\
\bottomrule
\end{tabular}
}
\end{table*}

\begin{table*}[t]
\centering
\caption{\textbf{Effect of Scenario-Aware CoT.} Scenario-Aware CoT improves both assessment and prediction over (deterministic) standard CoT on Test A.
}
\label{tab:e_vs_i}
\vspace{-3pt}
\adjustbox{width=0.930\linewidth}{
\begin{tabular}{l|cccc|c|cccccc|c}
\toprule
\multirow{2.5}{*}{\textbf{CoT}} & \multicolumn{5}{c|}{\textit{Assessment}} & \multicolumn{7}{c}{\textit{Prediction}} \\
\cmidrule(lr){2-6} \cmidrule(lr){7-13}
 & \textbf{Draw.} & \textbf{Vol.} & \textbf{Trend} & \textbf{Corr.} & \textbf{Avg.} & \textbf{Event} & \textbf{S/R} & \textbf{DDR} & \textbf{V.F.} & \textbf{R.P.} & \textbf{P.C.} & \textbf{Avg.} \\
\midrule
Standard & 99.1 & 92.7 & \first{99.3} & 86.9 & 94.5 & 51.0 & 80.0 & 74.3 & 69.3 & 51.1 & 65.4 & 65.2 \\
\rowcolor{yellow!15} Scenario-Aware & \first{99.3} & \first{93.5} & \first{99.3} & \first{88.1} & \first{95.0} & \first{57.2} & \first{80.5} & \first{74.4} & \first{79.5} & \first{51.5} & \first{65.8} & \first{68.2} \\
\midrule
$\Delta$ & \up{+0.2} & \up{+0.8} & +0.0 & \up{+1.2} & \up{+0.5} & \up{+6.2} & \up{+0.5} & \up{+0.1} & \up{+10.2} & \up{+0.4} & \up{+0.4} & \up{+3.0} \\
\bottomrule
\end{tabular}
}
\vspace{-6pt}
\end{table*}

\textbf{Training setup.} Backbones and LoRA 
hyperparameters are described in Appendix~\ref{app:training_setup}. Unless otherwise stated, FinSTaR refers to TimeOmni-1-7B fine-tuned with our CoT strategies.

\subsection{Main Results}
Table~\ref{tab:main_results} presents the comparison across all three test splits.
FinSTaR achieves \textbf{78.9\%} overall accuracy on Test~A, outperforming the best zero-shot LLM (Qwen2.5-7B, 51.6\%) by \textbf{+27.3\%}, the best SFT baseline (Qwen2.5-7B w/ CoT, 57.8\%) by \textbf{+21.1\%}, and the best TSRM (TimeOmni-1-7B, 51.3\%) by \textbf{+27.6\%}, with consistent performance across Test~B (78.3\%) and Test~C (78.1\%).

\textbf{Key observations.}
1)~Assessment tasks reach near-perfect accuracy ($>$93\%), as Compute-in-CoT makes the computation explicit and verifiable.
2)~On prediction tasks, FinSTaR consistently outperforms all baselines, with the largest gains on Volatility Forecast (+21.3\% over the best zero-shot baseline) and Support/Resistance (+26.4\%).
3)~TimeMQA models, despite being fine-tuned on TS QA, perform \textit{worse} than general LLMs, indicating that non-financial TS training \textit{does not transfer} to financial reasoning.
4)~Many models exhibit low success rates 
as shown in Table~\ref{tab:sr}, failing to produce valid output format, consistent with the findings of TimeOmni-1 \cite{timeomni1}.

\begin{table*}[t]
\centering
\begin{minipage}[t]{0.67\linewidth}
\centering
\caption{\textbf{Leave-One-Category-Out ablation study.} Each row removes one capability category from training and evaluates on all tasks. Together with Table~\ref{tab:solo_joint}, the results reveal the dependencies among task categories and demonstrate the \textit{effectiveness of joint training across all four capabilities}.}
\label{tab:loco_full}
\vspace{-4pt}
\adjustbox{width=\linewidth}{
\begin{tabular}{l|cccc|cccccc|c}
\toprule
\multirow{2}{*}{ } & \multicolumn{4}{c|}{\textit{Assessment}} & \multicolumn{6}{c|}{\textit{Prediction}} & \multirow{2.5}{*}{\textbf{Avg.}} \\
\cmidrule(lr){2-5} \cmidrule(lr){6-11}
 & \textbf{Draw.} & \textbf{Vol.} & \textbf{Trend} & \textbf{Corr.} & \textbf{Event} & \textbf{S/R} & \textbf{DDR} & \textbf{V.F.} & \textbf{R.P.} & \textbf{P.C.} & \\
\midrule
\rowcolor{yellow!15} Full & \first{99.3} & \first{93.5} & \second{99.3} & \first{88.1} & \first{57.2} & \first{80.5} & 74.4 & \first{79.5} & \first{51.5} & \first{65.8} & \first{78.9} \\
\midrule
w/o AS & 43.2 & 32.2 & 29.2 & \second{88.0} & 56.8 & 79.8 & \first{75.3} & 78.8 & \first{51.5} & \second{65.4} & 60.0 \\
w/o AM & 98.8 & 92.3 & 99.2 & 33.4 & 53.9 & \second{80.4} & \second{75.2} & \second{78.9} & \first{51.5} & 64.4 & 72.8 \\
w/o PS & \second{99.2} & 90.6 & \first{99.4} & \second{88.0} & 47.7 & 48.1 & 47.4 & 55.1 & \first{51.5} & 64.1 & 69.1 \\
w/o PM & 98.9 & \second{92.3} & 67.5 & \first{88.1} & \second{57.1} & \second{80.4} & 74.8 & 76.6 & \second{47.8} & 49.2 & \second{73.3} \\
\bottomrule
\end{tabular}
}
\end{minipage}
\hfill
\begin{minipage}[t]{0.292\linewidth}
\centering
\captionof{table}{Solo vs.\ joint training.}
\label{tab:solo_joint}
\vspace{-4pt}
\adjustbox{width=\linewidth}{
\footnotesize
\begin{tabular}{c|l|cc|c}
\toprule
\textbf{Cat.} & \textbf{Task} & \textbf{Solo} & \textbf{Joint} & \textbf{$\Delta$} \\
\midrule
\multirow{3}{*}{AS} & Draw. & 98.8 & \cellcolor{yellow!15} \first{99.3} & \up{+0.5} \\
 & Vol. & 91.1 & \cellcolor{yellow!15} \first{93.5} & \up{+2.4} \\
 & Trend & 99.1 & \cellcolor{yellow!15} \first{99.3} & \up{+0.2} \\
\midrule
AM & Corr. & 85.7 & \cellcolor{yellow!15} \first{88.1} & \up{+2.4} \\
\midrule
\multirow{4}{*}{PS} & Event & 55.9 & \cellcolor{yellow!15} \first{57.2} & \up{+1.3} \\
 & S/R & 79.5 & \cellcolor{yellow!15} \first{80.5} & \up{+1.0} \\
 & DDR & \first{74.4} & \cellcolor{yellow!15} \first{74.4} & +0.0 \\
 & V.F. & 78.4 & \cellcolor{yellow!15} \first{79.5} & \up{+1.1} \\
\midrule
\multirow{2}{*}{PM} & R.P. &\first{51.5} & \cellcolor{yellow!15} \first{51.5} & +0.0 \\
 & P.C. & 64.5 & \cellcolor{yellow!15} \first{65.8} & \up{+1.3} \\
\bottomrule
\end{tabular}
}
\end{minipage}

\vspace{8pt}

\centering
\captionof{table}{Success rate on Test~A.}
\label{tab:sr}
\vspace{3pt}
\adjustbox{width=0.927\linewidth}{
\scriptsize
\begin{tabular}{l|ccc|ccc|c|c}
\toprule
& \multicolumn{3}{c|}{\textit{Language Models}} & \multicolumn{3}{c|}{\textit{TS Language Models}} & \textit{TSRM} & \textit{Ours} \\
 \cmidrule(lr){2-4} \cmidrule(lr){5-7}  \cmidrule(lr){8-8}  \cmidrule(lr){9-9}
 & Gemma & Llama & Qwen & TMQA-Q & TMQA-M & TMQA-L & TimeOmni & \textbf{FinSTaR} \\
\midrule
SR   & 51.8 & 80.3 & 99.9 & 4.8 & 16.3 & 25.4 & 98.8 & \cellcolor{yellow!15}\first{100.0} \\
Acc. & 28.3 & 42.1 & 51.6 & 2.1 & 6.4  & 10.4 & 51.3 & \cellcolor{yellow!15}\first{78.9} \\
\bottomrule
\end{tabular}
}
\vspace{5pt}
\end{table*}

\subsection{Analysis}
\label{sec:analysis}

\textbf{Ablation: Value of Reasoning.}
Table~\ref{tab:ao_vs_cot} compares w/o CoT (answer-only) vs.\ w/ CoT (FinSTaR) on the TimeOmni-1-7B. 
CoT provides the largest gains on assessment tasks where computation becomes explicit: Correlation (+56.0), Volatility (+56.3), and Trend (+31.0).
On prediction tasks, gains are moderate but consistent: Support/Resistance (+24.2), Volatility Forecast (+10.4), and Pair Convergence (+11.2).
Relative Performance shows the smallest gain (+4.6), reflecting the inherent difficulty of stock return prediction.

\begin{figure}[t]
\centering
\begin{subfigure}[t]{0.535\linewidth}
\adjustbox{width=\linewidth}{%
\includegraphics[width=\linewidth]{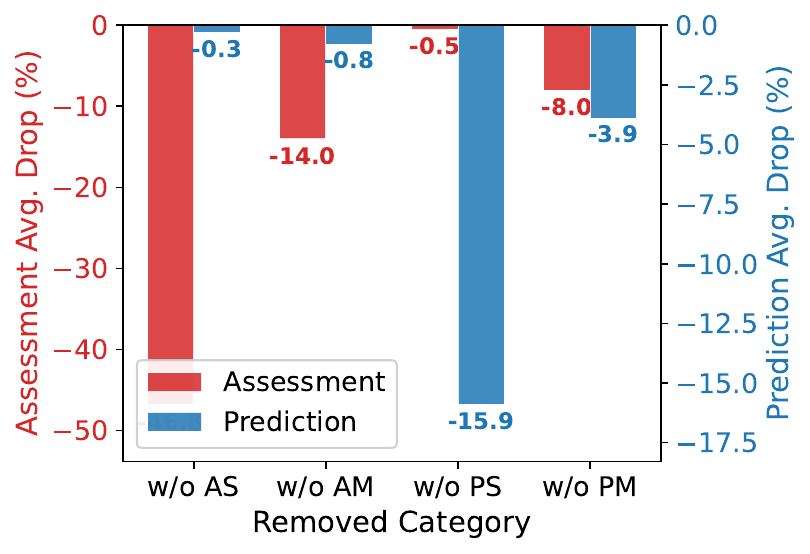}
}
\caption{LOCO ablation.}
\end{subfigure}%
\hfill
\begin{subfigure}[t]{0.45\linewidth}
\adjustbox{width=\linewidth}{%
\includegraphics[width=\linewidth]{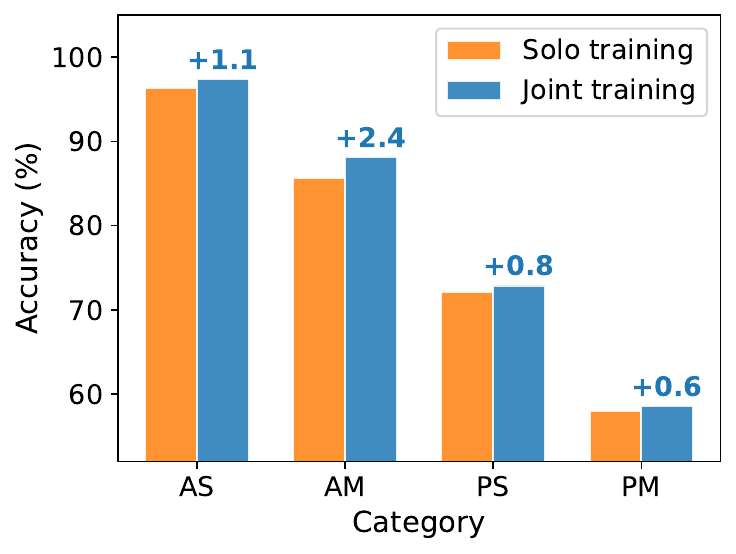}
}
\caption{Solo vs.\ joint training.}
\end{subfigure}
\vspace{-4pt}
\caption{\textbf{Taxonomy dependency analysis.} (a) Each category is essential to its own tasks: removing AS causes the largest assessment drop 
and removing PS causes the largest prediction drop. 
(b) Joint training across all four categories outperforms solo training, indicating that the categories are 
mutually reinforcing.}
\label{fig:loco_analysis}
\vspace{-8pt}
\end{figure}

\textbf{Scenario-Aware CoT Effect.}
To isolate the effect of Scenario-Aware CoT, we compare two CoT configurations for prediction tasks: (deterministic) standard CoT vs. (probabilistic) Scenario-Aware CoT,
with standard CoT fixed for assessment tasks in both.
As shown in Table~\ref{tab:e_vs_i}, assessment accuracy is nearly identical between the two.
On prediction tasks, Scenario-Aware CoT achieves +3\% average improvement, with the largest gains on Volatility Forecast (+10.2) and Event Response (+6.2).
These results confirm that \textit{scenario reasoning consistently improves performance on uncertain outcomes}.

\textbf{Capability dependencies.}
We study how the \textit{four capability categories} --- \textbf{AS} (Assessment-Single), \textbf{AM} (Assessment-Multi), \textbf{PS} (Prediction-Single), and \textbf{PM} (Prediction-Multi) --- interact during training via two complementary analyses on TimeOmni-1-7B on Test A.
First, we conduct a \textit{Leave-One-Category-Out} (LOCO) ablation: remove one category's tasks from training, retrain, and evaluate all ten tasks.
As shown in Figure~\ref{fig:loco_analysis}(a), each category is \textit{essential to its own tasks} --- removing AS causes the largest assessment drop and removing PS causes the largest prediction drop --- while removals do not catastrophically harm the other axis.
Second, Figure~\ref{fig:loco_analysis}(b) compares solo training (each category alone) with joint training (all four together): joint training \textit{consistently outperforms} solo training across all categories.
Together, these results show that the four categories are \textit{complementary and mutually reinforcing through joint training}, rather than redundant or interfering.
Full results are shown in Tables~\ref{tab:loco_full} and~\ref{tab:solo_joint}.

\textbf{Success rate.}
Table~\ref{tab:sr} reports the success rate (SR), the percentage of outputs containing a valid answer tag.
Most baselines fail to follow the required format: TimeMQA variants and Phi-3.5-mini achieve SR $<$ 25\%, rendering their accuracy unreliable.
In contrast, FinSTaR achieves 100\% SR, confirming that structured CoT supervision also teaches reliable output formatting.

\textbf{Prediction tasks: TSRM vs.\ TS forecasting models.} FinSTaR is further compared against statistical and deep-learning forecasting baselines on prediction tasks via a \textit{forecast-then-classify} pipeline; FinSTaR outperforms all such baselines (see Appendix~\ref{app:forecast_protocol}, Table~\ref{tab:forecast_vs_reasoning}).

\textbf{Backbone sensitivity.} Scenario-Aware CoT improves prediction only when the backbone already possesses TS reasoning ability: it boosts TimeOmni-1 but \textit{hurts} the general-purpose Qwen2.5-7B. This indicates that \textit{general TS reasoning ability is a prerequisite for financial TS reasoning}---without it, scenario-based reasoning generates plausible but ungrounded narratives rather than useful signal (see Appendix~\ref{app:backbone_sensitivity}, Table~\ref{tab:backbone}).

\textbf{Effect of training set size.} A data-efficiency study shows that FinSTaR reaches 71.7\% overall accuracy with only 10\% of the training data, confirming that our CoT strategies are sample-efficient (see Appendix~\ref{app:data_efficiency}, Table~\ref{tab:data_eff_full}).


\textbf{CoT quality verification.}
We verify that FinSTaR generates meaningful reasoning rather than template copies along four dimensions, with details provided in Appendix~\ref{app:tasks}:
\setlist[itemize]{leftmargin=0.3cm, itemsep=0.2pt, parsep=0pt, topsep=1pt}
\begin{itemize}
\item \textbf{Format compliance.} Over 99\% of outputs contain valid \texttt{<think>...<answer>} structure.
\item \textbf{Arithmetic faithfulness.} Since each assessment label is uniquely determined by a numerical computation, the near-perfect accuracy 
implies faithful 
reasoning. 
Manual inspection confirms that computed values align with ground truth.
\item \textbf{Scenario diversity.} Across 100 sampled predictions, 85+ scenario descriptions per task confirm that the model \textit{generates varied reasoning, not memorized templates}.
While scenario templates are defined per (task, answer) pair during training, this does not constitute label leakage: the model generates all three scenarios \textit{before} selecting an answer, and the template serves as a reasoning scaffold rather than an answer key.
\end{itemize}



\section{Conclusion}
We introduce FinSTaR, 
\textit{the first Financial TSRM},
which separates financial TS reasoning into assessment and prediction.
Our general $2 \times 2$ capability taxonomy, instantiated as ten financial reasoning tasks, reveals that the four categories are complementary and mutually reinforcing through joint training, and our scenario-based reasoning consistently outperforms standard CoT on prediction tasks.
FinSTaR achieves 78.9\% overall accuracy, outperforming over 15 baselines.
%

\section*{Limitations}
FinSTaR has several limitations.
First, FinTSR-Bench is built from the top 250 S\&P~500 stocks over 2010--2025, so our findings may not transfer to small-cap, international, or non-equity assets (e.g., FX, commodities, crypto) with distinct dynamics.
Second, we cast financial reasoning as ten closed-form classification tasks, leaving open-ended problems such as numerical forecasting, portfolio optimization, and free-form analysis out of scope.
Third, Scenario-Aware CoT is bounded by human-designed templates, and we evaluate only at the 7B scale (the sole publicly released TSRM at this size), so how the gains scale to larger models or template-free supervision remains open.
Finally, all results are offline and exclude transaction costs and market impact; FinSTaR is a research artifact and decision-support tool, not an autonomous trading system or investment advice.

\section*{Ethical Considerations}
This work advances time series reasoning by introducing a general capability taxonomy and instantiating it in the financial domain through FinTSR-Bench and FinSTaR. The positive societal impact includes enabling more interpretable and transparent financial analysis---our structured chain-of-thought framework produces grounded, step-by-step reasoning that is auditable by analysts and regulators, in contrast to opaque black-box forecasters. By distinguishing deterministic assessment from stochastic prediction, our approach also promotes epistemically honest modeling: FinSTaR explicitly acknowledges uncertainty in prediction tasks through scenario-based reasoning rather than producing overconfident point predictions. However, we caution against unintended misuse: the model is trained on historical S\&P~500 data and should not be deployed as an autonomous trading system or investment advisor, as prediction tasks are inherently uncertain (reflecting market efficiency) and outputs may reflect biases in the training distribution. We recommend using FinSTaR as a decision-support tool that augments, rather than replaces, human financial judgment, and we emphasize that model outputs should be validated before any consequential financial decisions.

\bibliography{references}

\clearpage
\appendix
\renewcommand{\thefigure}{\thesection.\arabic{figure}}
\renewcommand{\thetable}{\thesection.\arabic{table}}
\counterwithin{figure}{section}
\counterwithin{table}{section}
\section{Extended Related Work}
\label{app:related_work}

\textbf{Time series reasoning models (TSRMs).}
Recent work extends LLMs beyond forecasting toward TS reasoning.
TimeOmni-1~\cite{timeomni1} introduces TSR-Suite with human-annotated CoT but focuses on non-financial domains.
Thoth~\cite{thoth2026} proposes mid-training for TS understanding,
PATRA~\cite{patra2026} addresses pattern-aware alignment, and SenTSR-Bench~\cite{sentsr2026} injects domain knowledge for TS reasoning.
\citet{merrill2024towards} formalize TS reasoning as QA over numerical sequences, while S2TS-LLM~\cite{s2tsllm2025} uses spectral symbolization to bridge TS and language representations.
More recent work explores tool-augmented TSRMs and RL-based reasoning~\cite{timeart2025, timer1_2025}, as well as cross-domain temporal reasoning benchmarks~\cite{mtbench2025}.
However, \textit{all existing TSRMs target general domains} and none has been designed for financial TS, which require distinct reasoning about market-specific phenomena.

\textbf{LLMs in financial domain.}
FinTSB~\cite{liu2025fintsb} benchmarks forecasting on stocks but lacks reasoning tasks.
FinBen~\cite{finben2024} evaluates LLMs across financial NLP datasets but targets text-based tasks rather than raw price TS.
News-to-Forecast~\cite{newforecast2024} integrates event analysis with reflection and CausalStock~\cite{causalstock2024} leverages LLMs as denoised news encoders, but both require textual inputs rather than reasoning from raw prices.
No existing work combines \textit{raw financial TS} with \textit{reasoning QA} and \textit{CoT supervision}, the gap that FinSTaR addresses.
Recent work on RL-based financial reasoning~\cite{fin_rl_reason2025} and multimodal financial TS reasoning~\cite{vta2025} explores complementary directions but does not address the assessment--prediction distinction central to our approach.
A comparison with related work is shown in Table~\ref{tab:comparison}.

\textbf{CoT for structured data.}
CoT~\cite{wei2022chain} improves LLM reasoning on math~\cite{cobbe2021gsm8k} and science tasks, and STaR~\cite{star2022} bootstraps reasoning from self-generated rationales.
Program-of-Thought (PoT)~\cite{pot2023} executes code for arithmetic faithfulness, and verifiable-reward RL~\cite{counts2025} trains on outcome correctness.
For TS reasoning, \textit{existing CoT methods rely on human annotation~\cite{timeomni1} or GPT-generated rationales}, which are difficult to scale.
Our Compute-in-CoT ensures the same arithmetic correctness within text-based chains, needing no code execution environment and thus compatible with standard LLM inference.
Scenario-Aware CoT further extends CoT to stochastic prediction via multi-scenario reasoning, which PoT and verifiable-reward methods, focused on deterministic correctness, do not address.

\vspace{20pt}
\section{Dataset Construction}
\label{app:dataset}

\textbf{Dataset statistics.}
Each task has ${\sim}3{,}500$ training samples and ${\sim}1{,}000$ test samples per split, totaling ${\sim}35{,}000$ for training and ${\sim}10{,}000$ per test set.
All splits use a 120-day input window of daily closing prices.
Assessment tasks evaluate the current state and have no prediction horizon, whereas prediction tasks use task-specific forward horizons: Event Response and Support/Resistance use a 10-day horizon, while Drawdown Recovery, Volatility Forecast, Relative Performance, and Pair Convergence use a 20-day horizon.

\textbf{QA generation parameters.}
Table~\ref{tab:qa_params} lists the QA generation parameters used for constructing FinTSR-Bench, including window length, sample caps per task, and task-specific thresholds (e.g., correlation pos/neg thresholds, event $|z|$ threshold, relative-performance margin, drawdown-recovery minimum drawdown, volatility-forecast change threshold, and pair-convergence spread margin).

\begin{table}[h]
\centering
\caption{QA generation parameters.}
\label{tab:qa_params}
\vspace{5pt}
\adjustbox{width=\linewidth}{
\begin{tabular}{ll}
\toprule
\textbf{Parameter} & \textbf{Value} \\
\midrule
Window length & 120 trading days \\
Samples per task (raw) & 100{,}000 \\
Cap per task (train) & 3{,}500 \\
Cap per task (test) & 1{,}000 \\
Correlation thresholds & \begin{tabular}[t]{@{}l@{}}pos $>$ 0.30,\\ neg $<$ $-$0.10\end{tabular} \\
Event threshold & $|z|$ $>$ 2.5 \\
Rel.\ Perf.\ margin & 5\% \\
Rel.\ Perf.\ forward window & 20 days \\
S/R breakout margin & 3\% \\
Drawdown Recovery min drawdown & 5\% \\
Volatility Forecast change threshold & 25\% \\
Pair Convergence spread margin & 3\% \\
\bottomrule
\end{tabular}
}
\end{table}

\textbf{Class distributions.}
Table~\ref{tab:class_dist} reports the training-set class distribution. All tasks are approximately class-balanced via undersampling at 3,500 samples per task, so chance-level accuracy is $\approx 1/K$ for $K$-class tasks (e.g., 25\% for Drawdown's 4 classes, 50\% for binary prediction tasks).

\begin{table}[h]
\centering
\caption{\textbf{Training set class distribution.} All tasks are approximately class-balanced via undersampling (3,500 samples per task).}
\label{tab:class_dist}
\vspace{5pt}
\adjustbox{width=\linewidth}{
\begin{tabular}{l|l}
\toprule
\textbf{Task} & \textbf{Class distribution (\%)} \\
\midrule
Drawdown & A: 24.2, B: 24.3, C: 26.0, D: 25.5 \\
Volatility Regime & A: 33.5, B: 33.1, C: 33.4 \\
Trend Direction & A: 20.2, B: 20.4, C: 19.6, D: 21.0, E: 18.7 \\
Correlation & A: 32.9, B: 34.2, C: 32.9 \\
\midrule
Event Response & A: 49.0, B: 51.0 \\
Support/Resistance & A: 50.1, B: 49.9 \\
Drawdown Recovery & A: 49.0, B: 51.0 \\
Volatility Forecast & A: 48.9, B: 51.1 \\
Relative Performance & A: 51.4, B: 48.6 \\
Pair Convergence & A: 50.5, B: 49.5 \\
\bottomrule
\end{tabular}
}
\end{table}

\newpage
\textbf{Label construction algorithms.}
We detail the labeling procedure for each prediction task to ensure reproducibility.
\begin{itemize}
\item \textbf{Event Response.} We compute daily returns and their z-scores over the 120-day window. A trading day with $|z| > 2.5$ is flagged as an ``event.'' The label is determined by comparing the 10-day post-event return direction with the event-day return: same direction $\rightarrow$ Persistence, opposite $\rightarrow$ Mean-reversion.
\item \textbf{Support/Resistance.} We identify support (resistance) as the minimum (maximum) closing price over a rolling 60-day lookback window. If the current price is within 5\% of a key level, the sample is included. The label is determined by whether the price moves beyond the level by $>$3\% (Breakout) or stays within it (Bounce) over the next 10 days.
\item \textbf{Drawdown Recovery.} We compute peak price as the running maximum over the window. Stocks with drawdown $>$5\% from peak are included. The label is Recovery if the price moves $>$3\% toward the peak over the next 20 days, and Deepens otherwise.
\item \textbf{Volatility Forecast.} We compute the ratio of 20-day to 120-day return standard deviation. The label is Increase if the 20-day volatility computed at $t{+}20$ exceeds the current 20-day volatility by $>$25\%, and Decrease otherwise.
\item \textbf{Relative Performance.} Given two stocks, the label is determined by which stock achieves a higher cumulative return over the next 20 days. Pairs with $<$5\% return difference are excluded to avoid ambiguous labels.
\item \textbf{Pair Convergence.} We compute the normalized spread as $(P_A - P_B) / (P_A + P_B)$, where $P_A$ and $P_B$ are the closing prices. The label is Convergence if $|\text{spread}_{t+20}| < |\text{spread}_t|$, and Divergence otherwise. Pairs with spread $<$3\% are excluded.
\end{itemize}
To avoid overlapping-window leakage, each task uses a task-specific random seed for sample selection, and we enforce a minimum gap of 20 trading days between consecutive samples from the same stock.

\vspace{20pt}
\section{Training Setup}
\label{app:training_setup}

We fine-tune two backbones via LoRA~\cite{hu2022lora}: TimeOmni-1-7B~\cite{timeomni1}, an LLM trained with TS reasoning SFT and RL on top of Qwen2.5-7B, and Qwen2.5-7B-Instruct~\cite{qwen2025qwen25}, the same base LLM without TS-specific training.
As shown in Table~\ref{tab:main_results}, these are the best-performing models from their respective categories.
We use LoRA with $r{=}32$, $\alpha{=}64$, learning rate $5{\times}10^{-5}$, 4 epochs, batch size 1 with gradient accumulation 16, max sequence length 4096, trained on 2$\times$ NVIDIA L40S GPUs.

\vspace{20pt}
\section{Forecasting Baseline Protocol}
\label{app:forecast_protocol}

\begin{table*}[t]
\centering
\caption{\textbf{Forecasting baselines on prediction.} Values are (Test~A, Test~B, Test~C). FinSTaR outperforms forecasting baselines, with the largest gains on tasks requiring scenario reasoning.}
\label{tab:forecast_vs_reasoning}
\vspace{3pt}
\adjustbox{width=\linewidth}{
\footnotesize
\begin{tabular}{l|cccccc|c}
\toprule
\textbf{Method} & \textbf{Event} & \textbf{S/R} & \textbf{DDR} & \textbf{V.F.} & \textbf{R.P.} & \textbf{P.C.} & \textbf{Avg.} \\
\midrule
\midrule
\multicolumn{8}{l}{\textit{Statistical Forecasting}} \\
\midrule
Last Value & (42.7, 44.5, 46.4) & (61.7, 62.1, 59.6) & (49.9, 51.2, 49.3) & (48.0, 53.6, 51.9) & (48.5, 49.9, 47.0) & (50.3, 48.3, 49.1) & (50.2, 51.6, 50.5) \\
ETS & (49.2, 52.0, 50.3) & (62.1, 59.5, 58.9) & (46.3, 47.6, 49.9) & (48.0, 53.6, 51.9) & (49.5, \second{53.9}, 50.2) & (54.6, 56.5, 52.3) & (51.6, 53.8, 52.2) \\
MA & (51.2, 52.7, 52.1) & (60.1, 57.2, 57.8) & (45.1, 46.1, 46.3) & (50.6, 56.8, 53.9) & (50.8, 53.2, 51.3) & (\second{59.0}, \second{57.6}, \second{56.5}) & (52.8, 53.9, 53.0) \\
Momentum & (49.5, 47.2, 50.4) & (68.9, 68.8, 68.4) & (57.1, 55.6, 57.6) & (48.0, 53.6, 51.9) & (49.7, 45.9, 49.0) & (50.5, 49.5, 52.3) & (53.9, 53.4, 54.9) \\
Drift & (52.9, 50.7, 50.4) & (71.0, 72.5, 68.1) & (\second{63.5}, \second{66.0}, \second{64.1}) & (48.0, 53.6, 51.9) & (48.4, 47.8, 50.4) & (50.3, 48.3, 49.1) & (55.7, 56.5, 55.7) \\
\midrule
\midrule
\multicolumn{8}{l}{\textit{DL Forecasting}} \\
\midrule
TFT & (52.8, 53.0, 50.6) & (61.6, 60.2, 58.1) & (36.3, 38.9, 39.9) & (48.0, 53.7, 51.9) & (50.4, 49.2, \first{54.3}) & (52.2, 50.9, 50.4) & (50.2, 51.0, 50.9) \\
DeepAR & (49.1, 53.1, 51.2) & (64.0, 61.4, 60.6) & (50.4, 50.3, 51.0) & (48.2, 54.2, 52.0) & (48.4, 53.7, 49.9) & (56.4, 55.9, 56.2) & (52.8, 54.8, 53.5) \\
DLinear & (47.9, 48.7, 52.0) & (63.8, 62.1, 58.5) & (64.9, \second{66.9}, 63.0) & (58.2, 55.3, 57.7) & (46.4, 52.5, 50.2) & (55.0, 56.8, 54.8) & (56.0, \second{57.0}, 56.0) \\
TiDE & (\second{54.1}, \second{54.4}, \second{53.0}) & (57.4, 54.6, 56.7) & (58.1, 57.2, 56.7) & (\second{58.9}, 55.6, \second{58.0}) & (50.5, \first{54.1}, 50.3) & (56.2, \second{57.5}, \second{57.0}) & (55.9, 55.6, 55.3) \\
PatchTST & (48.6, 49.2, 49.6) & (\second{80.4}, \second{79.7}, \second{79.1}) & (49.3, 48.7, 49.1) & (69.5, \second{64.6}, 68.9) & (48.2, 48.6, 48.5) & (48.4, 44.9, 47.7) & (\second{57.4}, 56.0, \second{57.1}) \\
Chronos-1 & (53.4, \second{54.9}, 48.9) & (69.4, 71.0, 66.5) & (50.2, 53.9, 50.1) & (47.9, 54.2, 52.0) & (50.1, 48.0, 49.8) & (53.0, 51.0, 53.0) & (54.0, 55.5, 53.4) \\
Chronos-2 & (53.2, 53.0, 51.1) & (67.9, 69.5, 64.4) & (45.3, 47.0, 46.3) & (48.0, 53.6, 51.9) & (\second{51.4}, 49.1, \second{52.4}) & (53.2, 48.2, 51.3) & (53.2, 53.4, 52.9) \\
\midrule
\midrule
\multicolumn{8}{l}{\textit{Financial TSRM (Ours)}} \\
\midrule
\rowcolor{yellow!15} \textbf{FinSTaR} & (\first{57.2}, \first{55.6}, \first{53.5}) & (\first{80.5}, \first{80.0}, \first{79.5}) & (\first{74.4}, \first{75.0}, \first{72.9}) & (\first{79.5}, \first{77.3}, \first{78.8}) & (\first{51.5}, 50.1, 53.0) & (\first{65.8}, \first{64.3}, \first{64.6}) & (\first{68.2}, \first{67.0}, \first{67.0}) \\
\bottomrule
\end{tabular}
}
\end{table*}

\textbf{Prediction tasks: TSRM vs.\ TS forecasting models.}
While TSRMs can handle both assessment and prediction through reasoning, traditional TS forecasting models can also solve prediction tasks without reasoning, via a \textit{forecast-then-classify} pipeline (predict future prices numerically, then convert to classification labels).
We compare FinSTaR against such forecasting models: statistical methods (Last Value, MA, ETS, Drift, Momentum) and DL models (DLinear~\cite{dlinear}, Chronos-1/2~\cite{chronos2024}, DeepAR~\cite{deepar}, PatchTST~\cite{patchtst}, TFT~\cite{tft}, TiDE~\cite{tide}).
As shown in Table~\ref{tab:forecast_vs_reasoning}, FinSTaR outperforms the forecasting models on prediction tasks.

\textbf{Forecasting baseline protocol.}
All forecasting baselines use the same 120-day input window as FinSTaR and predict over the task-specific horizon (10 or 20 days).
No exogenous covariates are used; all models receive only univariate closing prices.
Predicted price trajectories are converted to classification labels using the same thresholds as the QA generation (e.g., direction of predicted return for Event Response, predicted volatility ratio for Volatility Forecast).

\textbf{Statistical methods}:
\begin{itemize}
\item \textbf{Last Value.} Predicts the last observed value for all future steps (naive baseline).
\item \textbf{Moving Average (MA).} Predicts the mean of the last $k$ observations ($k{=}20$).
\item \textbf{Exponential Smoothing (ETS)~\cite{ets}.} Weighted average with exponentially decaying weights; captures level and trend.
\item \textbf{Drift.} Extrapolates a linear trend from the first to the last observed value.
\item \textbf{Momentum.} Predicts continuation of the recent return direction over the past 20 days.
\end{itemize}

\textbf{Deep learning models} (trained per-task using AutoGluon TimeSeries with default hyperparameters and early stopping):
\begin{itemize}
\item \textbf{DLinear~\cite{dlinear}.} Decomposes TS into trend and remainder via moving average, then applies linear layers.
\item \textbf{DeepAR~\cite{deepar}.} Autoregressive RNN that produces probabilistic forecasts via learned likelihood parameters.
\item \textbf{PatchTST~\cite{patchtst}.} Patch-based Transformer that segments TS into subseries-level patches for efficient attention.
\item \textbf{TFT~\cite{tft}.} Temporal Fusion Transformer with variable selection, gating, and multi-horizon attention.
\item \textbf{TiDE~\cite{tide}.} MLP-based encoder-decoder with dense residual connections for long-term forecasting.
\end{itemize}

\textbf{Pre-trained forecasting models}:
\begin{itemize}
\item \textbf{Chronos-1~\cite{chronos2024}.} Tokenizes TS values into discrete bins and uses a T5-based language model for forecasting.
\item \textbf{Chronos-2~\cite{chronos2024}.} Updated version with improved tokenization and larger pre-training corpus.
\end{itemize}

\section{Backbone Sensitivity}
\label{app:backbone_sensitivity}

\begin{table*}[t]
\centering
\caption{\textbf{Backbone sensitivity.} Scenario-Aware CoT improves prediction only with a TS reasoning-trained backbone (TimeOmni-1, +11.3\%), while it hurts the general-purpose LLM (Qwen, $-$5.0\%), suggesting that \textit{prior TS reasoning ability is necessary} for effective scenario-based reasoning.}
\label{tab:backbone}
\vspace{4pt}
\makebox[\textwidth][c]{\adjustbox{width=0.818\textwidth}{
\begin{tabular}{c|l|cc|c}
\toprule
\textbf{Backbone} & \textbf{Config} & \textbf{Assessment Avg.} & \textbf{Prediction Avg.} & \textbf{Overall} \\
\midrule
\multirow{3.5}{*}{Qwen2.5-7B (w/o TS Reasoning SFT)}
& w/o CoT & 57.0 & \first{56.1} & 56.4 \\
 & \cellcolor{yellow!15}  w/ CoT & \cellcolor{yellow!15}  \first{67.7} & \cellcolor{yellow!15}  51.1 & \cellcolor{yellow!15}  \first{57.8} \\
 \cmidrule(lr){2-5}
 & $\Delta$ & \up{+10.7} & $-$5.0 & \up{+1.3} \\
\midrule
\multirow{3.5}{*}{TimeOmni-1-7B (w/ TS Reasoning SFT)}
& w/o CoT & 54.3 & 56.9 & 55.9 \\
 & \cellcolor{yellow!15}   w/ CoT & \cellcolor{yellow!15}  \first{95.0} & \cellcolor{yellow!15}  \first{68.2} & \cellcolor{yellow!15}  \first{78.9} \\
  \cmidrule(lr){2-5}
 & $\Delta$ & \up{+40.7} & \up{+11.3} & \up{+23.1} \\
\bottomrule
\end{tabular}
}}
\end{table*}

Table~\ref{tab:backbone} compares the effect of CoT across two backbones: TimeOmni-1-7B (trained with TS reasoning SFT + RL) and Qwen2.5-7B (general-purpose LLM without TS-specific training).\footnote{Both Qwen2.5-7B rows (w/o CoT and w/ CoT) are fine-tuned on the same FinTSR-Bench data via LoRA, serving as SFT baselines that isolate the effect of TS reasoning pre-training from our CoT strategies.}
CoT improves assessment accuracy for both backbones, since Compute-in-CoT involves explicit arithmetic that even general-purpose LLMs can perform.
However, for prediction tasks, CoT improves TimeOmni-1 (+11.3\%) but \textit{hurts} Qwen ($-$5.0\%).
We attribute this to the fact that Scenario-Aware CoT requires reasoning about price dynamics, which demands prior TS reasoning ability.
Without TS reasoning training, the general-purpose LLM generates plausible-sounding but ungrounded scenarios, introducing noise rather than useful reasoning signal.
This suggests that \textit{Scenario-Aware CoT is effective only when the backbone already possesses TS reasoning ability}, highlighting the synergy between TS reasoning pre-training and structured CoT.

\section{Training Data Efficiency}
\label{app:data_efficiency}

\begin{table*}[t]
\centering
\caption{\textbf{Effect of training set size.} FinSTaR performance with 350, 875, 1750, and 3500 training samples per task. With only 10\% of the full data (350 samples/task), FinSTaR already achieves 71.7\%, indicating that \textit{structured CoT strategies are sample-efficient.}}
\label{tab:data_eff_full}
\vspace{4pt}
\adjustbox{width=\textwidth}{
\begin{tabular}{l|l|cccc|cccccc|c}
\toprule
 \multirow{2.5}{*}{\textbf{Method}} &\multirow{2.5}{*}{\textbf{\# Samples}} & \multicolumn{4}{c|}{\textit{Assessment}} & \multicolumn{6}{c|}{\textit{Prediction}} & \multirow{2.5}{*}{\textbf{Avg.}} \\
\cmidrule(lr){3-6} \cmidrule(lr){7-12}
& & \textbf{Draw.} & \textbf{Vol.} & \textbf{Trend} & \textbf{Corr.} & \textbf{Event} & \textbf{S/R} & \textbf{DDR} & \textbf{V.F.} & \textbf{R.P.} & \textbf{P.C.} & \\
\midrule
\rowcolor{gray!15} TimeOmni-1-7B & 3500 (100\%) &  75.7 & 35.6 & 50.5 & 31.5 & 49.1 & 54.1 & 59.3 & 58.2 & 50.7 & 48.4 & 51.3 \\
\rowcolor{gray!15}Qwen2.5-7B (w/o CoT) & 3500 (100\%) &  77.5 & 40.4 & 75.2 & 34.7 & 53.0 & 53.7 & 67.1 & 60.6 & 50.7 & 51.3 & 56.4 \\
\rowcolor{gray!15}Qwen2.5-7B (w/ CoT)& 3500 (100\%) & 96.4 & 46.6 & 92.3 & 35.4 & 47.6 & 51.1 & 57.2 & 48.7 & 47.3 & 55.0 & 57.8 \\
\midrule
\multirow{4}{*}{\textbf{FinSTaR}} & 350 (10\%) &  98.1 & 80.4 & 98.0 & 71.6 & 53.9 & 54.8 & 70.6 & 73.9 & 51.4 & 63.8 & 71.7 \\
& 875 (25\%)& 98.0 & 87.3 & 99.0 & 83.1 & 57.4 & 53.9 & 73.6 & 78.2 & 51.5 & 64.5 & 74.7 \\
& 1750 (50\%)& 99.1 & 90.0 & 82.2 & 85.5 & 54.5 & 70.9 & 74.7 & 78.7 & 51.5 & 65.0 & 75.2 \\
& 3500  (100\%)& 99.3 & 93.5 & 99.3 & 88.1 & 57.2 & 80.5 & 74.4 & 79.5 & 51.5 & 65.8 & 78.9 \\
\bottomrule
\end{tabular}
}
\end{table*}

\textbf{Effect of training set size.}
We evaluate FinSTaR's performance with reduced training data size (350, 875, 1750, and 3500 samples per task).
As shown in Table~\ref{tab:data_eff_full}, FinSTaR already achieves 71.7\% overall accuracy with only 350 samples/task (10\% of full).
Performance improves to 74.7\% at 875 and 75.2\% at 1750 samples, with diminishing returns beyond that.
This indicates that our Compute-in-CoT and Scenario-Aware CoT strategies are \textit{data-efficient}, requiring relatively few examples to learn the underlying reasoning patterns.

\renewcommand{\dblfloatpagefraction}{0.1}
\section{Scenario Templates}
\label{app:scenario_templates}

Table~\ref{tab:scenario_templates} lists the scenario templates used in Scenario-Aware CoT for all six prediction tasks. For each task and answer, we define three scenarios (base, adverse, favorable) grounded in financial domain knowledge. Assessment tasks (Drawdown, Volatility Regime, Trend Direction, Correlation) use deterministic Compute-in-CoT and do not require scenario templates.

\begin{table*}[t]
\centering
\caption{\textbf{Scenario templates for prediction tasks.} Each (task, answer) pair has three domain-specific scenarios.}
\label{tab:scenario_templates}
\vspace{6pt}
\adjustbox{width=\linewidth}{
\footnotesize
\begin{tabular}{l|l|p{4cm}|p{4cm}|p{4cm}}
\toprule
\textbf{Task} & \textbf{Answer} & \textbf{Base Case} & \textbf{Adverse} & \textbf{Favorable} \\
\midrule
\multirow{4.5}{*}{\textbf{Event Response}} & Mean-reversion & Extreme move partially reverses as market recognizes overreaction & Secondary catalyst reinforces shock direction & Positive news accelerates recovery \\
\cmidrule{2-5}
 & Persistence & Shock reflects genuine new information being priced in & Mean-reversion forces prove it was an overreaction & Additional confirming news amplifies the move \\
\midrule
\multirow{4.5}{*}{\textbf{Support/Resistance}} & Breakout & Price breaks through with sustained momentum & False breakout traps traders, price reverses & High volume accelerates move beyond level \\
\cmidrule{2-5}
 & Bounce & Key level holds as institutional orders defend it & Sustained pressure overwhelms the level & Strong bounce triggers significant reversal \\
\midrule
\multirow{4.5}{*}{\textbf{Drawdown Recovery}} & Recovery & Drawdown stabilizes, value buyers accumulate & Cause worsens (earnings revision, guidance cut) & Positive catalyst triggers V-shaped recovery \\
\cmidrule{2-5}
 & Further decline & Selling pressure continues, drawdown deepens & Surprise positive development reverses decline & Risk-off sentiment compounds weakness \\
\midrule
\multirow{4.5}{*}{\textbf{Volatility Forecast}} & Increase & Volatility clustering continues, uncertainty persists & Uncertainty resolves, volatility collapses & New uncertainty compounds existing volatility \\
\cmidrule{2-5}
 & Decrease & Volatility mean-reverts as shock dissipates & Unexpected event triggers new spike & Broad market calming accelerates compression \\
\midrule
\multirow{4.5}{*}{\textbf{Relative Performance}} & Stock A & Momentum trends persist, Stock A continues trajectory & Stock A headwind, Stock B catalyst reverses gap & Sector tailwinds boost Stock A's advantage \\
\cmidrule{2-5}
 & Stock B & Stock B strengthening continues, Stock A fades & Stock A recovers, re-establishing outperformance & Stock B catalyst (upgrade, M\&A) widens lead \\
\midrule
\multirow{4.5}{*}{\textbf{Pair Convergence}} & Converge & Spread narrows as co-movement reasserts & Structural change sustains divergence & Common catalyst forces rapid convergence \\
\cmidrule{2-5}
 & Diverge & Different fundamentals sustain widening spread & Common catalyst narrows the spread & Company-specific event accelerates divergence \\
\bottomrule
\end{tabular}
}
\end{table*}

\section{Full Task Examples}
\label{app:tasks}

\begin{figure*}[h]
\centering
\adjustbox{width=\linewidth}{%
\includegraphics[width=0.95\linewidth]{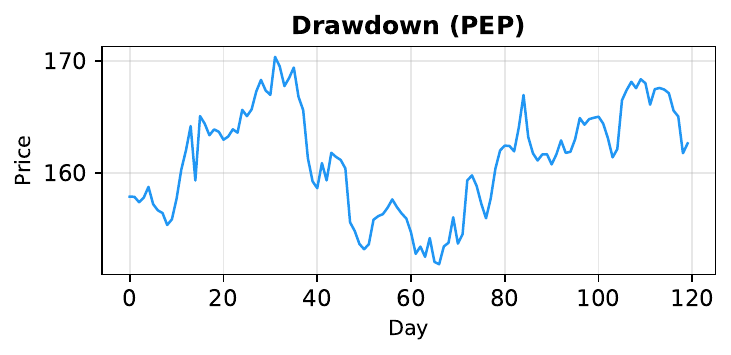}
}
\vspace{4pt}

\textbf{Task: Drawdown --- Ticker: PEP --- GT: (B) Pullback} \\[4pt]
\begin{minipage}{0.95\linewidth}
\fbox{\parbox{\dimexpr\linewidth-2\fboxsep-2\fboxrule}{
\textbf{Input:} You are analyzing the stock PEP. Below are the daily closing prices for the most recent 120 trading days: [\textit{prices shown above}]. Based on these prices, assess the current drawdown phase. \\
\hspace{6pt}(A) At/Near Peak ($<$3\%) \\
\hspace{6pt}(B) Pullback (3--10\%) \\
\hspace{6pt}(C) Correction (10--20\%) \\
\hspace{6pt}(D) Severe Decline ($>$20\%)
}} \\[4pt]
\fbox{\parbox{\dimexpr\linewidth-2\fboxsep-2\fboxrule}{
\textbf{Output:} \\[6pt]
\texttt{<think>} \\[4pt]
Step 1 --- Find the peak price: \\
Scanning the 120-day series, the highest price is 170.39 (around day 32). \\[6pt]
Step 2 --- Current price: \\
The last price in the series is 162.65. \\[6pt]
Step 3 --- Calculate drawdown: \\
Drawdown = (Peak $-$ Current) / Peak = (170.39 $-$ 162.65) / 170.39 = 4.5\% \\[6pt]
Step 4 --- Classify: \\
4.5\% is between 3\% and 10\%, which corresponds to (B) Pullback. \\[4pt]
\texttt{</think>} \\
\texttt{<answer>(B)</answer>}
}}
\end{minipage}
\caption{(Single-Assessment) Example of Drawdown task.}
\label{fig:ex_first}
\vspace{15pt}
\end{figure*}

\begin{figure*}[h]
\centering
\adjustbox{width=\linewidth}{%
\includegraphics[width=0.95\linewidth]{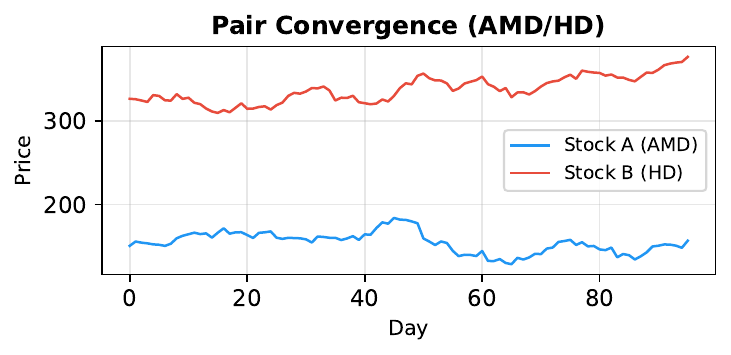}
}
\vspace{4pt}

\textbf{Task: Pair Convergence --- Tickers: AMD vs HD --- GT: (B) Divergence} \\[4pt]
\begin{minipage}{0.95\linewidth}
\fbox{\parbox{\dimexpr\linewidth-2\fboxsep-2\fboxrule}{
\textbf{Input:} You are comparing two stocks. Stock A (AMD) and Stock B (HD) daily closing prices: [\textit{prices shown above}]. The price spread (normalized difference) is 0.113. Predict how the spread will change over the next 20 days. \\
\hspace{6pt}(A) Convergence \\
\hspace{6pt}(B) Divergence
}} \\[4pt]
\fbox{\parbox{\dimexpr\linewidth-2\fboxsep-2\fboxrule}{
\textbf{Output:} \\[6pt]
\texttt{<think>} \\[4pt]
Step 1 --- Pair: AMD, HD. \\[6pt]
Step 2 --- Current spread: 0.113. \\[6pt]
Step 3 --- Scenario analysis: \\
\hspace{2em}$\bullet$ \textit{Base case} (most probable --- no major external events): \\
\hspace{2.7em}Different fundamentals sustain the widening spread. The divergence continues as the two stocks respond to distinct sector dynamics. \\[4pt]
\hspace{2em}$\bullet$ \textit{Adverse scenario} (external shock reverses expectation): \\
\hspace{2.7em}A common catalyst (e.g., index rebalancing, macro shock) narrows the spread by forcing temporary co-movement. \\[4pt]
\hspace{2em}$\bullet$ \textit{Favorable scenario} (catalyst strengthens expectation): \\
\hspace{2.7em}Company-specific events accelerate the divergence further as one stock's fundamentals diverge from the other's. \\[6pt]
Step 4 --- Scenario assessment: \\
The base case and favorable scenario (2 of 3) support Divergence. \\
The adverse scenario requires a specific external trigger not present in current data. \\
Without new information, the base case is the most probable outcome. \\[6pt]
Step 5 --- Judgment: Base case favors Divergence $\rightarrow$ (B). \\[4pt]
\texttt{</think>} \\
\texttt{<answer>(B)</answer>}
}}
\end{minipage}
\caption{(Multi-Prediction) Example of Pair Convergence task.}
\label{fig:ex_last}
\end{figure*}

We present actual model input, generated CoT reasoning, and output for all ten tasks. Each example shows a correctly predicted sample from Test~A, with the corresponding price series visualization.
Figures~\ref{fig:ex_first}--\ref{fig:ex_last} present one representative example per task, covering the four assessment tasks followed by the six prediction tasks.


\begin{figure*}[h]
\centering
\adjustbox{width=\linewidth}{%
\includegraphics[width=0.95\linewidth]{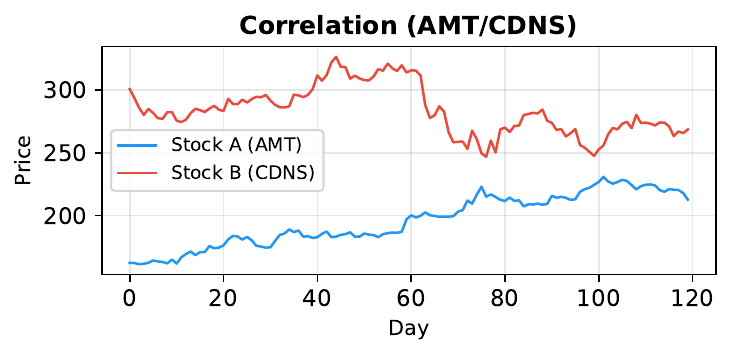}
}
\vspace{4pt}

\textbf{Task: Correlation --- Tickers: AMT vs CDNS --- GT: (B) Negative correlation} \\[4pt]
\begin{minipage}{0.95\linewidth}
\fbox{\parbox{\dimexpr\linewidth-2\fboxsep-2\fboxrule}{
\textbf{Input:} You are analyzing the relationship between two stocks. Stock A (AMT) and Stock B (CDNS) daily closing prices (120 days each): [\textit{prices shown above}]. Determine the correlation between their daily returns. \\
\hspace{6pt}(A) Positive \\
\hspace{6pt}(B) Negative \\
\hspace{6pt}(C) No significant correlation
}} \\[4pt]
\fbox{\parbox{\dimexpr\linewidth-2\fboxsep-2\fboxrule}{
\textbf{Output:} \\[6pt]
\texttt{<think>} \\[4pt]
Step 1 --- Compare price changes in 12 periods ($\sim$10 days each): \\
\hspace{1em}Days 1-10: A=+1.6\%, B=$-$6.0\% \\
\hspace{1em}Days 11-20: A=+7.8\%, B=+3.2\% \\
\hspace{1em}Days 21-30: A=$-$1.0\%, B=+4.5\% \\
\hspace{1em}Days 31-40: A=+2.3\%, B=$-$5.1\% \\
\hspace{1em}Days 41-50: A=$-$0.8\%, B=+1.9\% \\
\hspace{1em}Days 51-60: A=+3.1\%, B=$-$2.4\% \\
\hspace{1em}Days 61-70: A=$-$2.6\%, B=+4.0\% \\
\hspace{1em}Days 71-80: A=+1.5\%, B=$-$1.8\% \\
\hspace{1em}Days 81-90: A=$-$1.9\%, B=+2.7\% \\
\hspace{1em}Days 91-100: A=+0.4\%, B=$-$0.6\% \\
\hspace{1em}Days 101-110: A=$-$1.5\%, B=+2.2\% \\
\hspace{1em}Days 111-120: A=+2.4\%, B=$-$3.1\% \\[6pt]
Step 2 --- Pattern: \\
When A rises, B tends to fall, and when A falls, B tends to rise. The two stocks move in opposite directions across most periods. \\[6pt]
Step 3 --- negative correlation $\rightarrow$ (B). \\[4pt]
\texttt{</think>} \\
\texttt{<answer>(B)</answer>}
}}
\end{minipage}
\caption{(Multi-Assessment) Example of Correlation task.}
\end{figure*}

\begin{figure*}[h]
\centering
\adjustbox{width=\linewidth}{%
\includegraphics[width=0.95\linewidth]{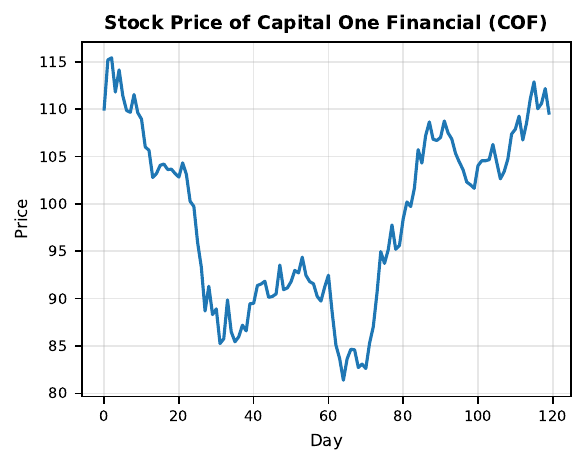}
}
\vspace{4pt}

\textbf{Task: Volatility Regime --- Ticker: COF --- GT: (B) Normal volatility} \\[4pt]
\begin{minipage}{0.95\linewidth}
\fbox{\parbox{\dimexpr\linewidth-2\fboxsep-2\fboxrule}{
\textbf{Input:} You are analyzing the stock COF. Below are the daily closing prices (120 days): [\textit{prices shown above}]. Compute the ratio of recent (last 20 days) to overall (120-day) volatility. \\
\hspace{6pt}(A) Low ($<$0.6) \\
\hspace{6pt}(B) Normal (0.6--1.6) \\
\hspace{6pt}(C) High ($>$1.6)
}} \\[4pt]
\fbox{\parbox{\dimexpr\linewidth-2\fboxsep-2\fboxrule}{
\textbf{Output:} \\[6pt]
\texttt{<think>} \\[4pt]
Step 1 --- Compute daily returns from prices: \\
\hspace{1em}e.g., (54.32$\rightarrow$54.51): +0.35\%, (54.51$\rightarrow$54.10): $-$0.75\%, (54.10$\rightarrow$54.78): +1.26\% \\
\hspace{1em}\ldots \\[6pt]
Step 2 --- Overall volatility (std of all returns): 2.02\% \\[6pt]
Step 3 --- Recent 20-day volatility: 1.50\% \\[6pt]
Step 4 --- Ratio = 1.50 / 2.02 = 0.74 \\
Recent volatility is close to the long-term average --- normal conditions. \\
$\rightarrow$ (B) normal volatility. \\[4pt]
\texttt{</think>} \\
\texttt{<answer>(B)</answer>}
}}
\end{minipage}
\caption{(Single-Assessment) Example of Volatility Regime task.}
\vspace{15pt}
\end{figure*}

\begin{figure*}[h]
\centering
\adjustbox{width=\linewidth}{%
\includegraphics[width=0.95\linewidth]{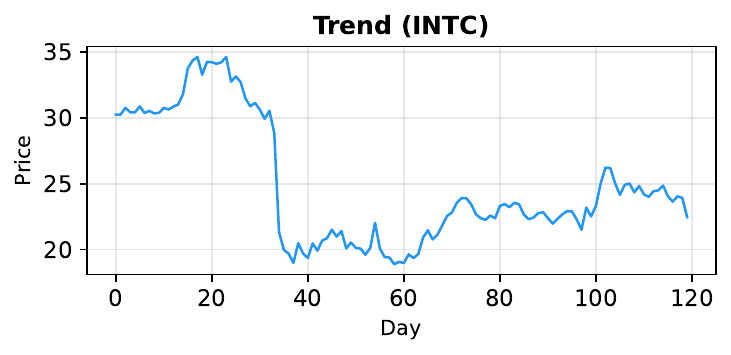}
}
\vspace{4pt}

\textbf{Task: Trend Direction --- Ticker: INTC --- GT: (E) Strong Downtrend} \\[4pt]
\begin{minipage}{0.95\linewidth}
\fbox{\parbox{\dimexpr\linewidth-2\fboxsep-2\fboxrule}{
\textbf{Input:} You are analyzing the stock INTC. Below are the daily closing prices (120 days): [\textit{prices shown above}]. Classify the overall trend direction. \\
\hspace{6pt}(A) Strong Uptrend ($>$20\%) \\
\hspace{6pt}(B) Mild Uptrend (5--20\%) \\
\hspace{6pt}(C) Sideways ($-$5\% to +5\%) \\
\hspace{6pt}(D) Mild Downtrend (5--20\% decline) \\
\hspace{6pt}(E) Strong Downtrend ($>$20\% decline)
}} \\[4pt]
\fbox{\parbox{\dimexpr\linewidth-2\fboxsep-2\fboxrule}{
\textbf{Output:} \\[6pt]
\texttt{<think>} \\[4pt]
Step 1 --- Read start and end prices: \\
Start (day 1): 30.27 \\
End (day 120): 22.47 \\[6pt]
Step 2 --- Compute cumulative return: \\
Return = (22.47 $-$ 30.27) / 30.27 = $-$25.8\% \\[6pt]
Step 3 --- Classify: \\
$-$25.8\% $\rightarrow$ (E) Strong Downtrend ($<-$20\%). \\[4pt]
\texttt{</think>} \\
\texttt{<answer>(E)</answer>}
}}
\end{minipage}
\caption{(Single-Assessment) Example of Trend Direction task.}
\end{figure*}


\begin{figure*}[h]
\vspace{-10pt}
\centering
\adjustbox{width=0.745\linewidth}{
\includegraphics[width=0.73\linewidth]{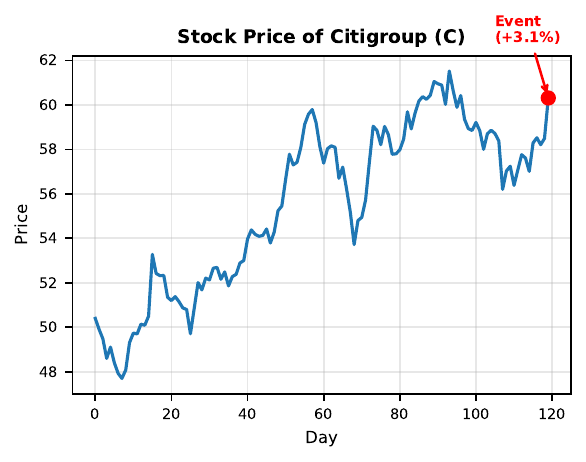}
}
\vspace{1pt}

\textbf{Task: Event Response --- Ticker: C --- GT: (B) Persistence} \\[4pt]
\begin{minipage}{0.95\linewidth}
\fbox{\parbox{\dimexpr\linewidth-2\fboxsep-2\fboxrule}{
\textbf{Input:} You are analyzing the stock C. Below are the daily closing prices (120 days): [\textit{prices shown above}]. The stock just experienced a significant positive shock. Predict the outcome over the next 10 days. \\
\hspace{6pt}(A) Mean-reversion \\
\hspace{6pt}(B) Persistence
}} \\[4pt]
\fbox{\parbox{\dimexpr\linewidth-2\fboxsep-2\fboxrule}{
\textbf{Output:} \\[6pt]
\texttt{<think>} \\[4pt]
Step 1 --- Event: positive shock, z=2.20, return=+1.85\%. \\[6pt]
Step 2 --- Pre-event trend: flat. \\[6pt]
Step 3 --- Scenario analysis: \\
\hspace{2em}$\bullet$ \textit{Base case} (most probable --- no major external events): \\
\hspace{2.7em}The positive shock reflects genuine new information being priced in. The market continues to absorb the news over the following days, sustaining the move in the same direction. \\[4pt]
\hspace{2em}$\bullet$ \textit{Adverse scenario} (external shock reverses expectation): \\
\hspace{2.7em}Mean-reversion forces kick in as profit-taking and overreaction corrections pull the price back toward pre-event levels. \\[4pt]
\hspace{2em}$\bullet$ \textit{Favorable scenario} (catalyst strengthens expectation): \\
\hspace{2.7em}Additional confirming news (e.g., earnings beat, upgrade) amplifies the move and accelerates persistence. \\[6pt]
Step 4 --- Scenario assessment: \\
The base case and favorable scenario (2 of 3) support Persistence. \\
The adverse scenario requires a specific external trigger not present in current data. \\
Without new information, the base case is the most probable outcome. \\[6pt]
Step 5 --- Judgment: Base case favors Persistence $\rightarrow$ (B). \\[4pt]
\texttt{</think>} \\
\texttt{<answer>(B)</answer>}
}}
\end{minipage}
\caption{(Single-Prediction) Example of Event Response task.}
\vspace{-40pt}
\end{figure*}

\begin{figure*}[h]
\centering
\adjustbox{width=\linewidth}{%
\includegraphics[width=0.95\linewidth]{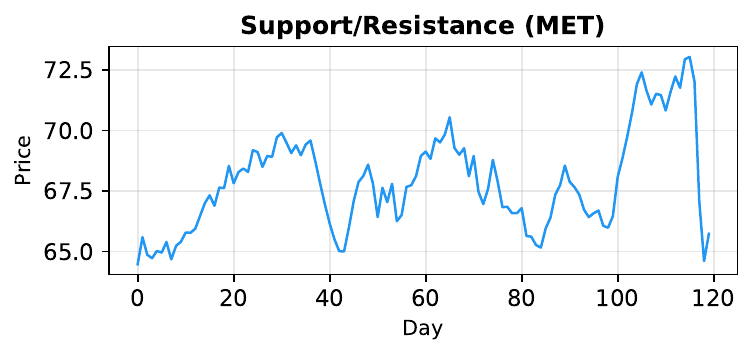}
}
\vspace{4pt}

\textbf{Task: Support/Resistance --- Ticker: MET --- GT: (B) Bounce} \\[4pt]
\begin{minipage}{0.95\linewidth}
\fbox{\parbox{\dimexpr\linewidth-2\fboxsep-2\fboxrule}{
\textbf{Input:} You are analyzing the stock MET. Below are the daily closing prices (120 days): [\textit{prices shown above}]. The stock is currently near a key support level. Predict the behavior over the next 10 days. \\
\hspace{6pt}(A) Breakout \\
\hspace{6pt}(B) Bounce
}} \\[4pt]
\fbox{\parbox{\dimexpr\linewidth-2\fboxsep-2\fboxrule}{
\textbf{Output:} \\[6pt]
\texttt{<think>} \\[4pt]
Step 1 --- Key level: support at 64.47. \\[6pt]
Step 2 --- Proximity: 1.9\%. \\[6pt]
Step 3 --- Scenario analysis: \\
\hspace{2em}$\bullet$ \textit{Base case} (most probable --- no major external events): \\
\hspace{2.7em}The key level holds as institutional orders defend it. Price reverses direction after testing the level. \\[4pt]
\hspace{2em}$\bullet$ \textit{Adverse scenario} (external shock reverses expectation): \\
\hspace{2.7em}Sustained pressure eventually overwhelms the level, leading to a delayed breakout. \\[4pt]
\hspace{2em}$\bullet$ \textit{Favorable scenario} (catalyst strengthens expectation): \\
\hspace{2.7em}A strong bounce off the level triggers a significant reversal move in the opposite direction. \\[6pt]
Step 4 --- Scenario assessment: \\
The base case and favorable scenario (2 of 3) support Bounce. \\
The adverse scenario requires a specific external trigger not present in current data. \\
Without new information, the base case is the most probable outcome. \\[6pt]
Step 5 --- Judgment: Base case favors Bounce $\rightarrow$ (B). \\[4pt]
\texttt{</think>} \\
\texttt{<answer>(B)</answer>}
}}
\end{minipage}
\caption{(Single-Prediction) Example of Support/Resistance task.}
\end{figure*}

\begin{figure*}[h]
\centering
\adjustbox{width=\linewidth}{%
\includegraphics[width=0.95\linewidth]{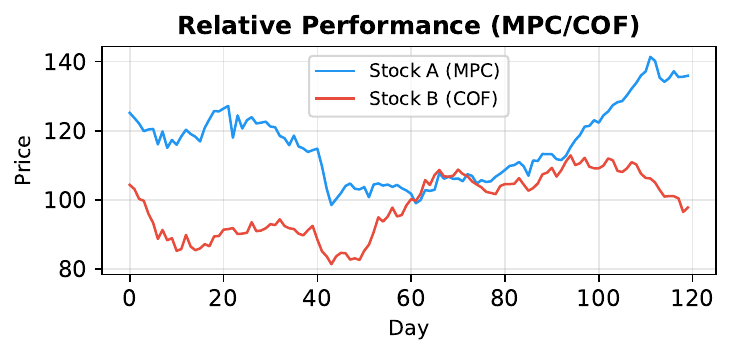}
}
\vspace{4pt}

\textbf{Task: Relative Performance --- Tickers: MPC vs COF --- GT: (A) Stock A outperforms} \\[4pt]
\begin{minipage}{0.95\linewidth}
\fbox{\parbox{\dimexpr\linewidth-2\fboxsep-2\fboxrule}{
\textbf{Input:} You are comparing two stocks. Stock A (MPC) and Stock B (COF) daily closing prices (120 days each): [\textit{prices shown above}]. Predict which stock will have a higher return over the next 20 days. \\
\hspace{6pt}(A) Stock A outperforms \\
\hspace{6pt}(B) Stock B outperforms
}} \\[4pt]
\fbox{\parbox{\dimexpr\linewidth-2\fboxsep-2\fboxrule}{
\textbf{Output:} \\[6pt]
\texttt{<think>} \\[4pt]
Step 1 --- Comparing MPC vs COF. \\[6pt]
Step 2 --- Forward window: 20 days. \\[6pt]
Step 3 --- Scenario analysis: \\
\hspace{2em}$\bullet$ \textit{Base case} (most probable --- no major external events): \\
\hspace{2.7em}Stock A's recent strengthening continues, while Stock B's momentum fades. The relative performance gap widens further in A's favor. \\[4pt]
\hspace{2em}$\bullet$ \textit{Adverse scenario} (external shock reverses expectation): \\
\hspace{2.7em}A company-specific headwind hits Stock A (e.g., earnings miss, regulatory issue), eroding its lead. \\[4pt]
\hspace{2em}$\bullet$ \textit{Favorable scenario} (catalyst strengthens expectation): \\
\hspace{2.7em}Sector tailwinds (e.g., favorable commodity prices) boost Stock A and widen its outperformance. \\[6pt]
Step 4 --- Scenario assessment: \\
The base case and favorable scenario (2 of 3) support Stock A outperforms. \\
The adverse scenario requires a specific external trigger not present in current data. \\
Without new information, the base case is the most probable outcome. \\[6pt]
Step 5 --- Judgment: Base case favors Stock A outperforms $\rightarrow$ (A). \\[4pt]
\texttt{</think>} \\
\texttt{<answer>(A)</answer>}
}}
\end{minipage}
\caption{(Multi-Prediction) Example of Relative Performance task.}
\vspace{15pt}
\end{figure*}

\begin{figure*}[h]
\centering
\adjustbox{width=\linewidth}{%
\includegraphics[width=0.95\linewidth]{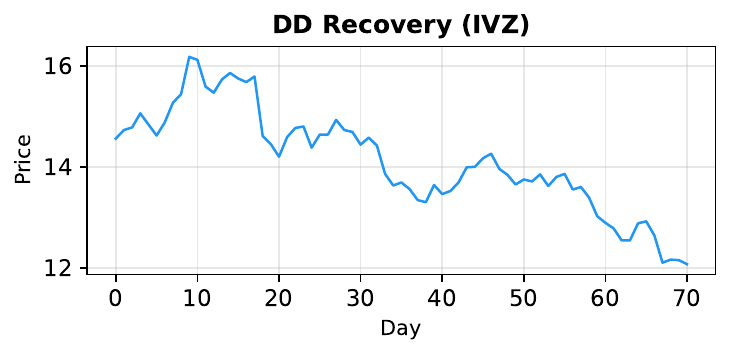}
}
\vspace{4pt}

\textbf{Task: Drawdown Recovery --- Ticker: IVZ --- GT: (B) Deepens} \\[4pt]
\begin{minipage}{0.95\linewidth}
\fbox{\parbox{\dimexpr\linewidth-2\fboxsep-2\fboxrule}{
\textbf{Input:} You are analyzing the stock IVZ. Below are the daily closing prices: [\textit{prices shown above}]. The stock has experienced a drawdown of 25.4\% from its recent peak. Predict the behavior over the next 20 days. \\
\hspace{6pt}(A) Recovery \\
\hspace{6pt}(B) Deepens
}} \\[4pt]
\fbox{\parbox{\dimexpr\linewidth-2\fboxsep-2\fboxrule}{
\textbf{Output:} \\[6pt]
\texttt{<think>} \\[4pt]
Step 1 --- Drawdown: 25.4\% (peak=16.18, current=12.07). \\[6pt]
Step 2 --- Peak was 62 days ago. \\[6pt]
Step 3 --- Scenario analysis: \\
\hspace{2em}$\bullet$ \textit{Base case} (most probable --- no major external events): \\
\hspace{2.7em}Selling pressure continues as the fundamental weakness persists. The drawdown deepens before eventually finding a floor. \\[4pt]
\hspace{2em}$\bullet$ \textit{Adverse scenario} (external shock reverses expectation): \\
\hspace{2.7em}A surprise positive development reverses the decline --- but this requires a specific external trigger. \\[4pt]
\hspace{2em}$\bullet$ \textit{Favorable scenario} (catalyst strengthens expectation): \\
\hspace{2.7em}The decline accelerates as market-wide risk-off sentiment compounds the stock-specific weakness. \\[6pt]
Step 4 --- Scenario assessment: \\
The base case and favorable scenario (2 of 3) support Deepens. \\
The adverse scenario requires a specific external trigger not present in current data. \\
Without new information, the base case is the most probable outcome. \\[6pt]
Step 5 --- Judgment: Base case favors Deepens $\rightarrow$ (B). \\[4pt]
\texttt{</think>} \\
\texttt{<answer>(B)</answer>}
}}
\end{minipage}
\caption{(Single-Prediction) Example of Drawdown Recovery task.}
\end{figure*}

\begin{figure*}[h]
\centering
\adjustbox{width=\linewidth}{%
\includegraphics[width=0.95\linewidth]{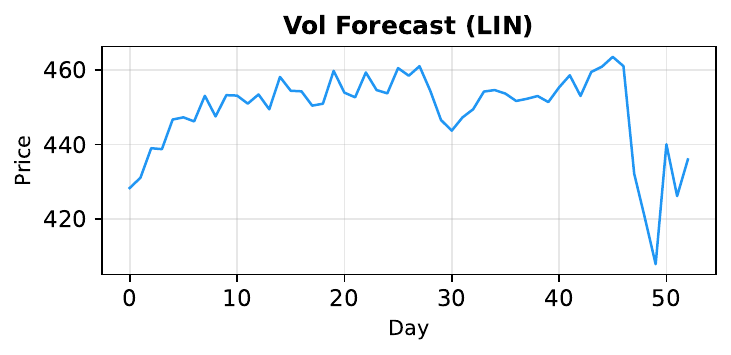}
}
\vspace{4pt}

\textbf{Task: Volatility Forecast --- Ticker: LIN --- GT: (B) Volatility decreases} \\[4pt]
\begin{minipage}{0.95\linewidth}
\fbox{\parbox{\dimexpr\linewidth-2\fboxsep-2\fboxrule}{
\textbf{Input:} You are analyzing the stock LIN. Below are the daily closing prices: [\textit{prices shown above}]. The current volatility ratio (recent 20-day vs overall) is 1.43. Predict how volatility will change over the next 20 days. \\
\hspace{6pt}(A) Increases \\
\hspace{6pt}(B) Decreases
}} \\[4pt]
\fbox{\parbox{\dimexpr\linewidth-2\fboxsep-2\fboxrule}{
\textbf{Output:} \\[6pt]
\texttt{<think>} \\[4pt]
Step 1 --- Vol: recent=2.59\%, long=1.81\%, ratio=1.43. \\[6pt]
Step 2 --- Volatility pattern assessed. \\[6pt]
Step 3 --- Scenario analysis: \\
\hspace{2em}$\bullet$ \textit{Base case} (most probable --- no major external events): \\
\hspace{2.7em}Volatility mean-reverts toward the long-run level as the recent shock dissipates and the market normalizes. \\[4pt]
\hspace{2em}$\bullet$ \textit{Adverse scenario} (external shock reverses expectation): \\
\hspace{2.7em}A new source of uncertainty (earnings, macro news) triggers another volatility spike. \\[4pt]
\hspace{2em}$\bullet$ \textit{Favorable scenario} (catalyst strengthens expectation): \\
\hspace{2.7em}Market calming accelerates as uncertainty resolves, pushing volatility down faster than baseline. \\[6pt]
Step 4 --- Scenario assessment: \\
The base case and favorable scenario (2 of 3) support Vol decreases. \\
The adverse scenario requires a specific external trigger not present in current data. \\
Without new information, the base case is the most probable outcome. \\[6pt]
Step 5 --- Judgment: Base case favors Vol decreases $\rightarrow$ (B). \\[4pt]
\texttt{</think>} \\
\texttt{<answer>(B)</answer>}
}}
\end{minipage}
\caption{(Single-Prediction) Example of Volatility Forecast task.}
\vspace{15pt}
\end{figure*}

\section{Comparison with Other Methods}
\label{app:qualitative}

We present three qualitative examples per task where only FinSTaR answers correctly while both baselines (Qwen2.5-7B and TimeOmni-1-7B, zero-shot) fail.
Figures~\ref{fig:cmp_first}--\ref{fig:cmp_last} show these comparisons, grouped by task (three examples each, in the order the tasks are introduced).


\begin{figure*}[h]
\centering
\caption{Example 1: Drawdown (Assessment) --- Ticker: ABBV --- GT: (C)}
\label{fig:cmp_first}
\vspace{2pt}

\noindent
\begin{minipage}[t]{0.28\linewidth}
\colorbox{red!8}{\parbox{\dimexpr\linewidth-2\fboxsep}{
\textbf{Qwen2.5-7B} \textcolor{red}{(B)} \ding{55}\\[2pt]
\tiny
[think]\newline To determine the current drawdown phase of ABBV, we need to identify the recent peak price and then calculate how far the current price is from that peak.\newline 1. Identify the highest price in the given data.\newline 2. Calculate the difference between the highest price and the current price.\newline 3. Determine the percentage decline based on the highest price.\newline From the data provided, the highest price is 17 [...]
}}
\end{minipage}
\hfill
\begin{minipage}[t]{0.28\linewidth}
\colorbox{red!8}{\parbox{\dimexpr\linewidth-2\fboxsep}{
\textbf{TimeOmni-1} \textcolor{red}{(B)} \ding{55}\\[2pt]
\tiny
[think]\newline Step 1: Identify the recent peak  \newline - The data covers 120 trading days.  \newline - The highest value in the series is 170.00 (last value), which is at index 119.  \newline - The previous day's price is 168.75 (index 118).  \newline - The day before that is 168.68 (index 117).  \newline - The day before that is 165.54 (index 116).  \newline - The day before that is 156.74 (index 115).  \newline - The day before that is 158.70 (index 114) [...]
}}
\end{minipage}
\hfill
\begin{minipage}[t]{0.41\linewidth}
\colorbox{green!10}{\parbox{\dimexpr\linewidth-2\fboxsep}{
\textbf{FinSTaR (Ours)} \textcolor{green!50!black}{(C)} \ding{51}\\[2pt]
\tiny
[think]\newline Step 1 -- Find the peak price:\newline Scanning the 120-day series, the highest price is 170.00 (around day 108).\newline Step 2 -- Current price:\newline The last price in the series is 152.33.\newline Step 3 -- Calculate drawdown:\newline Drawdown = (Peak - Current) / Peak = (170.00 - 152.33) / 170.00 = 10.4 pct\newline Step 4 -- Classify:\newline 10.4 pct is between 10 pct and 20 pct, which corresponds to (C) Correction.\newline [/think]\newline [answer]C[/answer]
}}
\end{minipage}
\vspace{8pt}
\end{figure*}

\begin{figure*}[h]
\centering
\caption{Example 2: Drawdown (Assessment) --- Ticker: DHR --- GT: (C)}
\vspace{2pt}

\noindent
\begin{minipage}[t]{0.28\linewidth}
\colorbox{red!8}{\parbox{\dimexpr\linewidth-2\fboxsep}{
\textbf{Qwen2.5-7B} \textcolor{red}{(A)} \ding{55}\\[2pt]
\tiny
[think]\newline To determine the current drawdown phase of the stock DHR, we need to identify its recent peak price and then compare the current price to this peak.\newline 1. Identify the Recent Peak Price:\newline    - The data provided shows the most recent 120 trading days' closing prices.\newline    - The highest value in this list is 211.53, which occurred on the last day of the provided data.\newline 2. Determine the Curren [...]
}}
\end{minipage}
\hfill
\begin{minipage}[t]{0.28\linewidth}
\colorbox{red!8}{\parbox{\dimexpr\linewidth-2\fboxsep}{
\textbf{TimeOmni-1} \textcolor{red}{(A)} \ding{55}\\[2pt]
\tiny
[think]\newline Step 1: Identify the most recent peak  \newline - The data shows a clear upward trend with some fluctuations, but the highest value in the series is at the end: 211.53 (last value).\newline - The previous high before this was 211.47 (second-to-last value), so the most recent peak is at or very close to the last value.\newline Step 2: Estimate the current price  \newline - The most recent value is 211.53.\newline - The next-to-la [...]
}}
\end{minipage}
\hfill
\begin{minipage}[t]{0.41\linewidth}
\colorbox{green!10}{\parbox{\dimexpr\linewidth-2\fboxsep}{
\textbf{FinSTaR (Ours)} \textcolor{green!50!black}{(C)} \ding{51}\\[2pt]
\tiny
[think]\newline Step 1 -- Find the peak price:\newline Scanning the 120-day series, the highest price is 211.53 (around day 103).\newline Step 2 -- Current price:\newline The last price in the series is 189.45.\newline Step 3 -- Calculate drawdown:\newline Drawdown = (Peak - Current) / Peak = (211.53 - 189.45) / 211.53 = 10.5 pct\newline Step 4 -- Classify:\newline 10.5 pct is between 10 pct and 20 pct, which corresponds to (C) Correction.\newline [/think]\newline [answer]C[/answer]
}}
\end{minipage}
\vspace{8pt}
\end{figure*}

\begin{figure*}[h]
\centering
\caption{Example 3: Drawdown (Assessment) --- Ticker: CRWD --- GT: (C)}
\vspace{2pt}

\noindent
\begin{minipage}[t]{0.28\linewidth}
\colorbox{red!8}{\parbox{\dimexpr\linewidth-2\fboxsep}{
\textbf{Qwen2.5-7B} \textcolor{red}{(B)} \ding{55}\\[2pt]
\tiny
[think]\newline To determine the current drawdown phase of the stock CRWD, we need to identify the recent peak price and then compare the current price to this peak.\newline 1. Identify the Recent Peak Price:\newline    - The data provided shows the most recent 120 trading days' closing prices.\newline    - The highest value in the list is 334.55, which occurred on one of the last few days.\newline 2. Determine the Current Price:* [...]
}}
\end{minipage}
\hfill
\begin{minipage}[t]{0.28\linewidth}
\colorbox{red!8}{\parbox{\dimexpr\linewidth-2\fboxsep}{
\textbf{TimeOmni-1} \textcolor{red}{(B)} \ding{55}\\[2pt]
\tiny
[think]\newline Step 1: Identify the recent peak  \newline - The data shows a clear upward trend with a few dips and rises.  \newline - The highest value in the series is 334.55 (at index 95).  \newline - The last value is 310.77 (index 119), which is lower than the peak but still relatively high.\newline Step 2: Estimate the current price  \newline - The last value is 310.77.  \newline - The previous value was 329.57 (index 118), so there has been a d [...]
}}
\end{minipage}
\hfill
\begin{minipage}[t]{0.41\linewidth}
\colorbox{green!10}{\parbox{\dimexpr\linewidth-2\fboxsep}{
\textbf{FinSTaR (Ours)} \textcolor{green!50!black}{(C)} \ding{51}\\[2pt]
\tiny
[think]\newline Step 1 -- Find the peak price:\newline Scanning the 120-day series, the highest price is 334.55 (around day 76).\newline Step 2 -- Current price:\newline The last price in the series is 299.15.\newline Step 3 -- Calculate drawdown:\newline Drawdown = (Peak - Current) / Peak = (334.55 - 299.15) / 334.55 = 10.6 pct\newline Step 4 -- Classify:\newline 10.6 pct is between 10 pct and 20 pct, which corresponds to (C) Correction.\newline [/think]\newline [answer]C[/answer]
}}
\vspace{-30pt}
\end{minipage}
\end{figure*}

\clearpage

\begin{figure*}[h]
\centering
\caption{Example 1: Correlation (Assessment) --- Ticker: AMD\_BAC --- GT: (A)}
\vspace{2pt}

\noindent
\begin{minipage}[t]{0.28\linewidth}
\colorbox{red!8}{\parbox{\dimexpr\linewidth-2\fboxsep}{
\textbf{Qwen2.5-7B} \textcolor{red}{(C)} \ding{55}\\[2pt]
\tiny
[think]To determine the correlation between the daily returns of Stock A and Stock B, we need to calculate the correlation coefficient. However, given the complexity and the fact that we are asked to choose from predefined options, we can analyze the general trend visually or through a simplified approach.\newline Firstly, let's observe the trends:\newline - Stock A's prices show a significant fluctuation with so [...]
}}
\end{minipage}
\hfill
\begin{minipage}[t]{0.28\linewidth}
\colorbox{red!8}{\parbox{\dimexpr\linewidth-2\fboxsep}{
\textbf{TimeOmni-1} \textcolor{red}{(B)} \ding{55}\\[2pt]
\tiny
[think]\newline Step 1: Trend Consistency (Perceptual Matching)\newline - Both Stock A (AMD) and Stock B (BAC) start with relatively stable/flat periods at the beginning.\newline - Both show a gradual rise in the first half of the series (e.g., both increase from  137-144 for AMD and  44-46 for BAC).\newline - Both then experience a decline or stabilization period (e.g., AMD falls from  144 down to  110, BAC falls from  46 down  [...]
}}
\end{minipage}
\hfill
\begin{minipage}[t]{0.41\linewidth}
\colorbox{green!10}{\parbox{\dimexpr\linewidth-2\fboxsep}{
\textbf{FinSTaR (Ours)} \textcolor{green!50!black}{(A)} \ding{51}\\[2pt]
\tiny
[think]\newline Step 1 -- Compare price changes in 12 periods ( 10 days each):\newline   Days 1-10: A=+2.8 pct, B=+1.2 pct\newline   Days 11-20: A=-14.3 pct, B=-7.2 pct\newline   Days 21-30: A=+8.6 pct, B=+2.8 pct\newline   Days 31-40: A=-2.8 pct, B=-0.6 pct\newline   Days 41-50: A=-8.9 pct, B=+1.5 pct\newline   Days 51-60: A=+3.6 pct, B=-5.1 pct\newline   Days 61-70: A=-10.8 pct, B=-7.5 pct\newline   Days 71-80: A=+10.0 pct, B=+2.6 pct\newline   Days 81-90: A=-23.9 pct, B=-12.4 pct\newline   Days 91-100: A=+3.1 pct, B=+8.8 pct\newline   Days 101-110: A=+10.1 pct, B=+11.2 pct\newline   Days 111-120: A=+20.0 pct, B=+8.7 pct\newline Step 2 -- Pattern:\newline When A rises, B also tends to rise, and when A falls, B also falls. The magnitudes are similar.\newline Step 3 -- positive correlation -> (A).\newline [/think]\newline [answer]A[/answer]
}}
\end{minipage}
\vspace{8pt}
\end{figure*}

\begin{figure*}[h]
\centering
\caption{Example 2: Correlation (Assessment) --- Ticker: CRM\_DECK --- GT: (A)}
\vspace{2pt}

\noindent
\begin{minipage}[t]{0.28\linewidth}
\colorbox{red!8}{\parbox{\dimexpr\linewidth-2\fboxsep}{
\textbf{Qwen2.5-7B} \textcolor{red}{(B)} \ding{55}\\[2pt]
\tiny
[think]To determine the correlation between the daily returns of Stock A and Stock B, we need to calculate the correlation coefficient. However, given the complexity and the fact that we are only asked to determine the nature of the correlation (positive, negative, or no significant), we can infer this by observing the general trend of the price movements.\newline For Stock A (CRM):\newline - The prices show some [...]
}}
\end{minipage}
\hfill
\begin{minipage}[t]{0.28\linewidth}
\colorbox{red!8}{\parbox{\dimexpr\linewidth-2\fboxsep}{
\textbf{TimeOmni-1} \textcolor{red}{(B)} \ding{55}\\[2pt]
\tiny
[think]\newline Step 1: Trend Consistency (Perceptual Matching)\newline - Both Stock A (CRM) and Stock B (DECK) start with relatively stable values for the first few days.\newline - Around day 10-15, both begin to rise: CRM goes from  282 to  339, DECK from  152 to  178.\newline - Both peak around day 15-16, then both decline slightly.\newline - Both show a pronounced rise again starting around day 25-30: CRM rises from  320 to  365, DE [...]
}}
\end{minipage}
\hfill
\begin{minipage}[t]{0.41\linewidth}
\colorbox{green!10}{\parbox{\dimexpr\linewidth-2\fboxsep}{
\textbf{FinSTaR (Ours)} \textcolor{green!50!black}{(A)} \ding{51}\\[2pt]
\tiny
[think]\newline Step 1 -- Compare price changes in 12 periods ( 10 days each):\newline   Days 1-10: A=+3.4 pct, B=+5.3 pct\newline   Days 11-20: A=+8.3 pct, B=+7.8 pct\newline   Days 21-30: A=+2.5 pct, B=+15.1 pct\newline   Days 31-40: A=-4.6 pct, B=+3.0 pct\newline   Days 41-50: A=-2.0 pct, B=+1.5 pct\newline   Days 51-60: A=-2.5 pct, B=+1.3 pct\newline   Days 61-70: A=+3.8 pct, B=-20.3 pct\newline   Days 71-80: A=-4.4 pct, B=-8.1 pct\newline   Days 81-90: A=-11.5 pct, B=-12.3 pct\newline   Days 91-100: A=-4.5 pct, B=-15.3 pct\newline   Days 101-110: A=-3.3 pct, B=-4.8 pct\newline   Days 111-120: A=-6.0 pct, B=-11.8 pct\newline Step 2 -- Pattern:\newline When A rises, B also tends to rise, and when A falls, B also falls. The magnitudes are similar.\newline Step 3 -- positive correlation -> (A).\newline [/think]\newline [answer]A[/answer]
}}
\end{minipage}
\vspace{8pt}
\end{figure*}

\begin{figure*}[h]
\centering
\caption{Example 3: Correlation (Assessment) --- Ticker: MRK\_EL --- GT: (A)}
\vspace{2pt}

\noindent
\begin{minipage}[t]{0.28\linewidth}
\colorbox{red!8}{\parbox{\dimexpr\linewidth-2\fboxsep}{
\textbf{Qwen2.5-7B} \textcolor{red}{(B)} \ding{55}\\[2pt]
\tiny
[think]To determine the correlation between the daily returns of Stock A and Stock B, we need to calculate the correlation coefficient. However, given the complexity and the fact that we are asked to choose from predefined options, we can analyze the general trend visually or through a simplified approach.\newline Firstly, let's observe the trends:\newline - Stock A's prices show fluctuations but generally seem t [...]
}}
\end{minipage}
\hfill
\begin{minipage}[t]{0.28\linewidth}
\colorbox{red!8}{\parbox{\dimexpr\linewidth-2\fboxsep}{
\textbf{TimeOmni-1} \textcolor{red}{(B)} \ding{55}\\[2pt]
\tiny
[think]\newline Step 1: Trend Consistency (Perceptual Matching)\newline - Both Stock A and Stock B start with relatively stable values for about the first 10-15 days.\newline - Around day 15-20, both begin to rise: Stock A goes from  91 to  97, while Stock B rises from  62 to  72.\newline - Both then continue rising, peaking around day 25-30 (A:  98, B:  78).\newline - After this peak, both decline: Stock A falls from  98 down to  85 by [...]
}}
\end{minipage}
\hfill
\begin{minipage}[t]{0.41\linewidth}
\colorbox{green!10}{\parbox{\dimexpr\linewidth-2\fboxsep}{
\textbf{FinSTaR (Ours)} \textcolor{green!50!black}{(A)} \ding{51}\\[2pt]
\tiny
[think]\newline Step 1 -- Compare price changes in 12 periods ( 10 days each):\newline   Days 1-10: A=-2.7 pct, B=-24.0 pct\newline   Days 11-20: A=-3.0 pct, B=+4.6 pct\newline   Days 21-30: A=+4.0 pct, B=+14.0 pct\newline   Days 31-40: A=-4.8 pct, B=-9.0 pct\newline   Days 41-50: A=+1.6 pct, B=-0.8 pct\newline   Days 51-60: A=-3.2 pct, B=+8.8 pct\newline   Days 61-70: A=-7.8 pct, B=-22.2 pct\newline   Days 71-80: A=+2.5 pct, B=+10.4 pct\newline   Days 81-90: A=+3.7 pct, B=-1.9 pct\newline   Days 91-100: A=-1.4 pct, B=-7.1 pct\newline   Days 101-110: A=-11.8 pct, B=-21.5 pct\newline   Days 111-120: A=-4.1 pct, B=+2.7 pct\newline Step 2 -- Pattern:\newline When A rises, B also tends to rise, and when A falls, B also falls. The magnitudes are similar.\newline Step 3 -- positive correlation -> (A).\newline [/think]\newline [answer]A[/answer]
}}
\vspace{-30pt}
\end{minipage}
\end{figure*}

\clearpage

\begin{figure*}[h]
\centering
\caption{Example 1: Volatility Regime (Assessment) --- Ticker: BKNG --- GT: (C)}
\vspace{2pt}

\noindent
\begin{minipage}[t]{0.28\linewidth}
\colorbox{red!8}{\parbox{\dimexpr\linewidth-2\fboxsep}{
\textbf{Qwen2.5-7B} \textcolor{red}{(B)} \ding{55}\\[2pt]
\tiny
[think]\newline To determine the volatility regimes, we need to calculate both the recent (last 20 days) and overall (120-day) volatility. Volatility can be measured using the standard deviation of the returns.\newline 1. Calculate the daily returns for each day.\newline 2. Compute the standard deviation of these returns for the last 20 days (recent volatility).\newline 3. Compute the standard deviation of these returns for the  [...]
}}
\end{minipage}
\hfill
\begin{minipage}[t]{0.28\linewidth}
\colorbox{red!8}{\parbox{\dimexpr\linewidth-2\fboxsep}{
\textbf{TimeOmni-1} \textcolor{red}{(A)} \ding{55}\\[2pt]
\tiny
[think]\newline Step 1: Calculate the overall (120-day) volatility  \newline - The first 120 values are given.  \newline - Compute the standard deviation (sigma) of these 120 values.\newline Step 2: Calculate the recent (last 20 days) volatility  \newline - Take the last 20 values from the list.\newline - Compute the standard deviation (sigma recent) of these 20 values.\newline Step 3: Compute the ratio  \newline - Ratio = sigma recent / sigma overall\newline Step 4: Classify the vol [...]
}}
\end{minipage}
\hfill
\begin{minipage}[t]{0.41\linewidth}
\colorbox{green!10}{\parbox{\dimexpr\linewidth-2\fboxsep}{
\textbf{FinSTaR (Ours)} \textcolor{green!50!black}{(C)} \ding{51}\\[2pt]
\tiny
[think]\newline Step 1 -- Compute daily returns from prices:\newline   e.g. (3441.67->3435.57): -0.17 pct, (3435.57->3414.34): -0.61 pct, (3414.34->3431.52): +0.50 pct\newline   ...\newline Step 2 -- Overall volatility (std of all returns): 1.60 pct\newline Step 3 -- Recent 20-day volatility: 2.59 pct\newline Step 4 -- Ratio = 2.59 / 1.60 = 1.62\newline Recent volatility is well above the long-term average -- unusually turbulent.\newline -> (C) high volatility.\newline [/think]\newline [answer]C[/answer]
}}
\end{minipage}
\vspace{8pt}
\end{figure*}

\begin{figure*}[h]
\centering
\caption{Example 2: Volatility Regime (Assessment) --- Ticker: BKNG --- GT: (C)}
\vspace{2pt}

\noindent
\begin{minipage}[t]{0.28\linewidth}
\colorbox{red!8}{\parbox{\dimexpr\linewidth-2\fboxsep}{
\textbf{Qwen2.5-7B} \textcolor{red}{(B)} \ding{55}\\[2pt]
\tiny
[think]\newline To determine the volatility regime, we need to calculate both the recent (last 20 days) and overall (120-day) volatility and then compute their ratio.\newline 1. Calculate the overall volatility (120-day):\newline    - First, find the mean of the 120-day closing prices.\newline    - Then, calculate the standard deviation of these prices.\newline 2. Calculate the recent volatility (last 20 days):\newline    - Find the mea [...]
}}
\end{minipage}
\hfill
\begin{minipage}[t]{0.28\linewidth}
\colorbox{red!8}{\parbox{\dimexpr\linewidth-2\fboxsep}{
\textbf{TimeOmni-1} \textcolor{red}{(A)} \ding{55}\\[2pt]
\tiny
[think]\newline Step 1: Calculate the overall (120-day) volatility  \newline - The first 120 values are given.  \newline - To estimate volatility, we can use the standard deviation of these values.\newline Let's compute the mean and standard deviation for the full 120-day series:\newline Mean (approximate calculation):\newline Sum = 3086.97 + ... + 3496.08 + 3558.52 + 3578.10 + 3640.90 + 3769.80 + 3689.26 + 3720.72 + 3678.88 + 3669.75 + 3681.88 [...]
}}
\end{minipage}
\hfill
\begin{minipage}[t]{0.41\linewidth}
\colorbox{green!10}{\parbox{\dimexpr\linewidth-2\fboxsep}{
\textbf{FinSTaR (Ours)} \textcolor{green!50!black}{(C)} \ding{51}\\[2pt]
\tiny
[think]\newline Step 1 -- Compute daily returns from prices:\newline   e.g. (3086.97->3083.60): -0.11 pct, (3083.60->3046.54): -1.20 pct, (3046.54->3085.46): +1.28 pct\newline   ...\newline Step 2 -- Overall volatility (std of all returns): 1.62 pct\newline Step 3 -- Recent 20-day volatility: 2.67 pct\newline Step 4 -- Ratio = 2.67 / 1.62 = 1.65\newline Recent volatility is well above the long-term average -- unusually turbulent.\newline -> (C) high volatility.\newline [/think]\newline [answer]C[/answer]
}}
\end{minipage}
\vspace{8pt}
\end{figure*}

\begin{figure*}[h]
\centering
\caption{Example 3: Volatility Regime (Assessment) --- Ticker: GL --- GT: (C)}
\vspace{2pt}

\noindent
\begin{minipage}[t]{0.28\linewidth}
\colorbox{red!8}{\parbox{\dimexpr\linewidth-2\fboxsep}{
\textbf{Qwen2.5-7B} \textcolor{red}{(A)} \ding{55}\\[2pt]
\tiny
[think]\newline To determine the volatility regimes, we need to calculate both the recent (last 20 days) and overall (120-day) volatilities and then compute their ratio.\newline 1. Calculate the overall volatility (120-day):\newline    - First, find the mean of the 120-day closing prices.\newline    - Then, calculate the standard deviation of these prices.\newline 2. Calculate the recent volatility (last 20 days):\newline    - Find the  [...]
}}
\end{minipage}
\hfill
\begin{minipage}[t]{0.28\linewidth}
\colorbox{red!8}{\parbox{\dimexpr\linewidth-2\fboxsep}{
\textbf{TimeOmni-1} \textcolor{red}{(B)} \ding{55}\\[2pt]
\tiny
[think]\newline Step 1: Calculate the overall volatility (120-day standard deviation)\newline First, I'll compute the mean of the entire series.\newline Summing up the first few values:\newline - The first 20 values are mostly in the 114-122 range.\newline - The next 20 values are in the 117-124 range.\newline - The next 20 values are in the 120-126 range.\newline - The next 20 values are in the 123-126 range.\newline - The next 20 values are in the 124-126 ra [...]
}}
\end{minipage}
\hfill
\begin{minipage}[t]{0.41\linewidth}
\colorbox{green!10}{\parbox{\dimexpr\linewidth-2\fboxsep}{
\textbf{FinSTaR (Ours)} \textcolor{green!50!black}{(C)} \ding{51}\\[2pt]
\tiny
[think]\newline Step 1 -- Compute daily returns from prices:\newline   e.g. (114.11->114.12): +0.01 pct, (114.12->113.63): -0.43 pct, (113.63->114.89): +1.10 pct\newline   ...\newline Step 2 -- Overall volatility (std of all returns): 4.51 pct\newline Step 3 -- Recent 20-day volatility: 10.75 pct\newline Step 4 -- Ratio = 10.75 / 4.51 = 2.38\newline Recent volatility is well above the long-term average -- unusually turbulent.\newline -> (C) high volatility.\newline [/think]\newline [answer]C[/answer]
}}
\vspace{-30pt}
\end{minipage}
\end{figure*}

\clearpage

\begin{figure*}[h]
\centering
\caption{Example 1: Trend Direction (Assessment) --- Ticker: BKNG --- GT: (B)}
\vspace{2pt}

\noindent
\begin{minipage}[t]{0.28\linewidth}
\colorbox{red!8}{\parbox{\dimexpr\linewidth-2\fboxsep}{
\textbf{Qwen2.5-7B} \textcolor{red}{(A)} \ding{55}\\[2pt]
\tiny
[think]\newline To determine the trend direction, we need to analyze the overall change in the stock price over the 120-day period.\newline 1. Calculate the first and last prices:\newline    - First price: 2864.00\newline    - Last price: 3501.24\newline 2. Calculate the percentage change:\newline    [\newline    (Percentage Change) = ( ((Last Price) - (First Price))((First Price)) )  100\newline    ]\newline    [\newline    (Percentage Change) = ( (3501.24 - 2864.00)(2864.00 [...]
}}
\end{minipage}
\hfill
\begin{minipage}[t]{0.28\linewidth}
\colorbox{red!8}{\parbox{\dimexpr\linewidth-2\fboxsep}{
\textbf{TimeOmni-1} \textcolor{red}{(C)} \ding{55}\\[2pt]
\tiny
[think]\newline Step 1: Trend Consistency (Perceptual Matching)\newline Let's visually scan the series for broad structural trends.\newline - The first  40 data points show a general rise: from around 2864 up to about 3183 by day 12.\newline - There is a noticeable drop after day 12, with prices falling back toward 2700s by day 50.\newline - After day 50, there is another rise, peaking at about 3500 by day 110.\newline - The last 10-15 points s [...]
}}
\end{minipage}
\hfill
\begin{minipage}[t]{0.41\linewidth}
\colorbox{green!10}{\parbox{\dimexpr\linewidth-2\fboxsep}{
\textbf{FinSTaR (Ours)} \textcolor{green!50!black}{(B)} \ding{51}\\[2pt]
\tiny
[think]\newline Step 1 -- Read start and end prices:\newline Start (day 1): 2864.00\newline End (day 120): 3425.24\newline Step 2 -- Compute cumulative return:\newline Return = (3425.24 - 2864.00) / 2864.00 = +19.6 pct\newline Step 3 -- Classify:\newline +19.6 pct -> (B) Mild Uptrend (+5 pct to +20 pct).\newline [/think]\newline [answer]B[/answer]
}}
\end{minipage}
\vspace{8pt}
\end{figure*}

\begin{figure*}[h]
\centering
\caption{Example 2: Trend Direction (Assessment) --- Ticker: NVR --- GT: (B)}
\vspace{2pt}

\noindent
\begin{minipage}[t]{0.28\linewidth}
\colorbox{red!8}{\parbox{\dimexpr\linewidth-2\fboxsep}{
\textbf{Qwen2.5-7B} \textcolor{red}{(A)} \ding{55}\\[2pt]
\tiny
[think]\newline To determine the trend direction, we need to analyze the overall change in the stock price over the 120-day period.\newline 1. Calculate the first and last prices:\newline    - First price: 5173.62\newline    - Last price: 6466.00\newline 2. Calculate the percentage change:\newline    [\newline    (Percentage Change) = ((Last Price) - (First Price))((First Price))  100\newline    ]\newline    [\newline    (Percentage Change) = (6466.00 - 5173.62)(5173.62)  100 [...]
}}
\end{minipage}
\hfill
\begin{minipage}[t]{0.28\linewidth}
\colorbox{red!8}{\parbox{\dimexpr\linewidth-2\fboxsep}{
\textbf{TimeOmni-1} \textcolor{red}{(A)} \ding{55}\\[2pt]
\tiny
[think]\newline Step 1: Trend Consistency (Perceptual Matching)\newline First, I visually scan the series for overall trend direction. The data starts at  5174 and rises steadily through the first half, with some fluctuations but generally rising. Around day 60-70, there is a noticeable flattening or slight decline, then another rise toward the end. There are no clear, sustained declines; the largest drops are on [...]
}}
\end{minipage}
\hfill
\begin{minipage}[t]{0.41\linewidth}
\colorbox{green!10}{\parbox{\dimexpr\linewidth-2\fboxsep}{
\textbf{FinSTaR (Ours)} \textcolor{green!50!black}{(B)} \ding{51}\\[2pt]
\tiny
[think]\newline Step 1 -- Read start and end prices:\newline Start (day 1): 5173.62\newline End (day 120): 6078.96\newline Step 2 -- Compute cumulative return:\newline Return = (6078.96 - 5173.62) / 5173.62 = +17.5 pct\newline Step 3 -- Classify:\newline +17.5 pct -> (B) Mild Uptrend (+5 pct to +20 pct).\newline [/think]\newline [answer]B[/answer]
}}
\end{minipage}
\vspace{8pt}
\end{figure*}

\begin{figure*}[h]
\centering
\caption{Example 3: Trend Direction (Assessment) --- Ticker: BLK --- GT: (D)}
\vspace{2pt}

\noindent
\begin{minipage}[t]{0.28\linewidth}
\colorbox{red!8}{\parbox{\dimexpr\linewidth-2\fboxsep}{
\textbf{Qwen2.5-7B} \textcolor{red}{(E)} \ding{55}\\[2pt]
\tiny
[think]\newline To determine the trend direction, we need to analyze the overall movement of the stock price over the 120-day period.\newline 1. Calculate the percentage change from the first day to the last day:\newline    - First day price: 1021.89\newline    - Last day price: 800.99\newline    - Percentage change = ((Last day price - First day price) / First day price) * 100\newline    - Percentage change = ((800.99 - 1021.89) / 1021.89) [...]
}}
\end{minipage}
\hfill
\begin{minipage}[t]{0.28\linewidth}
\colorbox{red!8}{\parbox{\dimexpr\linewidth-2\fboxsep}{
\textbf{TimeOmni-1} \textcolor{red}{(E)} \ding{55}\\[2pt]
\tiny
[think]\newline Step 1: Trend Consistency (Perceptual Matching)\newline Let's visually scan the series for broad structural trends.\newline - The first  40 data points show a general rise: from around 1020 down to about 990, then up to 1030-1040, then back down to 980-990, and so on.\newline - There is a noticeable "bump" or local peak around index 22-24 (values near 1030-1040), followed by a fall.\newline - After that, there's a gradua [...]
}}
\end{minipage}
\hfill
\begin{minipage}[t]{0.41\linewidth}
\colorbox{green!10}{\parbox{\dimexpr\linewidth-2\fboxsep}{
\textbf{FinSTaR (Ours)} \textcolor{green!50!black}{(D)} \ding{51}\\[2pt]
\tiny
[think]\newline Step 1 -- Read start and end prices:\newline Start (day 1): 1021.89\newline End (day 120): 896.49\newline Step 2 -- Compute cumulative return:\newline Return = (896.49 - 1021.89) / 1021.89 = -12.3 pct\newline Step 3 -- Classify:\newline -12.3 pct -> (D) Mild Downtrend (-5 pct to -20 pct).\newline [/think]\newline [answer]D[/answer]
}}
\vspace{-30pt}
\end{minipage}
\end{figure*}

\clearpage

\begin{figure*}[h]
\centering
\caption{Example 1: Event Response (Prediction) --- Ticker: AMAT --- GT: (B)}
\vspace{2pt}

\noindent
\begin{minipage}[t]{0.28\linewidth}
\colorbox{red!8}{\parbox{\dimexpr\linewidth-2\fboxsep}{
\textbf{Qwen2.5-7B} \textcolor{red}{(A)} \ding{55}\\[2pt]
\tiny
[think]\newline To determine whether the price is likely to mean-revert or persist, we need to analyze the recent trends and the nature of the shock.\newline 1. Trend Analysis:\newline    - The data shows a mix of increases and decreases over the past 120 days, with no clear long-term upward or downward trend.\newline    - There are periods of volatility, but overall, the stock price has been relatively stable around a centr [...]
}}
\end{minipage}
\hfill
\begin{minipage}[t]{0.28\linewidth}
\colorbox{red!8}{\parbox{\dimexpr\linewidth-2\fboxsep}{
\textbf{TimeOmni-1} \textcolor{red}{(A)} \ding{55}\\[2pt]
\tiny
[think]\newline Step 1: Identify the baseline trend and recent pattern  \newline - The series shows a general upward trend with some volatility.  \newline - There is a noticeable rise around the first half (e.g., from  180 to  193), then a decline, then another rise, and then a more pronounced drop towards the end ( 140s).  \newline - The last few days show a sharp drop (from  180 down to  125), suggesting a major event or rever [...]
}}
\end{minipage}
\hfill
\begin{minipage}[t]{0.41\linewidth}
\colorbox{green!10}{\parbox{\dimexpr\linewidth-2\fboxsep}{
\textbf{FinSTaR (Ours)} \textcolor{green!50!black}{(B)} \ding{51}\\[2pt]
\tiny
[think]\newline Step 1 -- Event: positive shock, z=4.52, return=+13.70 pct.\newline Step 2 -- Pre-event trend: downward.\newline Step 3 -- Scenario analysis:\newline   Base case (most probable -- no major external events):\newline     The shock reflects genuine new information being priced in. The trend continues as the market adjusts to the new fundamental value.\newline   Adverse scenario (external shock reverses expectation):\newline     Mean-reversion forces kick in -- the initial shock proves to be an overreaction and the price reverts.\newline   Favorable scenario (catalyst strengthens expectation):\newline     Additional confirming news amplifies the original move, with momentum traders piling in.\newline Step 4 -- Scenario assessment:\newline   The base case and favorable scenario (2 of 3) support Persistence.\newline   The adverse scenario requires a specific external trigger not present in current data.\newline   Without new information, the base case is the most probable outcome.\newline Step 5 -- Judgment: Base case favors Persistence -> (B).\newline [/think]\newline [answer]B[/answer]
}}
\end{minipage}
\vspace{8pt}
\end{figure*}

\begin{figure*}[h]
\centering
\caption{Example 2: Event Response (Prediction) --- Ticker: DXCM --- GT: (B)}
\vspace{2pt}

\noindent
\begin{minipage}[t]{0.28\linewidth}
\colorbox{red!8}{\parbox{\dimexpr\linewidth-2\fboxsep}{
\textbf{Qwen2.5-7B} \textcolor{red}{(A)} \ding{55}\\[2pt]
\tiny
[think]\newline To determine whether the price is likely to mean-revert or persist, we need to analyze the recent trends and the nature of the shock.\newline 1. Trend Analysis:\newline    - The data shows a mix of increases and decreases, but there's no clear long-term upward or downward trend over the past 120 days. There are periods of increase followed by decrease, indicating some volatility but no strong directio [...]
}}
\end{minipage}
\hfill
\begin{minipage}[t]{0.28\linewidth}
\colorbox{red!8}{\parbox{\dimexpr\linewidth-2\fboxsep}{
\textbf{TimeOmni-1} \textcolor{red}{(A)} \ding{55}\\[2pt]
\tiny
[think]\newline Step 1: Identify the baseline trend and recent pattern  \newline - The series shows a general upward trend with some volatility.  \newline - There is a noticeable rise around the middle third of the series (e.g., from  75 to  88), then a slight fall, then another rise, and finally a sharp spike at the end (last value: 81.62).\newline - The last few values before the spike were in the 80s, so the spike is a signif [...]
}}
\end{minipage}
\hfill
\begin{minipage}[t]{0.41\linewidth}
\colorbox{green!10}{\parbox{\dimexpr\linewidth-2\fboxsep}{
\textbf{FinSTaR (Ours)} \textcolor{green!50!black}{(B)} \ding{51}\\[2pt]
\tiny
[think]\newline Step 1 -- Event: positive shock, z=5.03, return=+15.23 pct.\newline Step 2 -- Pre-event trend: downward.\newline Step 3 -- Scenario analysis:\newline   Base case (most probable -- no major external events):\newline     The shock reflects genuine new information being priced in. The trend continues as the market adjusts to the new fundamental value.\newline   Adverse scenario (external shock reverses expectation):\newline     Mean-reversion forces kick in -- the initial shock proves to be an overreaction and the price reverts.\newline   Favorable scenario (catalyst strengthens expectation):\newline     Additional confirming news amplifies the original move, with momentum traders piling in.\newline Step 4 -- Scenario assessment:\newline   The base case and favorable scenario (2 of 3) support Persistence.\newline   The adverse scenario requires a specific external trigger not present in current data.\newline   Without new information, the base case is the most probable outcome.\newline Step 5 -- Judgment: Base case favors Persistence -> (B).\newline [/think]\newline [answer]B[/answer]
}}
\end{minipage}
\vspace{8pt}
\end{figure*}

\begin{figure*}[h]
\centering
\caption{Example 3: Event Response (Prediction) --- Ticker: NOW --- GT: (B)}
\vspace{2pt}

\noindent
\begin{minipage}[t]{0.28\linewidth}
\colorbox{red!8}{\parbox{\dimexpr\linewidth-2\fboxsep}{
\textbf{Qwen2.5-7B} \textcolor{red}{(A)} \ding{55}\\[2pt]
\tiny
[think]\newline To determine whether the price is likely to mean-revert or persist, we need to analyze the recent trends and the magnitude of the shock.\newline 1. Trend Analysis:\newline    - The data shows a mix of increases and decreases, but there's no clear long-term upward or downward trend over the past 120 days.\newline    - The last few days before the shock show a slight downward trend, with prices around 160-170.\newline  [...]
}}
\end{minipage}
\hfill
\begin{minipage}[t]{0.28\linewidth}
\colorbox{red!8}{\parbox{\dimexpr\linewidth-2\fboxsep}{
\textbf{TimeOmni-1} \textcolor{red}{(A)} \ding{55}\\[2pt]
\tiny
[think]\newline Step 1: Identify the baseline trend and recent pattern  \newline - The series shows a clear upward trend for the first  80 days, with a few dips but generally rising prices.\newline - Around day 80-90, there is a noticeable flattening or slight decline, suggesting a possible top or reversal.\newline - The last 10-15 days show a sharp drop, then a partial recovery, but still below previous highs.\newline Step 2: Assess th [...]
}}
\end{minipage}
\hfill
\begin{minipage}[t]{0.41\linewidth}
\colorbox{green!10}{\parbox{\dimexpr\linewidth-2\fboxsep}{
\textbf{FinSTaR (Ours)} \textcolor{green!50!black}{(B)} \ding{51}\\[2pt]
\tiny
[think]\newline Step 1 -- Event: positive shock, z=4.03, return=+12.93 pct.\newline Step 2 -- Pre-event trend: downward.\newline Step 3 -- Scenario analysis:\newline   Base case (most probable -- no major external events):\newline     The shock reflects genuine new information being priced in. The trend continues as the market adjusts to the new fundamental value.\newline   Adverse scenario (external shock reverses expectation):\newline     Mean-reversion forces kick in -- the initial shock proves to be an overreaction and the price reverts.\newline   Favorable scenario (catalyst strengthens expectation):\newline     Additional confirming news amplifies the original move, with momentum traders piling in.\newline Step 4 -- Scenario assessment:\newline   The base case and favorable scenario (2 of 3) support Persistence.\newline   The adverse scenario requires a specific external trigger not present in current data.\newline   Without new information, the base case is the most probable outcome.\newline Step 5 -- Judgment: Base case favors Persistence -> (B).\newline [/think]\newline [answer]B[/answer]
}}
\vspace{-30pt}
\end{minipage}
\end{figure*}

\clearpage

\begin{figure*}[h]
\centering
\caption{Example 1: Support/Resistance (Prediction) --- Ticker: BLK --- GT: (A)}
\vspace{2pt}

\noindent
\begin{minipage}[t]{0.28\linewidth}
\colorbox{red!8}{\parbox{\dimexpr\linewidth-2\fboxsep}{
\textbf{Qwen2.5-7B} \textcolor{red}{(B)} \ding{55}\\[2pt]
\tiny
[think]\newline To predict the stock's behavior, we need to analyze the recent trends and the current position relative to the resistance level.\newline 1. Identify the Resistance Level: The resistance level appears to be around 1163.39, which is the highest price in the given data.\newline 2. Recent Trends:\newline    - The stock has been fluctuating around this resistance level.\newline    - There have been attempts to break a [...]
}}
\end{minipage}
\hfill
\begin{minipage}[t]{0.28\linewidth}
\colorbox{red!8}{\parbox{\dimexpr\linewidth-2\fboxsep}{
\textbf{TimeOmni-1} \textcolor{red}{(B)} \ding{55}\\[2pt]
\tiny
[think]\newline Step 1: Identify the current trend and key levels  \newline - The data shows a clear upward trend with some volatility.  \newline - There is a noticeable rise starting around day 40-45, peaking at  1087.49 on day 49, then a decline to  1030.46 by day 52, followed by a new rise to  1167.06 by day 100.  \newline - The last few days (days 91-100) show a plateau or slight uptrend, but no sharp breakouts.  \newline - The most [...]
}}
\end{minipage}
\hfill
\begin{minipage}[t]{0.41\linewidth}
\colorbox{green!10}{\parbox{\dimexpr\linewidth-2\fboxsep}{
\textbf{FinSTaR (Ours)} \textcolor{green!50!black}{(A)} \ding{51}\\[2pt]
\tiny
[think]\newline Step 1 -- Key level: resistance at 1094.78.\newline Step 2 -- Proximity: 2.4 pct.\newline Step 3 -- Scenario analysis:\newline   Base case (most probable -- no major external events):\newline     The price breaks through the key level with sustained momentum, establishing a new trading range beyond the previous barrier.\newline   Adverse scenario (external shock reverses expectation):\newline     The breakout fails -- a false breakout traps traders and the price reverses sharply back through the level.\newline   Favorable scenario (catalyst strengthens expectation):\newline     The breakout is accompanied by high volume, leading to an accelerated move as stops are triggered.\newline Step 4 -- Scenario assessment:\newline   The base case and favorable scenario (2 of 3) support Breakout.\newline   The adverse scenario requires a specific external trigger not present in current data.\newline   Without new information, the base case is the most probable outcome.\newline Step 5 -- Judgment: Base case favors Breakout -> (A).\newline [/think]\newline [answer]A[/answer]
}}
\end{minipage}
\vspace{8pt}
\end{figure*}

\begin{figure*}[h]
\centering
\caption{Example 2: Support/Resistance (Prediction) --- Ticker: MSFT --- GT: (A)}
\vspace{2pt}

\noindent
\begin{minipage}[t]{0.28\linewidth}
\colorbox{red!8}{\parbox{\dimexpr\linewidth-2\fboxsep}{
\textbf{Qwen2.5-7B} \textcolor{red}{(B)} \ding{55}\\[2pt]
\tiny
[think]\newline To predict the stock's behavior, we need to analyze the recent trends and the current position relative to the resistance level.\newline 1. Identify the Resistance Level: The resistance level is a key price level where the stock has previously faced selling pressure, preventing it from rising further. From the data, it appears to be around 520 - 530.\newline 2. Recent Trends:\newline    - The stock has be [...]
}}
\end{minipage}
\hfill
\begin{minipage}[t]{0.28\linewidth}
\colorbox{red!8}{\parbox{\dimexpr\linewidth-2\fboxsep}{
\textbf{TimeOmni-1} \textcolor{red}{(B)} \ding{55}\\[2pt]
\tiny
[think]\newline Step 1: Identify the current context and trend  \newline - The data covers 120 trading days, with the last value being 515.52.\newline - There is a clear upward trend visible, with several periods of rising prices and some minor dips.\newline - The last few values (last 10-15) show a slight downtrend, but overall, the price is still higher than at the start of the series.\newline - A "key resistance level" is mentioned,  [...]
}}
\end{minipage}
\hfill
\begin{minipage}[t]{0.41\linewidth}
\colorbox{green!10}{\parbox{\dimexpr\linewidth-2\fboxsep}{
\textbf{FinSTaR (Ours)} \textcolor{green!50!black}{(A)} \ding{51}\\[2pt]
\tiny
[think]\newline Step 1 -- Key level: resistance at 510.74.\newline Step 2 -- Proximity: 0.9 pct.\newline Step 3 -- Scenario analysis:\newline   Base case (most probable -- no major external events):\newline     The price breaks through the key level with sustained momentum, establishing a new trading range beyond the previous barrier.\newline   Adverse scenario (external shock reverses expectation):\newline     The breakout fails -- a false breakout traps traders and the price reverses sharply back through the level.\newline   Favorable scenario (catalyst strengthens expectation):\newline     The breakout is accompanied by high volume, leading to an accelerated move as stops are triggered.\newline Step 4 -- Scenario assessment:\newline   The base case and favorable scenario (2 of 3) support Breakout.\newline   The adverse scenario requires a specific external trigger not present in current data.\newline   Without new information, the base case is the most probable outcome.\newline Step 5 -- Judgment: Base case favors Breakout -> (A).\newline [/think]\newline [answer]A[/answer]
}}
\end{minipage}
\vspace{8pt}
\end{figure*}

\begin{figure*}[h]
\centering
\caption{Example 3: Support/Resistance (Prediction) --- Ticker: HCA --- GT: (A)}
\vspace{2pt}

\noindent
\begin{minipage}[t]{0.28\linewidth}
\colorbox{red!8}{\parbox{\dimexpr\linewidth-2\fboxsep}{
\textbf{Qwen2.5-7B} \textcolor{red}{(B)} \ding{55}\\[2pt]
\tiny
[think]\newline To predict the stock's behavior, we need to analyze the recent trends and the current position relative to the resistance level.\newline 1. Identify the Resistance Level: The resistance level is a key price level where the stock has previously faced selling pressure, preventing it from rising further. From the data, the highest price in the last 120 days is around 339.58. We can assume this as [...]
}}
\end{minipage}
\hfill
\begin{minipage}[t]{0.28\linewidth}
\colorbox{red!8}{\parbox{\dimexpr\linewidth-2\fboxsep}{
\textbf{TimeOmni-1} \textcolor{red}{(B)} \ding{55}\\[2pt]
\tiny
[think]\newline Step 1: Identify the current context and trend  \newline - The data covers 120 trading days, with the last value being 339.58.\newline - There is a clear upward trend visible, with several periods of rising prices and some minor dips.\newline - The last few values (last 10-15) show a slight decline after a period of higher prices, but overall, the trend is upward.\newline - A "key resistance level" is mentioned, which is [...]
}}
\end{minipage}
\hfill
\begin{minipage}[t]{0.41\linewidth}
\colorbox{green!10}{\parbox{\dimexpr\linewidth-2\fboxsep}{
\textbf{FinSTaR (Ours)} \textcolor{green!50!black}{(A)} \ding{51}\\[2pt]
\tiny
[think]\newline Step 1 -- Key level: resistance at 328.67.\newline Step 2 -- Proximity: 3.3 pct.\newline Step 3 -- Scenario analysis:\newline   Base case (most probable -- no major external events):\newline     The price breaks through the key level with sustained momentum, establishing a new trading range beyond the previous barrier.\newline   Adverse scenario (external shock reverses expectation):\newline     The breakout fails -- a false breakout traps traders and the price reverses sharply back through the level.\newline   Favorable scenario (catalyst strengthens expectation):\newline     The breakout is accompanied by high volume, leading to an accelerated move as stops are triggered.\newline Step 4 -- Scenario assessment:\newline   The base case and favorable scenario (2 of 3) support Breakout.\newline   The adverse scenario requires a specific external trigger not present in current data.\newline   Without new information, the base case is the most probable outcome.\newline Step 5 -- Judgment: Base case favors Breakout -> (A).\newline [/think]\newline [answer]A[/answer]
}}
\vspace{-30pt}
\end{minipage}
\end{figure*}

\clearpage

\begin{figure*}[h]
\centering
\caption{Example 1: Relative Performance (Prediction) --- Ticker: MSCI\_NVDA --- GT: (A)}
\vspace{2pt}

\noindent
\begin{minipage}[t]{0.28\linewidth}
\colorbox{red!8}{\parbox{\dimexpr\linewidth-2\fboxsep}{
\textbf{Qwen2.5-7B} \textcolor{red}{(B)} \ding{55}\\[2pt]
\tiny
[think]\newline To determine which stock is likely to outperform over the next 20 trading days, we need to analyze recent trends and patterns in the closing prices.\newline For Stock A (MSCI):\newline - The closing prices show a significant drop from around 580 to 434.85 over the first half of the period.\newline - There's a recovery phase where prices rise to around 500 but then decline again.\newline - The last few days show a slight  [...]
}}
\end{minipage}
\hfill
\begin{minipage}[t]{0.28\linewidth}
\colorbox{red!8}{\parbox{\dimexpr\linewidth-2\fboxsep}{
\textbf{TimeOmni-1} \textcolor{red}{(B)} \ding{55}\\[2pt]
\tiny
[think]\newline Step 1: Trend Consistency (Perceptual Matching)\newline - Both MSCI and NVDA show a general upward trend in the first half of the series.\newline     - MSCI: Starts around 530, rises to  587 by day 10, then fluctuates with some declines but remains above 500 for most of the period.\newline     - NVDA: Starts at  57, rises to  92 by day 10, then fluctuates with some declines but remains above 70 for most of the pe [...]
}}
\end{minipage}
\hfill
\begin{minipage}[t]{0.41\linewidth}
\colorbox{green!10}{\parbox{\dimexpr\linewidth-2\fboxsep}{
\textbf{FinSTaR (Ours)} \textcolor{green!50!black}{(A)} \ding{51}\\[2pt]
\tiny
[think]\newline Step 1 -- Comparing MSCI vs NVDA.\newline Step 2 -- Forward window: 20 days.\newline Step 3 -- Scenario analysis:\newline   Base case (most probable -- no major external events):\newline     Current momentum trends persist. Stock A continues its recent trajectory relative to Stock B, as momentum typically sustains over 20-day windows.\newline   Adverse scenario (external shock reverses expectation):\newline     Stock A faces a company-specific headwind (earnings miss, product recall) while Stock B benefits from a positive catalyst, reversing the relative performance.\newline   Favorable scenario (catalyst strengthens expectation):\newline     Sector tailwinds or index inclusion further boosts Stock A's advantage over Stock B.\newline Step 4 -- Scenario assessment:\newline   The base case and favorable scenario (2 of 3) support Stock A outperforms.\newline   The adverse scenario requires a specific external trigger not present in current data.\newline   Without new information, the base case is the most probable outcome.\newline Step 5 -- Judgment: Base case favors Stock A outperforms -> (A).\newline [/think]\newline [answer](A)[/answer]
}}
\end{minipage}
\vspace{8pt}
\end{figure*}

\begin{figure*}[h]
\centering
\caption{Example 2: Relative Performance (Prediction) --- Ticker: CINF\_GRMN --- GT: (A)}
\vspace{2pt}

\noindent
\begin{minipage}[t]{0.28\linewidth}
\colorbox{red!8}{\parbox{\dimexpr\linewidth-2\fboxsep}{
\textbf{Qwen2.5-7B} \textcolor{red}{(B)} \ding{55}\\[2pt]
\tiny
[think]To determine which stock is likely to outperform the other over the next 20 trading days, we need to analyze recent trends and patterns in the closing prices of both stocks.\newline For Stock A (CINF), the closing prices show a mix of increases and decreases over the past 120 days. However, there seems to be a general downward trend towards the end of the period, with some fluctuations around this  [...]
}}
\end{minipage}
\hfill
\begin{minipage}[t]{0.28\linewidth}
\colorbox{red!8}{\parbox{\dimexpr\linewidth-2\fboxsep}{
\textbf{TimeOmni-1} \textcolor{red}{(B)} \ding{55}\\[2pt]
\tiny
[think]\newline Step 1: Trend Consistency (Perceptual Matching)\newline - Both CINF and GRMN start with relatively stable/flat periods for about the first 10-15 days.\newline - Around day 10-15, both begin to rise: CINF rises from  97 to  106, while GRMN rises from  90 to  96-97.\newline - Both then continue to rise, peaking around day 15-20: CINF peaks at  118, GRMN peaks at  98-99.\newline - After their respective peaks, both decline  [...]
}}
\end{minipage}
\hfill
\begin{minipage}[t]{0.41\linewidth}
\colorbox{green!10}{\parbox{\dimexpr\linewidth-2\fboxsep}{
\textbf{FinSTaR (Ours)} \textcolor{green!50!black}{(A)} \ding{51}\\[2pt]
\tiny
[think]\newline Step 1 -- Comparing CINF vs GRMN.\newline Step 2 -- Forward window: 20 days.\newline Step 3 -- Scenario analysis:\newline   Base case (most probable -- no major external events):\newline     Current momentum trends persist. Stock A continues its recent trajectory relative to Stock B, as momentum typically sustains over 20-day windows.\newline   Adverse scenario (external shock reverses expectation):\newline     Stock A faces a company-specific headwind (earnings miss, product recall) while Stock B benefits from a positive catalyst, reversing the relative performance.\newline   Favorable scenario (catalyst strengthens expectation):\newline     Sector tailwinds or index inclusion further boosts Stock A's advantage over Stock B.\newline Step 4 -- Scenario assessment:\newline   The base case and favorable scenario (2 of 3) support Stock A outperforms.\newline   The adverse scenario requires a specific external trigger not present in current data.\newline   Without new information, the base case is the most probable outcome.\newline Step 5 -- Judgment: Base case favors Stock A outperforms -> (A).\newline [/think]\newline [answer](A)[/answer]
}}
\end{minipage}
\vspace{8pt}
\end{figure*}

\begin{figure*}[h]
\centering
\caption{Example 3: Relative Performance (Prediction) --- Ticker: NTRS\_CTRA --- GT: (A)}
\vspace{2pt}

\noindent
\begin{minipage}[t]{0.28\linewidth}
\colorbox{red!8}{\parbox{\dimexpr\linewidth-2\fboxsep}{
\textbf{Qwen2.5-7B} \textcolor{red}{(B)} \ding{55}\\[2pt]
\tiny
[think]\newline To determine which stock is likely to outperform over the next 20 trading days, we need to analyze the recent trends and volatility of both stocks.\newline For Stock A (NTRS):\newline - The closing prices show a general upward trend with some fluctuations.\newline - There are periods of decline followed by recovery, indicating some volatility but overall positive movement.\newline For Stock B (CTRA):\newline - The closing prices [...]
}}
\end{minipage}
\hfill
\begin{minipage}[t]{0.28\linewidth}
\colorbox{red!8}{\parbox{\dimexpr\linewidth-2\fboxsep}{
\textbf{TimeOmni-1} \textcolor{red}{(B)} \ding{55}\\[2pt]
\tiny
[think]\newline Step 1: Trend Consistency (Perceptual Matching)\newline - Both NTRS and CTRA show a general downward trend in the first  20 days, then a stabilization or slight uptick around day 20-30, followed by a more pronounced rise in both stocks.\newline - Both stocks have a clear "stable -> rise" pattern after about day 30.\newline - The shapes match: both decline early, stabilize, then rise together.\newline - No major periods wh [...]
}}
\end{minipage}
\hfill
\begin{minipage}[t]{0.41\linewidth}
\colorbox{green!10}{\parbox{\dimexpr\linewidth-2\fboxsep}{
\textbf{FinSTaR (Ours)} \textcolor{green!50!black}{(A)} \ding{51}\\[2pt]
\tiny
[think]\newline Step 1 -- Comparing NTRS vs CTRA.\newline Step 2 -- Forward window: 20 days.\newline Step 3 -- Scenario analysis:\newline   Base case (most probable -- no major external events):\newline     Current momentum trends persist. Stock A continues its recent trajectory relative to Stock B, as momentum typically sustains over 20-day windows.\newline   Adverse scenario (external shock reverses expectation):\newline     Stock A faces a company-specific headwind (earnings miss, product recall) while Stock B benefits from a positive catalyst, reversing the relative performance.\newline   Favorable scenario (catalyst strengthens expectation):\newline     Sector tailwinds or index inclusion further boosts Stock A's advantage over Stock B.\newline Step 4 -- Scenario assessment:\newline   The base case and favorable scenario (2 of 3) support Stock A outperforms.\newline   The adverse scenario requires a specific external trigger not present in current data.\newline   Without new information, the base case is the most probable outcome.\newline Step 5 -- Judgment: Base case favors Stock A outperforms -> (A).\newline [/think]\newline [answer](A)[/answer]
}}
\vspace{-40pt}
\end{minipage}
\end{figure*}

\clearpage

\begin{figure*}[h]
\centering
\caption{Example 1: Drawdown Recovery (Prediction) --- Ticker: AZO --- GT: (A)}
\vspace{2pt}

\noindent
\begin{minipage}[t]{0.28\linewidth}
\colorbox{red!8}{\parbox{\dimexpr\linewidth-2\fboxsep}{
\textbf{Qwen2.5-7B} \textcolor{red}{(B)} \ding{55}\\[2pt]
\tiny
[think]\newline To predict the stock's behavior over the next 20 trading days, we need to analyze the recent trends and patterns in the given data.\newline 1. Identify the Recent Peak and Current Price: The recent peak is 3880.15, and the current price is 3603.17, indicating a significant drawdown of 7.1 pct.\newline 2. Analyze Recent Trends:\newline    - The last few prices show a fluctuating trend but generally moving  [...]
}}
\end{minipage}
\hfill
\begin{minipage}[t]{0.28\linewidth}
\colorbox{red!8}{\parbox{\dimexpr\linewidth-2\fboxsep}{
\textbf{TimeOmni-1} \textcolor{red}{(B)} \ding{55}\\[2pt]
\tiny
[think]\newline Step 1: Identify the recent trend and key patterns  \newline - The series shows a clear upward trend with some volatility, peaking at 3880.15 on day 89.\newline - There is a noticeable drop after this peak, reaching a trough at 3603.17 on day 90.\newline - The last 20 days show a gradual decline from the peak, with some minor fluctuations but no sharp upward moves.\newline Step 2: Estimate the direction of movement  \newline - T [...]
}}
\end{minipage}
\hfill
\begin{minipage}[t]{0.41\linewidth}
\colorbox{green!10}{\parbox{\dimexpr\linewidth-2\fboxsep}{
\textbf{FinSTaR (Ours)} \textcolor{green!50!black}{(A)} \ding{51}\\[2pt]
\tiny
[think]\newline Step 1 -- Drawdown: 7.1 pct (peak=3880.15, current=3603.17).\newline Step 2 -- Peak was 20 days ago.\newline Step 3 -- Scenario analysis:\newline   Base case (most probable -- no major external events):\newline     The drawdown stabilizes and the stock gradually recovers as value buyers accumulate shares. This is the typical pattern for moderate drawdowns in quality stocks.\newline   Adverse scenario (external shock reverses expectation):\newline     The underlying cause of the drawdown worsens (earnings revision, guidance cut), leading to further decline despite technical stabilization signals.\newline   Favorable scenario (catalyst strengthens expectation):\newline     A positive catalyst (analyst upgrade, strong earnings) triggers a sharp V-shaped recovery back to the peak.\newline Step 4 -- Scenario assessment:\newline   The base case and favorable scenario (2 of 3) support Recovery.\newline   The adverse scenario requires a specific external trigger not present in current data.\newline   Without new information, the base case is the most probable outcome.\newline Step 5 -- Judgment: Base case favors Recovery -> (A).\newline [/think]\newline [answer]A[/answer]
}}
\end{minipage}
\vspace{8pt}
\end{figure*}

\begin{figure*}[h]
\centering
\caption{Example 2: Drawdown Recovery (Prediction) --- Ticker: NVR --- GT: (A)}
\vspace{2pt}

\noindent
\begin{minipage}[t]{0.28\linewidth}
\colorbox{red!8}{\parbox{\dimexpr\linewidth-2\fboxsep}{
\textbf{Qwen2.5-7B} \textcolor{red}{(B)} \ding{55}\\[2pt]
\tiny
[think]\newline To predict the stock's behavior, we need to analyze the recent trends and patterns in the given data.\newline 1. Identify the Peak and Current Price: The peak price is 5967.22, and the current price is 5621.21, indicating a drawdown of 5.8 pct.\newline 2. Analyze Recent Trends:\newline    - The last few prices show a fluctuating trend but generally decreasing from around 5900 to 5600.\newline    - There have been [...]
}}
\end{minipage}
\hfill
\begin{minipage}[t]{0.28\linewidth}
\colorbox{red!8}{\parbox{\dimexpr\linewidth-2\fboxsep}{
\textbf{TimeOmni-1} \textcolor{red}{(B)} \ding{55}\\[2pt]
\tiny
[think]\newline Step 1: Identify the recent trend and key patterns  \newline - The series shows a clear upward trend with some volatility, peaking at 5967.22 on day 49.\newline - After the peak, there is a decline, reaching a local minimum around 5585.70 by day 66.\newline - The last 20 days show a gradual decline from  5887 to  5611, with no sharp upward moves.\newline Step 2: Estimate the current context  \newline - The most recent value is 5 [...]
}}
\end{minipage}
\hfill
\begin{minipage}[t]{0.41\linewidth}
\colorbox{green!10}{\parbox{\dimexpr\linewidth-2\fboxsep}{
\textbf{FinSTaR (Ours)} \textcolor{green!50!black}{(A)} \ding{51}\\[2pt]
\tiny
[think]\newline Step 1 -- Drawdown: 5.8 pct (peak=5967.22, current=5621.21).\newline Step 2 -- Peak was 26 days ago.\newline Step 3 -- Scenario analysis:\newline   Base case (most probable -- no major external events):\newline     The drawdown stabilizes and the stock gradually recovers as value buyers accumulate shares. This is the typical pattern for moderate drawdowns in quality stocks.\newline   Adverse scenario (external shock reverses expectation):\newline     The underlying cause of the drawdown worsens (earnings revision, guidance cut), leading to further decline despite technical stabilization signals.\newline   Favorable scenario (catalyst strengthens expectation):\newline     A positive catalyst (analyst upgrade, strong earnings) triggers a sharp V-shaped recovery back to the peak.\newline Step 4 -- Scenario assessment:\newline   The base case and favorable scenario (2 of 3) support Recovery.\newline   The adverse scenario requires a specific external trigger not present in current data.\newline   Without new information, the base case is the most probable outcome.\newline Step 5 -- Judgment: Base case favors Recovery -> (A).\newline [/think]\newline [answer]A[/answer]
}}
\end{minipage}
\vspace{8pt}
\end{figure*}

\begin{figure*}[h]
\centering
\caption{Example 3: Drawdown Recovery (Prediction) --- Ticker: NVDA --- GT: (A)}
\vspace{2pt}

\noindent
\begin{minipage}[t]{0.28\linewidth}
\colorbox{red!8}{\parbox{\dimexpr\linewidth-2\fboxsep}{
\textbf{Qwen2.5-7B} \textcolor{red}{(B)} \ding{55}\\[2pt]
\tiny
[think]\newline To predict the stock's behavior, we need to analyze the recent trends and patterns in the given data.\newline 1. Identify the Peak and Current Price: The peak price is 135.51, and the current price is 120.82, indicating a 10.8 pct drawdown from the peak.\newline 2. Trend Analysis:\newline    - The stock price has been fluctuating but generally shows a downward trend since reaching the peak.\newline    - There hav [...]
}}
\end{minipage}
\hfill
\begin{minipage}[t]{0.28\linewidth}
\colorbox{red!8}{\parbox{\dimexpr\linewidth-2\fboxsep}{
\textbf{TimeOmni-1} \textcolor{red}{(B)} \ding{55}\\[2pt]
\tiny
[think]\newline Step 1: Identify the recent trend and key patterns  \newline - The series shows a clear upward trend with several peaks and troughs.\newline - There is a notable rise starting around day 10-15, peaking at  135.51 (day 30), then a decline to  120.82 (day 98).\newline - The last 20 days show a gradual decline, with some minor fluctuations but no sharp upward moves.\newline Step 2: Estimate the current context  \newline - The most  [...]
}}
\end{minipage}
\hfill
\begin{minipage}[t]{0.41\linewidth}
\colorbox{green!10}{\parbox{\dimexpr\linewidth-2\fboxsep}{
\textbf{FinSTaR (Ours)} \textcolor{green!50!black}{(A)} \ding{51}\\[2pt]
\tiny
[think]\newline Step 1 -- Drawdown: 10.8 pct (peak=135.51, current=120.82).\newline Step 2 -- Peak was 68 days ago.\newline Step 3 -- Scenario analysis:\newline   Base case (most probable -- no major external events):\newline     The drawdown stabilizes and the stock gradually recovers as value buyers accumulate shares. This is the typical pattern for moderate drawdowns in quality stocks.\newline   Adverse scenario (external shock reverses expectation):\newline     The underlying cause of the drawdown worsens (earnings revision, guidance cut), leading to further decline despite technical stabilization signals.\newline   Favorable scenario (catalyst strengthens expectation):\newline     A positive catalyst (analyst upgrade, strong earnings) triggers a sharp V-shaped recovery back to the peak.\newline Step 4 -- Scenario assessment:\newline   The base case and favorable scenario (2 of 3) support Recovery.\newline   The adverse scenario requires a specific external trigger not present in current data.\newline   Without new information, the base case is the most probable outcome.\newline Step 5 -- Judgment: Base case favors Recovery -> (A).\newline [/think]\newline [answer]A[/answer]
}}
\vspace{-40pt}
\end{minipage}
\end{figure*}

\clearpage

\begin{figure*}[h]
\centering
\caption{Example 1: Volatility Forecast (Prediction) --- Ticker: BDX --- GT: (A)}
\vspace{2pt}

\noindent
\begin{minipage}[t]{0.28\linewidth}
\colorbox{red!8}{\parbox{\dimexpr\linewidth-2\fboxsep}{
\textbf{Qwen2.5-7B} \textcolor{red}{(B)} \ding{55}\\[2pt]
\tiny
[think]\newline To determine whether volatility will increase or decrease, we need to analyze the recent price movements and compare them with the historical volatility.\newline 1. Recent Price Movements: \newline    - The last 20 days' closing prices show a mix of increases and decreases, but there's no clear trend indicating a significant upward or downward movement.\newline    - The recent volatility ratio is 0.91, which  [...]
}}
\end{minipage}
\hfill
\begin{minipage}[t]{0.28\linewidth}
\colorbox{red!8}{\parbox{\dimexpr\linewidth-2\fboxsep}{
\textbf{TimeOmni-1} \textcolor{red}{(B)} \ding{55}\\[2pt]
\tiny
[think]\newline Step 1: Trend Consistency (Perceptual Matching)\newline - Looking at the series, there is a general downward trend with some fluctuations. The first half of the series shows a gradual decline from  185 to  170, then a slight uptick around day 20-25, followed by another decline and a small uptick near the end.\newline - The last 20 days (days 49-68) show a clear downtrend: prices start at  180 and graduall [...]
}}
\end{minipage}
\hfill
\begin{minipage}[t]{0.41\linewidth}
\colorbox{green!10}{\parbox{\dimexpr\linewidth-2\fboxsep}{
\textbf{FinSTaR (Ours)} \textcolor{green!50!black}{(A)} \ding{51}\\[2pt]
\tiny
[think]\newline Step 1 -- Vol: recent=1.06 pct, long=1.17 pct, ratio=0.91.\newline Step 2 -- Volatility pattern assessed.\newline Step 3 -- Scenario analysis:\newline   Base case (most probable -- no major external events):\newline     Volatility clustering continues -- the current elevated/rising volatility persists as uncertainty remains in the market.\newline   Adverse scenario (external shock reverses expectation):\newline     A resolution of the uncertainty driver (e.g., earnings release, policy decision) causes volatility to collapse rapidly.\newline   Favorable scenario (catalyst strengthens expectation):\newline     A new source of uncertainty emerges, compounding existing volatility and pushing it even higher.\newline Step 4 -- Scenario assessment:\newline   The base case and favorable scenario (2 of 3) support Vol increases.\newline   The adverse scenario requires a specific external trigger not present in current data.\newline   Without new information, the base case is the most probable outcome.\newline Step 5 -- Judgment: Base case favors Vol increases -> (A).\newline [/think]\newline [answer](A)[/answer]
}}
\end{minipage}
\vspace{8pt}
\end{figure*}

\begin{figure*}[h]
\centering
\caption{Example 2: Volatility Forecast (Prediction) --- Ticker: KEY --- GT: (A)}
\vspace{2pt}

\noindent
\begin{minipage}[t]{0.28\linewidth}
\colorbox{red!8}{\parbox{\dimexpr\linewidth-2\fboxsep}{
\textbf{Qwen2.5-7B} \textcolor{red}{(B)} \ding{55}\\[2pt]
\tiny
[think]\newline To determine whether volatility will increase or decrease, we need to analyze the recent price movements and the given volatility ratio.\newline 1. Price Trend Analysis:\newline    - The closing prices show a mix of increases and decreases, but there isn't a clear upward or downward trend in the last 77 days.\newline    - There are several periods where prices seem to fluctuate around a certain level before m [...]
}}
\end{minipage}
\hfill
\begin{minipage}[t]{0.28\linewidth}
\colorbox{red!8}{\parbox{\dimexpr\linewidth-2\fboxsep}{
\textbf{TimeOmni-1} \textcolor{red}{(B)} \ding{55}\\[2pt]
\tiny
[think]\newline Step 1: Trend Consistency (Perceptual Matching)\newline - Looking at the series, there is a clear downward trend in the first half (first  40 days), with prices gradually declining from around 10 to about 8. Then, after a brief flat period, there's a slight uptick in the last 37 days, but it's not as pronounced as the initial decline.\newline - The volatility ratio is given as 0.59, which means the recent [...]
}}
\end{minipage}
\hfill
\begin{minipage}[t]{0.41\linewidth}
\colorbox{green!10}{\parbox{\dimexpr\linewidth-2\fboxsep}{
\textbf{FinSTaR (Ours)} \textcolor{green!50!black}{(A)} \ding{51}\\[2pt]
\tiny
[think]\newline Step 1 -- Vol: recent=2.20 pct, long=3.74 pct, ratio=0.59.\newline Step 2 -- Volatility pattern assessed.\newline Step 3 -- Scenario analysis:\newline   Base case (most probable -- no major external events):\newline     Volatility clustering continues -- the current elevated/rising volatility persists as uncertainty remains in the market.\newline   Adverse scenario (external shock reverses expectation):\newline     A resolution of the uncertainty driver (e.g., earnings release, policy decision) causes volatility to collapse rapidly.\newline   Favorable scenario (catalyst strengthens expectation):\newline     A new source of uncertainty emerges, compounding existing volatility and pushing it even higher.\newline Step 4 -- Scenario assessment:\newline   The base case and favorable scenario (2 of 3) support Vol increases.\newline   The adverse scenario requires a specific external trigger not present in current data.\newline   Without new information, the base case is the most probable outcome.\newline Step 5 -- Judgment: Base case favors Vol increases -> (A).\newline [/think]\newline [answer](A)[/answer]
}}
\end{minipage}
\vspace{8pt}
\end{figure*}

\begin{figure*}[h]
\centering
\caption{Example 3: Volatility Forecast (Prediction) --- Ticker: MRK --- GT: (A)}
\vspace{2pt}

\noindent
\begin{minipage}[t]{0.28\linewidth}
\colorbox{red!8}{\parbox{\dimexpr\linewidth-2\fboxsep}{
\textbf{Qwen2.5-7B} \textcolor{red}{(B)} \ding{55}\\[2pt]
\tiny
[think]\newline To determine whether volatility will increase or decrease, we need to analyze the recent price movements and compare them with the historical volatility.\newline 1. Recent Price Movements: The last 20 days show a mix of increases and decreases, but generally, the prices have been trending upwards. The most recent prices are around 117-118, while the prices in the earlier part of the list were  [...]
}}
\end{minipage}
\hfill
\begin{minipage}[t]{0.28\linewidth}
\colorbox{red!8}{\parbox{\dimexpr\linewidth-2\fboxsep}{
\textbf{TimeOmni-1} \textcolor{red}{(B)} \ding{55}\\[2pt]
\tiny
[think]\newline Step 1: Trend Consistency (Perceptual Matching)\newline - Looking at the series, there is a clear upward trend in the first half (days 0-49), with prices rising from  94 to  110, then a plateau, followed by a second rise (days 50-74), peaking around 120, then a gradual decline back toward  115, and finally a small uptick at the end.\newline - The last 20 days (days 75-94) show a gentle decline from  120 d [...]
}}
\end{minipage}
\hfill
\begin{minipage}[t]{0.41\linewidth}
\colorbox{green!10}{\parbox{\dimexpr\linewidth-2\fboxsep}{
\textbf{FinSTaR (Ours)} \textcolor{green!50!black}{(A)} \ding{51}\\[2pt]
\tiny
[think]\newline Step 1 -- Vol: recent=0.83 pct, long=1.02 pct, ratio=0.81.\newline Step 2 -- Volatility pattern assessed.\newline Step 3 -- Scenario analysis:\newline   Base case (most probable -- no major external events):\newline     Volatility clustering continues -- the current elevated/rising volatility persists as uncertainty remains in the market.\newline   Adverse scenario (external shock reverses expectation):\newline     A resolution of the uncertainty driver (e.g., earnings release, policy decision) causes volatility to collapse rapidly.\newline   Favorable scenario (catalyst strengthens expectation):\newline     A new source of uncertainty emerges, compounding existing volatility and pushing it even higher.\newline Step 4 -- Scenario assessment:\newline   The base case and favorable scenario (2 of 3) support Vol increases.\newline   The adverse scenario requires a specific external trigger not present in current data.\newline   Without new information, the base case is the most probable outcome.\newline Step 5 -- Judgment: Base case favors Vol increases -> (A).\newline [/think]\newline [answer](A)[/answer]
}}
\vspace{-30pt}
\end{minipage}
\end{figure*}

\clearpage

\begin{figure*}[h]
\centering
\caption{Example 1: Pair Convergence (Prediction) --- Ticker: PANW\_CRWD --- GT: (A)}
\vspace{2pt}

\noindent
\begin{minipage}[t]{0.28\linewidth}
\colorbox{red!8}{\parbox{\dimexpr\linewidth-2\fboxsep}{
\textbf{Qwen2.5-7B} \textcolor{red}{(B)} \ding{55}\\[2pt]
\tiny
[think]\newline To determine whether the spread between PANW and CRWD will converge or diverge, we need to analyze the trends in both stock prices over the given period.\newline 1. Trend Analysis for PANW:\newline    - The closing prices of PANW show a general upward trend with some fluctuations.\newline    - There are periods where the stock price increases significantly (e.g., from around 180 to 200+), followed by slight d [...]
}}
\end{minipage}
\hfill
\begin{minipage}[t]{0.28\linewidth}
\colorbox{red!8}{\parbox{\dimexpr\linewidth-2\fboxsep}{
\textbf{TimeOmni-1} \textcolor{red}{(B)} \ding{55}\\[2pt]
\tiny
[think]\newline Step 1: Trend Consistency (Perceptual Matching)\newline - Both PANW and CRWD show an initial rise in the first  10 days.\newline - Both then decline for about 10-15 days.\newline - Both then rise again, with CRWD showing a more pronounced increase after day 20.\newline - Both have periods of stability and minor fluctuations.\newline - The overall shapes match: rises, falls, and recoveries occur at similar times, though CRWD is a [...]
}}
\end{minipage}
\hfill
\begin{minipage}[t]{0.41\linewidth}
\colorbox{green!10}{\parbox{\dimexpr\linewidth-2\fboxsep}{
\textbf{FinSTaR (Ours)} \textcolor{green!50!black}{(A)} \ding{51}\\[2pt]
\tiny
[think]\newline Step 1 -- Pair: ['PANW', 'CRWD'].\newline Step 2 -- Current spread: 0.325.\newline Step 3 -- Scenario analysis:\newline   Base case (most probable -- no major external events):\newline     The elevated spread narrows as the historical co-movement relationship reasserts itself. This is the typical mean-reversion pattern for correlated stocks.\newline   Adverse scenario (external shock reverses expectation):\newline     The divergence reflects a structural change (business pivot, sector reclassification) and the spread continues to widen.\newline   Favorable scenario (catalyst strengthens expectation):\newline     A common catalyst (sector earnings, index rebalancing) forces rapid convergence.\newline Step 4 -- Scenario assessment:\newline   The base case and favorable scenario (2 of 3) support Convergence.\newline   The adverse scenario requires a specific external trigger not present in current data.\newline   Without new information, the base case is the most probable outcome.\newline Step 5 -- Judgment: Base case favors Convergence -> (A).\newline [/think]\newline [answer](A)[/answer]
}}
\end{minipage}
\vspace{8pt}
\end{figure*}

\begin{figure*}[h]
\centering
\caption{Example 2: Pair Convergence (Prediction) --- Ticker: ABBV\_COIN --- GT: (A)}
\vspace{2pt}

\noindent
\begin{minipage}[t]{0.28\linewidth}
\colorbox{red!8}{\parbox{\dimexpr\linewidth-2\fboxsep}{
\textbf{Qwen2.5-7B} \textcolor{red}{(B)} \ding{55}\\[2pt]
\tiny
[think]\newline To determine whether the spread between ABBV and COIN will converge or diverge, we need to analyze the trends in their respective price movements.\newline 1. Trend Analysis for ABBV:\newline    - The closing prices of ABBV show a general upward trend with some fluctuations.\newline    - There are periods where the stock price increases significantly (e.g., from 156.50 to 189.56), followed by periods of stabil [...]
}}
\end{minipage}
\hfill
\begin{minipage}[t]{0.28\linewidth}
\colorbox{red!8}{\parbox{\dimexpr\linewidth-2\fboxsep}{
\textbf{TimeOmni-1} \textcolor{red}{(B)} \ding{55}\\[2pt]
\tiny
[think]\newline Step 1: Trend Consistency (Perceptual Matching)\newline - Both ABBV and COIN show an initial decline in the first  10 days.\newline - Both then rise for about 10-15 days, peaking around day 25-30.\newline - Both then fall again, with a more pronounced drop for COIN starting around day 30-35.\newline - Both have a secondary rise after the mid-point, peaking again around day 50-55.\newline - Both then fall again, with a smaller di [...]
}}
\end{minipage}
\hfill
\begin{minipage}[t]{0.41\linewidth}
\colorbox{green!10}{\parbox{\dimexpr\linewidth-2\fboxsep}{
\textbf{FinSTaR (Ours)} \textcolor{green!50!black}{(A)} \ding{51}\\[2pt]
\tiny
[think]\newline Step 1 -- Pair: ['ABBV', 'COIN'].\newline Step 2 -- Current spread: 0.346.\newline Step 3 -- Scenario analysis:\newline   Base case (most probable -- no major external events):\newline     The elevated spread narrows as the historical co-movement relationship reasserts itself. This is the typical mean-reversion pattern for correlated stocks.\newline   Adverse scenario (external shock reverses expectation):\newline     The divergence reflects a structural change (business pivot, sector reclassification) and the spread continues to widen.\newline   Favorable scenario (catalyst strengthens expectation):\newline     A common catalyst (sector earnings, index rebalancing) forces rapid convergence.\newline Step 4 -- Scenario assessment:\newline   The base case and favorable scenario (2 of 3) support Convergence.\newline   The adverse scenario requires a specific external trigger not present in current data.\newline   Without new information, the base case is the most probable outcome.\newline Step 5 -- Judgment: Base case favors Convergence -> (A).\newline [/think]\newline [answer](A)[/answer]
}}
\end{minipage}
\vspace{8pt}
\end{figure*}

\begin{figure*}[h]
\centering
\caption{Example 3: Pair Convergence (Prediction) --- Ticker: ABBV\_META --- GT: (A)}
\label{fig:cmp_last}
\vspace{2pt}

\noindent
\begin{minipage}[t]{0.28\linewidth}
\colorbox{red!8}{\parbox{\dimexpr\linewidth-2\fboxsep}{
\textbf{Qwen2.5-7B} \textcolor{red}{(B)} \ding{55}\\[2pt]
\tiny
[think]\newline To determine whether the spread between ABBV and META will converge or diverge, we need to analyze the trends in their respective price movements.\newline 1. Trend Analysis for ABBV:\newline    - The closing prices of ABBV show a significant fluctuation but generally seem to be decreasing over time. There are periods of slight increases, but overall, the trend is downward.\newline 2. Trend Analysis for META [...]
}}
\end{minipage}
\hfill
\begin{minipage}[t]{0.28\linewidth}
\colorbox{red!8}{\parbox{\dimexpr\linewidth-2\fboxsep}{
\textbf{TimeOmni-1} \textcolor{red}{(B)} \ding{55}\\[2pt]
\tiny
[think]\newline Step 1: Trend Consistency (Perceptual Matching)\newline - Both ABBV and META show an initial rise in the first 10-15 days.\newline     - ABBV: Starts  144.80, rises to  149.32 by day 10, then falls back to  134.80 by day 15.\newline     - META: Starts  212.18, rises to  241.29 by day 15, then falls back to  231.70 by day 20.\newline - Both then fall for about 10-15 days, with some minor fluctuations.\newline     - ABBV: Drops fr [...]
}}
\end{minipage}
\hfill
\begin{minipage}[t]{0.41\linewidth}
\colorbox{green!10}{\parbox{\dimexpr\linewidth-2\fboxsep}{
\textbf{FinSTaR (Ours)} \textcolor{green!50!black}{(A)} \ding{51}\\[2pt]
\tiny
[think]\newline Step 1 -- Pair: ['ABBV', 'META'].\newline Step 2 -- Current spread: 0.547.\newline Step 3 -- Scenario analysis:\newline   Base case (most probable -- no major external events):\newline     The elevated spread narrows as the historical co-movement relationship reasserts itself. This is the typical mean-reversion pattern for correlated stocks.\newline   Adverse scenario (external shock reverses expectation):\newline     The divergence reflects a structural change (business pivot, sector reclassification) and the spread continues to widen.\newline   Favorable scenario (catalyst strengthens expectation):\newline     A common catalyst (sector earnings, index rebalancing) forces rapid convergence.\newline Step 4 -- Scenario assessment:\newline   The base case and favorable scenario (2 of 3) support Convergence.\newline   The adverse scenario requires a specific external trigger not present in current data.\newline   Without new information, the base case is the most probable outcome.\newline Step 5 -- Judgment: Base case favors Convergence -> (A).\newline [/think]\newline [answer](A)[/answer]
}}
\vspace{-30pt}
\end{minipage}
\end{figure*}

\end{document}